\pdfoutput=1
\documentclass[10pt,twocolumn,letterpaper]{article}

\newif\ifarxiv
\arxivtrue

\usepackage[pagenumbers]{iccv}
\usepackage[absolute,overlay]{textpos}

\usepackage{amsmath,amssymb,amsfonts}
\usepackage{graphicx}
\usepackage{booktabs}

\usepackage{siunitx}
\usepackage{etoolbox} %
\robustify\bfseries

\usepackage{makecell}
\usepackage{adjustbox}
\usepackage{array}
\usepackage{multirow}
\usepackage{rotating}
\usepackage[english]{babel}

\usepackage{commath}

\definecolor{iccvblue}{rgb}{0.21,0.49,0.74}
\usepackage[breaklinks,colorlinks,allcolors=iccvblue]{hyperref}

\title{Unsupervised Joint Learning of Optical Flow and Intensity with Event Cameras} %

\author{Shuang Guo$^{1}$, Friedhelm Hamann$^{1,2}$ and Guillermo Gallego$^{1,2,3}$.\\
$^{1}$~TU Berlin and Robotics Institute Germany,\\
$^{2}$~Science of Intelligence Excellence Cluster, 
$^{3}$~Einstein Center for Digital Future.
}

\def\tref{t_\text{ref}} %
\def\pol{p} %

\def\cL{\mathcal{L}} %
\def\cE{\mathcal{E}} %
\def\cV{\mathcal{V}} %
\def\numEvents{N_e} %
\def\Warp{\mathbf{W}}
\def\WarpImg{\mathcal{W}}

\def\bx{\mathbf{x}}

\def\pol{p}
\def\flow{F}

\def\IWE{\text{IWE}}
\def\intensity{L}

\def\cN{\mathcal{N}} %
\newcommand{\bnum}[1]{\bfseries #1}

\newcommand{\novalue}{{\textendash}}

\definecolor{light-gray}{gray}{0.6}
\newcommand\gframe[1]{{\color{light-gray}\frame{#1}}}

\usepackage{pifont}
\newcommand{\yesmark}{\ding{51}}%
\newcommand{\nomark}{\ding{55}}%

\begin{document}

\maketitle

\definecolor{somegray}{gray}{0.5}
\newcommand{\darkgrayed}[1]{\textcolor{somegray}{#1}}
\begin{textblock}{11}(2.5, 0.6)
\begin{center}
\darkgrayed{This paper has been accepted for publication at the\\
IEEE International Conference on Computer Vision (ICCV), Honolulu, 2025.
\copyright IEEE}
\end{center}
\end{textblock}

\begin{abstract}
Event cameras rely on motion to obtain information about scene appearance. 
This means that appearance and motion are inherently linked: either both are present and recorded in the event data, or neither is captured.
Previous works treat the recovery of these two visual quantities as separate tasks, which does not fit with the above-mentioned nature of event cameras and overlooks the inherent relations between them. 
We propose an unsupervised learning framework that jointly estimates optical flow (motion) and image intensity (appearance) using a single network.
From the data generation model, we newly derive the event-based photometric error as a function of optical flow and image intensity. 
This error is further combined with the contrast maximization framework to form a comprehensive loss function that provides proper constraints for both flow and intensity estimation.
Exhaustive experiments show our method's state-of-the-art performance: 
in optical flow estimation, it reduces EPE by 20\% and AE by 25\% compared to unsupervised approaches, while delivering competitive intensity estimation results, particularly in high dynamic range scenarios.
Our method also achieves shorter inference time than all other optical flow methods and many of the image reconstruction methods, while they output only one quantity.
Project page: \href{https://github.com/tub-rip/e2fai}{https://github.com/tub-rip/E2FAI}
\end{abstract}
    
\section{Introduction}
\label{sec:intro}

\def\figWidth{0.24\linewidth}
\begin{figure}[t]
	\centering
    
    {\footnotesize
    \setlength{\tabcolsep}{1pt}
	\begin{tabular}{
	>{\centering\arraybackslash}m{\figWidth} 
	>{\centering\arraybackslash}m{\figWidth}
	>{\centering\arraybackslash}m{\figWidth}
	>{\centering\arraybackslash}m{\figWidth}
        }

        \multicolumn{4}{c}{\includegraphics[width=0.95\linewidth]{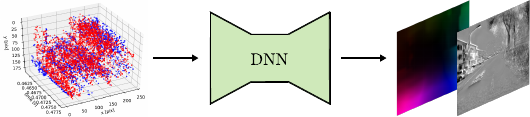}} \vspace{1ex}\\

	\gframe{\includegraphics[width=\linewidth]{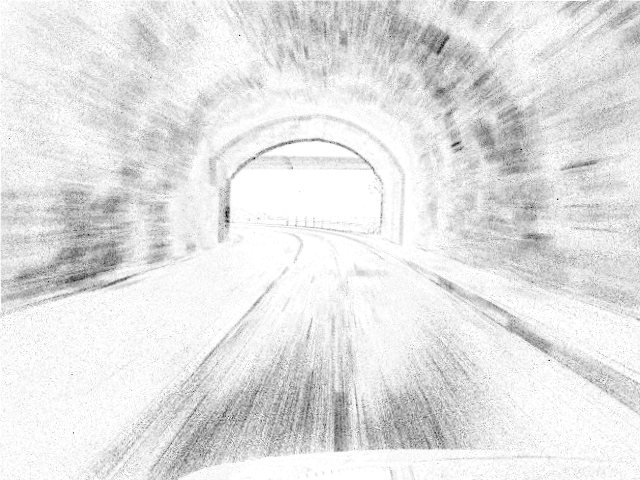}}
        &\includegraphics[width=\linewidth]{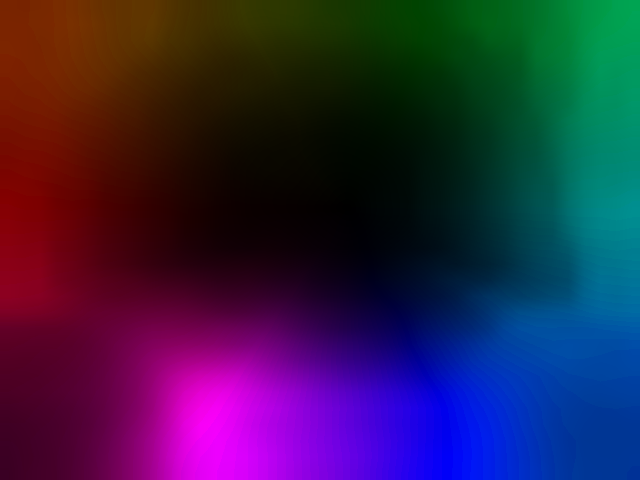}
	&\includegraphics[width=\linewidth]{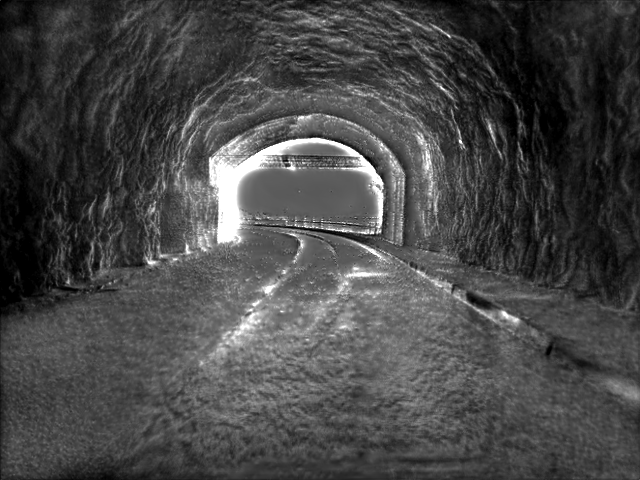}
        &\includegraphics[width=\linewidth]{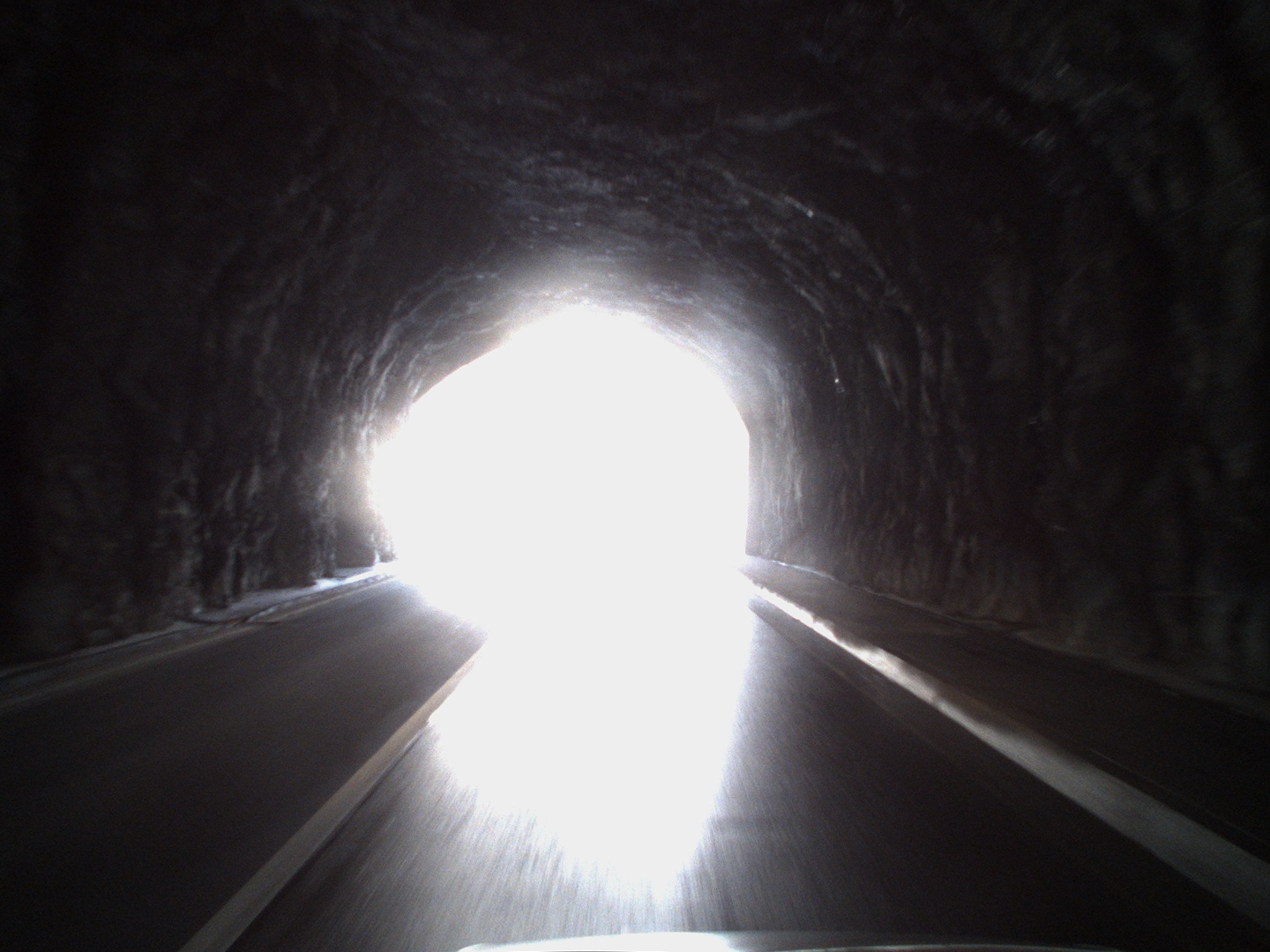}
	\\[-0.3ex]

        (a) Input Events
        & (b) Estimated Flow
        & (c) Estimated Intensity
        & (d) Image (reference)
        \\
	\end{tabular}
	}
    \caption{
    \emph{Our method} computes accurate optical flow and intensity images from event-camera data despite complex scenarios, fast motion and high dynamic range.
    The above result is obtained using data from interlaken\_00\_b sequence of DSEC \cite{Gehrig21threedv}.
    \label{fig:eye_catcher}
    }
\end{figure}

Event cameras \cite{Lichtsteiner08ssc,Posch14ieee} are novel bio-inspired vision sensors that offer attractive properties compared to traditional cameras: high temporal resolution, very high dynamic range (HDR), low power consumption and high pixel bandwidth, resulting in reduced motion blur. 
Hence, event cameras have a large potential for computer vision and robotics applications in challenging scenarios for traditional cameras, such as high speed motion and HDR illumination. 
However, novel methods are required to process the unconventional output of these sensors (a stream of asynchronous per-pixel brightness changes instead of conventional images) in order to unlock their potential \cite{Gallego20pami}.

In the past decade, a variety of computer vision algorithms have been developed to recover fundamental visual quantities from event streams
\cite{Gallego20pami,Zheng23arxiv,Chen20msp,Guo24eccv}, 
such as \mbox{EV-FlowNet} \cite{Zhu19cvpr} for optical flow estimation, 
and E2VID \cite{Rebecq19pami} for image intensity reconstruction.
Despite achieving good performance, most of these methods are designed to estimate a single visual quantity: optical flow \emph{or} image intensity. 
This philosophy does not fit well with the fact that, under constant illumination, motion and appearance are inherently entangled in the input data, since events are produced by \emph{moving intensity patterns} on the image plane (e.g., \cref{fig:eye_catcher}a).
More accurate and robust estimation becomes possible when the synergies between both visual quantities %
are properly leveraged.

To achieve this goal, we newly derive the event-based photometric error (PhE) as a function of optical flow and image intensity, and combine it with the state-of-the-art contrast maximization (CMax) framework \cite{Gallego18cvpr}, yielding a comprehensive loss function.
Both PhE and CMax provide constraints on scene appearance and motion. 
The former focuses more on appearance (i.e., intensity) while the latter focuses more on motion (i.e., optical flow).
Their complementary properties give rise to a well-behaved loss in both aspects.
In addition, our loss function also includes a full consideration of the internal synergies between estimated visual quantities:
during training, the predicted flow is leveraged to warp the predicted image intensity to the adjacent interval so that a temporal consistency (TC) loss can be calculated, which significantly encourages consistency and robustness.
We train a deep neural network (DNN) with this loss function in an unsupervised manner. 
The resulting model can predict precise optical flow and image intensity simultaneously from event data (see \cref{fig:eye_catcher}).

To the best of our knowledge, this paper presents the first unsupervised learning-based approach that jointly recovers optical flow and image intensity from event data, with a single network (\cref{tab:related}).
In the experiments, we evaluate our method in terms of optical flow and intensity on a variety of public datasets.
For optical flow, our method achieves the best accuracy in the unsupervised learning category in the DSEC benchmark \cite{Gehrig21threedv}, with 20\% and 25\% improvements in terms of EPE and AE, respectively.
For intensity, our method reports competitive results with respect to other unsupervised methods and even some supervised ones, especially in HDR scenarios.
In terms of speed, our method reports minimal inference time to obtain both visual quantities.
Our method robustly predicts precise flow and intensity on unseen data recorded in different scenarios and with different event cameras, thus demonstrating generalization.

Our contributions can be summarized as follows:

\begin{enumerate}
\item We propose the first unsupervised learning framework for the joint estimation, with a single network, of event-based optical flow and image intensity.
Its working principle fits naturally with the characteristics of event data.

\item We derive event-based PhE, and combine it with CMax, yielding a comprehensive and well-behaved loss function for the estimation of motion and scene appearance.
It can be directly adapted to various optimization and learning-based solutions for similar problems.

\item We conduct comprehensive experiments on public datasets, where our method shows state-of-the-art performance on optical flow, image intensity and inference time.
Furthermore, our method shows excellent generalization on unseen data recorded in different scenarios (HDR and fast motion) and with different event cameras.

\end{enumerate}

We hope the clear advantages of our approach and the source code provided will make its adoption appealing,  
thus bringing the tasks of optical flow estimation and intensity reconstruction closer than they currently are (until now, they have mostly been treated as independent problems).

\section{Related Work}
\label{sec:related}

\begin{table}[t]
\centering
\begin{adjustbox}{max width=\linewidth}
\setlength{\tabcolsep}{3pt}
\renewcommand{\arraystretch}{1.1}
\begin{tabular}{lccccll}
\toprule
\textbf{Method} & \textbf{Year} & \textbf{Type} & \textbf{DOFs} & \textbf{Joint} & \textbf{Loss function}\\
\midrule
IVM~\cite{Cook11ijcnn} & 2011 & MB & 3 & \yesmark & ``Consistency'' of the data (Max.~Likelihood) \\ 
SOFIE~\cite{Bardow16cvpr} & 2016 & MB & 6 & \yesmark & Pixel-wise brightness change + TC \\ 
E-cGAN~\cite{Mostafavi21ijcv} & 2021 & SL & 6 & \nomark & Supervised: Error w.r.t. ground truth \\
BTEB~\cite{Paredes21cvpr} & 2021 & USL & 6 & \nomark & FlowNet: IWE sharpness; ReconNet: LEGM \\
\textbf{Ours} & 2025 & USL & 6 & \yesmark & CMax + PhE of Flow and Intensity + TC \\ 
\bottomrule
\end{tabular}
\end{adjustbox}
\caption{
\emph{Event-based optical flow and intensity estimation methods.} 
The columns indicate: the method type,
the number of degrees of freedom (DOFs) of the camera motion that the method can handle (3$\equiv$rotation, 6$\equiv$free motion), whether the method estimates flow and intensity jointly or separately, and the loss used.
\label{tab:related}
}

\end{table}

Given the high-speed and HDR properties of event cameras, extensive research has been carried out to utilize them for the estimation of optical flow and image intensity. 
Let us review the literature on each of these tasks in \cref{sec:related:flow} and \cref{sec:related:intensity}, respectively, and summarize the approaches that solve for both quantities in \cref{sec:related:joint}.

\subsection{Event-based Optical Flow Estimation}
\label{sec:related:flow}

Previous works can be categorized into model-based (MB), supervised learning (SL) or unsupervised learning (USL).
State-of-the-art MB methods~\cite{Shiba24pami,Brebion21tits} formulate the problem using an objective function, and solve for optical flow through optimization.
The most commonly-used objective is the sharpness of images of warped events (IWEs), such as the CMax loss~\cite{Gallego18cvpr,Gallego19cvpr} used in \cite{Shiba24pami}, and the flow warping loss (FWL) \cite{Stoffregen20eccv} used in \cite{Brebion21tits}.
The optimization usually takes a number of iterations to converge, and it repeatedly warps all the involved events in every iteration, which makes these methods costly. 
Furthermore, the objectives derived from IWE sharpness may suffer from the issue of event collapse \cite{Shiba22sensors,Shiba22aisy}, which affects flow accuracy and requires strong regularization to mitigate it.

Due to the better fitting capabilities and short inference times of DNNs, researchers have adopted them to compute optical flow.
SL approaches \cite{Gehrig19iccv,Gehrig21threedv,Liu23iccv,Li23iros,Luo23iccv,Wu24icra} train DNNs to learn the mapping from event data to the ground truth (GT) optical flow.
However, the acquisition of GT relies on either simulation \cite{Li23iros,Luo23iccv} or calculation from external depth sensors using the motion field equation \cite{Zhu18rss,Gehrig21threedv}.
Both are expensive. The former suffers from the sim-to-real gap, that is, the trained model may not perform well on real-world data,
while the latter suffers from sensor inaccuracy and data sparsity \cite{Shiba24pami,Hamann24eccv}.
Although SL methods are generally ahead in optical flow benchmarks, the above issues about GT data are inherited by the trained model. 

In contrast, USL methods \cite{Paredes23iccv,Zhu19cvpr,Hamann24eccv,You24arxiv} forego costly data labeling, shifting the focus to the data and the loss function to learn optical flow patterns.
Similarly to MB methods, current mainstream loss functions are still based on IWE sharpness \cite{Gallego19cvpr}.
For instance, \mbox{EV-FlowNet} \cite{Zhu19cvpr} quantifies sharpness in terms of average timestamp images \cite{Mitrokhin18iros}.
More recently, \cite{Shiba24pami,Paredes23iccv,Hamann24eccv} adopted the CMax loss to train DNNs, which achieved obvious improvements with respect to MB methods that used the same loss functions.
In addition, USL performs expensive event warping only during training, hence it has much shorter and more constant inference time than equivalent MB methods.

\subsection{Event-based Image Intensity Reconstruction}
\label{sec:related:intensity}
Similarly, intensity estimation methods can be classified into MB, SL and USL categories.
For the MB category, early works recovered brightness from events by means of temporal filtering \cite{Scheerlinck18accv} or temporal integration with manifold denoising \cite{Munda18ijcv}.
A recent work \cite{Zhang22pami} formulated this task as a linear inverse problem with image regularization and solved for intensity using 
ADMM~\cite{Boyd2011admm}.
However, it required accurate optical flow as input, which limits its applicability.

SL approaches \cite{Rebecq19pami,Scheerlinck20wacv,Gantier21tip,Weng21cvpr} trained DNNs using synthetic data to learn the mapping from event data to image intensity.
In this way, the models were prevented from learning motion blur and under/over-exposure that happens in traditional cameras.
Besides, these methods were aimed at video reconstruction and therefore adopted recurrent network architectures while introducing temporal consistency loss to guarantee the continuity of sequential images.
However, they suffer from the sim-to-real gap and sometimes perform poorly in HDR scenarios (see, e.g., \cref{fig:image_array}).

USL methods for intensity reconstruction are underexplored.
Recent work \cite{Paredes21cvpr} proposed to supervise the training with event-based photometric consistency. 
The loss was defined using the linearized event generation model (LEGM), as the error between brightness increments measured by event data and those obtained from the spatial gradient of the predicted image.
However, this method requires accurate optical flow provided externally for both training and inference, for which the authors had to train a separate network to estimate optical flow (see details of the flow part in \cref{sec:related:joint}).
The need to infer sequentially with two separate models takes more time and amplifies error propagation, thus significantly limiting practicality.

\subsection{Event-based Flow-Intensity Estimation}
\label{sec:related:joint}

\Cref{tab:related} summarizes the main works that estimate both optical flow and image intensity with event cameras.
The earliest work \cite{Cook11ijcnn} proposed a joint estimation approach recovering optical flow, image intensity and camera ego-motion.
However, it was limited to pure rotational motion.
Bardow et al. \cite{Bardow16cvpr} developed a variational algorithm that simultaneously optimized optical flow and image intensity. 
However, optical flow was only constrained by the brightness constancy between estimated intensity images, while the motion information in event data was not utilized.
Hence, the resulting intensity and optical flow showed poor accuracy.

In the learning-based category, \cite{Mostafavi21ijcv} presented an SL method for image reconstruction from events using conditional generative adversarial networks (cGANs), and also showed the applicability to predict depth and optical flow. 
The networks were trained and performed inference for each quantity separately, so the bonds between them were ignored.
More recently, \cite{Paredes21cvpr} proposed training an intensity reconstruction DNN (same architecture as \cite{Rebecq19pami}), but in an unsupervised manner. 
To do so, they separately trained a network for optical flow estimation using the same loss as \cite{Zhu19cvpr}, and subsequently fed its output flow into a second network for intensity reconstruction (see \cref{sec:related:intensity}). 

Our method overcomes and simplifies previous designs: it jointly estimates intensity and flow in an unsupervised manner by leveraging their synergies via the proposed loss function and architecture (i.e., using a single DNN).

\section{Method}
\label{sec:method}
In this section, we first explain the preliminaries (\cref{sec:method:pre}), then introduce our proposed comprehensive loss function, including newly derived event-based PhE (\cref{sec:method:phe}), CMax loss (\cref{sec:method:cmax}) and regularization terms (\cref{sec:method:reg}). 
Finally, we present the training pipeline in \cref{sec:method:train}.

\subsection{Preliminaries}
\label{sec:method:pre}
\textbf{Event Generation Model (EGM)}. 
Every pixel of an event camera independently measures brightness changes, 
generating an event $e_k \doteq (\bx_k,t_k,\pol_k)$ when the logarithmic brightness change $\Delta L$ reaches a contrast threshold $C$ \cite{Gallego20pami}:
\begin{equation}
    \Delta L \doteq L(\bx_k,t_k) - L(\bx_k,t_k - \Delta t_k) = \pol_k C,
    \label{eq:EGM}
\end{equation}
where $\pol_k \in \left\{ +1,-1 \right\}$ indicates the polarity of the intensity increment, and $\Delta t_k$ is the time elapsed since the previous event at the same pixel $\bx_k$.

\noindent\textbf{Event warping}.  
Given a set of events $\cE \doteq \left\{e_k\right\}^{\numEvents}_{k=1}$, we can warp them to a reference time $\tref$ with a motion model $\Warp$, yielding a set of warped events $\cE'_{\tref} \doteq \left\{e'_k\right\}^{\numEvents}_{k=1}$, where:
$
    e_k \doteq (\bx_k, t_k, \pol_k) \rightarrow e'_k \doteq (\bx'_k, \tref, \pol_k).
$
Provided that the time span of $\cE$ is small, we can assume all pixels move with constant but different velocities on the image plane, that is, the motion model $\Warp$ is given by:
\begin{equation}
    \bx'_k = \bx_k + (\tref - t_k) \flow(\bx_k),
    \label{eq:flow_warp}
\end{equation}
where $\flow(\bx_k)$ is the optical flow at $\bx_k$ \cite{Shiba24pami}.

\begin{figure*}[t]
\centering
\includegraphics[width=.84\linewidth]{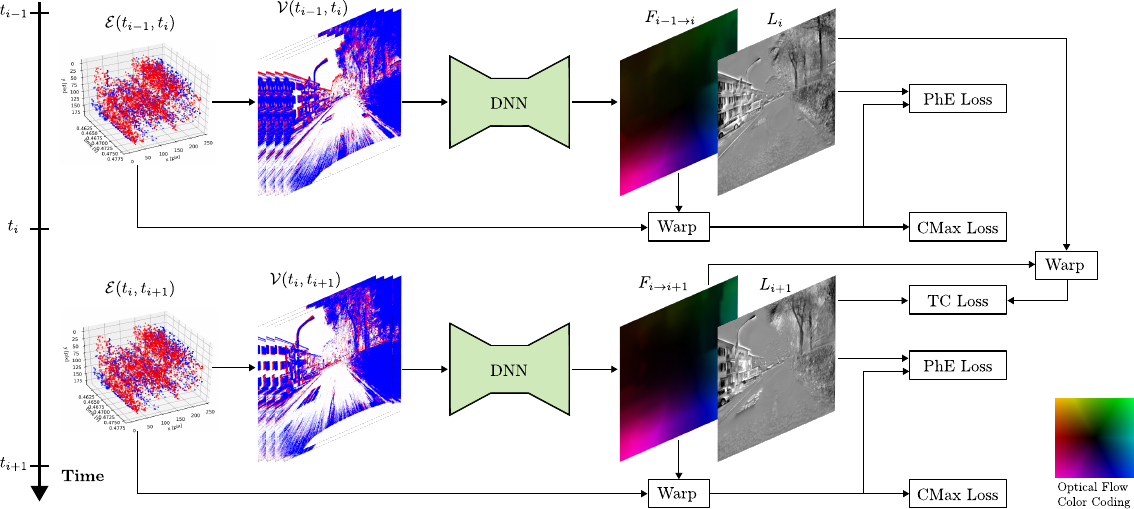}
\caption{
\emph{The training pipeline of our proposed flow-intensity joint estimation method.} 
In every training step, two consecutive event data samples are input to the network, respectively. 
The CMax and PhE losses for each sample are calculated with the output optical flow and image intensity. 
The TC loss is defined by the photometric error between the predicted $\intensity_{i+1}$ and the one warped from $\intensity_i$ through $\flow_{i \rightarrow i+1}$.
}
\label{fig:pipeline}
\end{figure*}

\subsection{Event-based Photometric Error (PhE)}
\label{sec:method:phe}
We leverage the original EGM to derive the event-based PhE.
Starting from \eqref{eq:EGM}, we warp an event $e_k \doteq (\bx_k,t_k,\pol_k)$ and its predecessor (at the same pixel) $e_{k-1} \doteq (\bx_k,t_k - \Delta t_k,\pol_{k-1})$ to a reference time $\tref$.
According to \eqref{eq:flow_warp}, their warped locations on the camera plane at $\tref$ are:
\begin{equation}
    \begin{aligned}
    \bx'_k &= \bx_k + (\tref - t_k) \flow(\bx_k), \\
    \bx'_{k-1} &= \bx_k + (\tref - (t_k - \Delta t_k)) \flow(\bx_k).
    \label{eq:phe_warp}
    \end{aligned}
\end{equation}

Let the image intensity at $\tref$ be\footnote{We omit ``logarithmic'' in the following text for simplification.} $\intensity(\bx)$, then we substitute \eqref{eq:phe_warp} into \eqref{eq:EGM} to obtain the PhE for the event $e_k$:
\begin{equation}
    \epsilon_k \doteq (\intensity(\bx'_k) - \intensity(\bx'_{k-1})) - \pol_k C,
    \label{eq:phe_term}
\end{equation}
which is the difference between the predicted intensity increment and the measured one.
In practice, $\bx'_k$ and $\bx'_{k-1}$ have sub-pixel precision, so we use bilinear interpolation to compute $\intensity(\bx'_k)$ and $\intensity(\bx'_{k-1})$.
Therefore, every PhE term \eqref{eq:phe_term} provides constraints for the intensity values at up to eight pixels around $\bx'_k$ and $\bx'_{k-1}$, as well as the optical flow value at one pixel $\bx_k$.
Finally, we sum the photometric error terms of all events in $\cE$, obtaining the PhE loss:
\begin{equation}
    \cL_\text{PhE} (\intensity, \flow) \doteq \frac{1}{\numEvents} \sum_{k=1}^{\numEvents} \,\abs{\epsilon_k}.
    \label{eq:phe_loss}
\end{equation}

It is an objective function of \emph{both} intensity and flow, which opens the door to the joint estimation of both quantities.
It is worth emphasizing that the PhE loss does not have the event collapse problem \cite{Shiba22sensors,Shiba22aisy} that CMax suffers from.

\subsection{Contrast Maximization (CMax)}
\label{sec:method:cmax}
The CMax framework \cite{Gallego18cvpr,Gallego19cvpr} assumes events are triggered by moving edges, so that optical flow estimation can be determined by seeking the best motion compensation.
Specifically, warped events $\cE'_{\tref}$ are aggregated on the image plane at $\tref$, forming an image of warped events:
\begin{equation}
    \IWE(\bx; \cE'_{\tref}, \flow) \doteq \sum_{k=1}^{\numEvents} \cN\bigl(\bx;\, \bx'_k,\, \sigma^2=1\text{px}\bigr),
\end{equation}
where every pixel $\bx$ counts its number of warped events.

In this work, we adopt the inverse of the $L^1$ magnitude of the gradient to quantify the CMax loss:
\begin{equation}
    \cL_\text{CMax} (\flow) \doteq 1 \bigg/ \left(\frac{1}{\abs{\Omega}} \int_{\Omega} \norm{\nabla \IWE(\bx)}_1 \,d\bx\right).
    \label{eq:cmax_loss}
\end{equation}
In contrast to PhE, the CMax loss recovers scene appearance in the form of a sharp edge map, which is the objective instead of the variable.
The only optimizable variable is optical flow, which reflects that CMax loss focuses more on motion parameters, as mentioned in \cref{sec:intro}.

\subsection{Regularization}
\label{sec:method:reg}

\textbf{Smoothness of Flow and Intensity}.
\label{sec:method:reg:tv}
As stated in \cref{sec:method:phe}, the PhE loss only provides supervisory constraints on pixels that contain events (called valid pixels).
To infer the values of the remaining pixels, regularization is needed; 
we use the total variation (TV) \cite{Rudin92physica} to encourage smoothness of optical flow and intensity predictions:
$\cL_\text{FTV}(\flow) \doteq \frac{1}{\abs{\Omega}} \int_{\Omega} \|\nabla \flow(\bx)\|_1 \,d\bx,$ and 
$\cL_\text{ITV}(\intensity) \doteq \frac{1}{\abs{\Omega}} \int_{\Omega} \|\nabla \intensity(\bx)\|_1 \,d\bx.$
Additionally, the smoothness of the flow mitigates the event collapse caused by the CMax loss.

\noindent\textbf{Temporal Consistency (TC)}.
\label{sec:method:reg:tc}
A key advantage of joint estimation (over a separate one) is being able to leverage the synergies between the quantities for better consistency.
We achieve this by establishing associations between temporally consecutive predictions.
As depicted in \cref{fig:pipeline}, in each training step, the network takes as input two consecutive data samples to predict the corresponding optical flow and image intensity.
We leverage flow $\flow_{i \rightarrow i+1}$ to transport $\intensity_i$ to $t_{i+1}$, yielding $\intensity'_{i+1}$, whose photometric error with respect to $\intensity_{i+1}$ (predicted from the other data sample) is calculated to encourage temporal consistency (TC):
\begin{equation}
    \cL_\text{TC} \doteq \frac{1}{\abs{\Omega}} \int_{\Omega} \abs{\intensity_{i+1}(\bx) - \WarpImg(\bx; \intensity_i, \flow_{i \rightarrow i+1})} \,d\bx,
    \label{eq:tc_loss}
\end{equation}
where $\WarpImg$ warps an image according to the optical flow.
The introduction of the TC loss greatly improves the prediction quality, especially in image intensity (see \cref{sec:experim:ablation}).

\subsection{Training Pipeline}
\label{sec:method:train}
An overview of the training pipeline is displayed in \cref{fig:pipeline}.
During training, the loss terms \eqref{eq:phe_loss}, \eqref{eq:cmax_loss}, etc. %
are evaluated with the predictions of each sample, while the TC loss \eqref{eq:tc_loss}, between consecutive samples.
Consequently, the total loss is the weighted sum of all these terms:
\begin{equation*}
\cL_\text{total} = \lambda_1 \cL_\text{PhE} + \lambda_2 \cL_\text{CMax} + \lambda_3 \cL_\text{FTV} + \lambda_4 \cL_\text{ITV} + \lambda_5 \cL_\text{TC},
\end{equation*}
where the first four terms are the sum for the two data samples (\cref{fig:pipeline}).
For every sample, the reference time of CMax is set to a random number within the time span of the event set \cite{Hamann24eccv}, that is, keeping the IWE sharp at any time, which helps reduce the possibility of event collapse,
while that of PhE is always set to the end time of the sample.
For inference, users just need to input one event voxel grid $\cV(t_{i-1}, t_i)$ to predict flow $\flow_{i-1 \rightarrow i}$ and intensity $\intensity_i$.

\section{Experiments}
\label{sec:experim}
We begin by describing the experimental setup (\cref{sec:experim:setup}), then present the results of optical flow evaluation (\cref{sec:experim:flow}) and of image intensity evaluation (\cref{sec:experim:intensity}), before finally introducing the ablation study (\cref{sec:experim:ablation}).

\subsection{Experimental setup}
\label{sec:experim:setup}

\textbf{Datasets}. 
\Cref{tab:datasets} summarizes the datasets used in our experiments.
The ECD dataset \cite{Mueggler17ijrr} is collected with a handheld DAVIS240C camera in indoor scenarios, where the motion speed varies from slow to fast.
The DSEC dataset \cite{Gehrig21threedv} features urban and highway driving scenarios in daylight and night. 
It is recorded with an on-board Prophesee Gen3 camera, and the GT optical flow is computed from the LiDAR disparity data.
The HDR dataset \cite{Rebecq19pami} contains event data captured in HDR illumination conditions, such as the Sun and a car driving out of a tunnel.
It highlights the HDR property of event cameras.
The BS-ERGB dataset \cite{Tulyakov22cvpr} provides high-resolution event data recorded by a handheld co-capture system composed of a Prophesee Gen4 camera (1 megapixel) and a FLIR RGB camera, in complex outdoor scenes.
The high quality of the GT frames makes them suitable for the evaluation of event-based intensity reconstruction.
For intensity evaluation, we use the EVREAL tool \cite{Ercan23cvprw}, and adopt its sequence selection.
The sole change made is the removal of the last seconds of the may29\_rooftop sequences from the evaluation, where the camera does not move and event data consists of pure noise.

\textbf{Metrics}.
For optical flow, we adopt standard metrics: end-point error (EPE), angular error (AE) and \%Out (the percentage of flow estimates whose EPE$>$3px).
We also report the flow warp loss (FWL) proposed in \cite{Stoffregen20eccv}.
For image intensity, we report full-reference metrics when GT images are available (e.g., BS-ERGB dataset): 
mean square error (MSE), structure similarity index (SSIM) \cite{Wang04tip} and perceptual similarity (LPIPS) \cite{Zhang18cvprLPIPS}.
Otherwise, we use no-reference metrics: BRISQUE~\cite{Mittal12tip}, NIQE~\cite{Mittal13spl}, MANIQA~\cite{Yang22cvprw} (e.g., HDR dataset).

\textbf{Implementation Details}.
We train our model only on the DSEC training split for 130 epochs with an AdamW optimizer \cite{Loshchilov2019iclr}, whose learning rate is $10^{-3}$ for the first 100 epochs and then decays to $10^{-4}$.
The training is carried out on four NVIDIA RTX A6000 GPUs with a total batch size of 24. 
This trained model is used for the evaluation of optical flow on the test split of DSEC and those of image intensity on BS-ERGB and HDR datasets, \emph{without fine-tuning}. 
The weights of the loss terms are: $\lambda_1 = 30, \lambda_2 = 1, \lambda_3 = 10, \lambda_4 = 0.001, \lambda_5 = 1$.
The contrast threshold $C$ is set to $0.2$.
We adopt the classical U-Net architecture \cite{Ronneberger15icmicci}, which has 15 input channels (number of time bins in the event voxel grid \cite{Zhu19cvpr}) and three output channels (first two: optical flow, third one: intensity).

During training and inference, we first apply average pooling with a kernel size of 16 to the raw output flow, and then perform bilinear interpolation to recover the original resolution for the output flow, to further guarantee flow smoothness.
For inference, like \cite{Paredes21cvpr}, we convert the predicted logarithmic intensity into the linear scale through $I = \exp{(\intensity)}$, then perform the robust min/max normalization before evaluation and visualization: $\hat{I} = (I - m)/(M - m)$, where $m$ and $M$ are the 1\% and 99\% percentiles of $I$, respectively.
The inference time reported in \cref{tab:exp:dsec,tab:intensity_inf_time} is measured on a single GPU of the same type (NVIDIA RTX A6000).

\begin{table}[t]
\centering
    \begin{adjustbox}{max width=\linewidth}
    \setlength{\tabcolsep}{4pt}
    \begin{tabular}{@{}lllll@{}}
    \toprule
    \textbf{Dataset} & \textbf{Camera} & \textbf{Pixels} & \textbf{Scenarios \& Features} \\
    \midrule
    ECD \cite{Mueggler17ijrr} & DAVIS240C & $240 \times 180$ & Indoor handheld, varying speed \\
    DSEC \cite{Gehrig21ral}  & Prophesee Gen3 & $640 \times 480$ & Outdoor driving, daylight \& night \\
    HDR \cite{Rebecq19pami}  & Samsung DVS Gen3 & $640 \times 480$ & \makecell[tl]{Indoor \& outdoor, handheld\\ \& driving, HDR} \\
    BS-ERGB \cite{Tulyakov22cvpr}  & PSEE Gen4 \& FLIR  & $970 \times 625$ & Outdoor handheld motion, HDR \\
    \bottomrule
    \end{tabular}
    \end{adjustbox}
    \caption{\label{tab:datasets} 
    \emph{Configurations of the datasets used in the experiments}. 
    The BS-ERGB data is cropped; its resolution is a bit smaller than the event camera resolution.}
\end{table}

\subsection{Optical Flow Evaluation}
\label{sec:experim:flow}

\begin{table*}[t!]
\centering
\adjustbox{max width=\linewidth}{%
\setlength{\tabcolsep}{3pt}

\begin{tabular}{ll*{17}{S[table-format=2.3]}}
\toprule
  &  &
  & \multicolumn{4}{c}{\textbf{All}}
  & \multicolumn{4}{c}{interlaken\_00\_b}
  & \multicolumn{4}{c}{interlaken\_01\_a}
  & \multicolumn{4}{c}{thun\_01\_a} \\
 \cmidrule(l{1mm}r{1mm}){4-7}
 \cmidrule(l{1mm}r{1mm}){8-11}
 \cmidrule(l{1mm}r{1mm}){12-15}
 \cmidrule(l{1mm}r{1mm}){16-19}

 Type & Method & \text{$t_\text{inf} [\text{ms}]$}
 & \text{EPE $\downarrow$} & \text{AE $\downarrow$} & \text{\%Out $\downarrow$} & \text{FWL $\uparrow$}
 & \text{EPE $\downarrow$} & \text{AE $\downarrow$} & \text{\%Out $\downarrow$} & \text{FWL $\uparrow$}
 & \text{EPE $\downarrow$} & \text{AE $\downarrow$} & \text{\%Out $\downarrow$} & \text{FWL $\uparrow$}
 & \text{EPE $\downarrow$} & \text{AE $\downarrow$} & \text{\%Out $\downarrow$} & \text{FWL $\uparrow$} \\
 
\midrule
\multirow{2}{*}{\shortstack{SL}}

 & E-RAFT~\cite{Gehrig21threedv}
 & 46.331
 & 0.788 & 10.56 & 2.684 & 1.286231422619447
 & 1.394 & 6.22 & 6.189 & 1.3233361693404235
 & 0.899 & 6.881 & 3.907 & 1.4233908392179435
 & 0.654 & 9.748 & 1.87 & 1.200584411 \\

 & IDNet~\cite{Wu24icra}
 &
 & \bnum{0.719} & \bnum{2.723} & \bnum{2.036} & \novalue 
 & \bnum{1.25} & \bnum{2.11} & \bnum{4.353} & \novalue
 & \bnum{0.774} & \bnum{2.25} & \bnum{2.596} & \novalue
 & \bnum{0.572} & \bnum{2.662} & \bnum{1.472} & \novalue \\

\midrule
\multirow{7}{*}{\shortstack{MB/\\USL}}

 & RTEF~\cite{Brebion21tits}
 &
 & 4.88 & \novalue & 41.95 & \bnum{2.51}
 & 8.59 & \novalue & 59.84 & \bnum{2.89}
 & 5.94 & \novalue & 47.33 & \bnum{2.92}
 & 3.01 & \novalue & 29.70 & \bnum{2.39}
 \\

 & MultiCM~\cite{Shiba22eccv}
 & \text{$9.9 \cdot 10^{3}$}
 & 3.472 & 13.983 & 30.855 & 1.365150407655252
 & 5.744 & 9.188 & 38.925 & 1.4992904511327119
 & 3.743 & 9.771 & 31.366 & 1.5137225860194
 & 2.116 & 11.057 & 17.684 & 1.2420968732859203 \\ 

 & BTEB~\cite{Paredes21cvpr}
 & 
 & 3.86 & \novalue & 31.45 & 1.30
 & 6.32 & \novalue & 47.95 & 1.46
 & 4.91 & \novalue & 36.07 & 1.42
 & 2.33 & \novalue & 20.92 & 1.32 \\

 & Paredes et al.~\cite{Paredes23iccv}
 & {40.1}
 & 2.33 & 10.56 & 17.771 & \novalue
 & 3.337 & 6.22 & 25.724 & \novalue
 & 2.489 & 6.881 & 19.153 & \novalue
 & 1.73 & 9.748 & 10.386 & \novalue \\

 & EV-FlowNet~\cite{Zhu19cvpr}
 &
 & 3.86 & \novalue & 31.45 & 1.30
 & 6.32  & \novalue & 47.95 & 1.46 
 & 4.91 & \novalue & 36.07 & 1.42
 & 2.33 & \novalue & 20.92 & 1.32  \\

 & MotionPriorCM~\cite{Hamann24eccv}
 & 17.856
 & 3.2  & 8.53 & 15.21 & 1.4603124856948853
 & 3.21 & 4.89 & \bnum{20.45} & 1.582527995109558
 & 2.38 & 5.46 & 17.4  & 1.701579213142395
 & 1.39 & 6.99 & 7.36  & 1.297378420829773 \\

 & VSA-SM~\cite{You24arxiv}
 & 
 & 2.22 & 8.859 & 16.825 & \novalue 
 & 3.204 & 6.232 & 24.611 & \novalue 
 & 2.462 & 6.997 & 20.228 & \novalue 
 & 1.553 & 6.626 & 10.667 & \novalue \\

 & \textbf{Ours}
 & \bnum{15.115}
 & \bnum{1.781} & \bnum{6.439} & \bnum{11.241} & 1.788
 & \bnum{3.079} & \bnum{3.871} & 20.758 & 1.923
 & \bnum{1.899} & \bnum{4.107} & \bnum{12.62} & 2.058
 & \bnum{1.256} & \bnum{5.685} & \bnum{6.612} & 1.561 \\

 \\[-0.5ex]
 
\midrule
  & & & \multicolumn{4}{c}{thun\_01\_b}
  & \multicolumn{4}{c}{zurich\_city\_12\_a}
  & \multicolumn{4}{c}{zurich\_city\_14\_c}
  & \multicolumn{4}{c}{zurich\_city\_15\_a} \\

 \cmidrule(l{1mm}r{1mm}){4-7}
 \cmidrule(l{1mm}r{1mm}){8-11}
 \cmidrule(l{1mm}r{1mm}){12-15}
 \cmidrule(l{1mm}r{1mm}){16-19}
 
 Type & Method & & \text{EPE $\downarrow$} & \text{AE $\downarrow$} & \text{\%Out $\downarrow$} & \text{FWL $\uparrow$}
 & \text{EPE $\downarrow$} & \text{AE $\downarrow$} & \text{\%Out $\downarrow$} & \text{FWL $\uparrow$}
 & \text{EPE $\downarrow$} & \text{AE $\downarrow$} & \text{\%Out $\downarrow$} & \text{FWL $\uparrow$}
 & \text{EPE $\downarrow$} & \text{AE $\downarrow$} & \text{\%Out $\downarrow$} & \text{FWL $\uparrow$} \\
 
\midrule
\multirow{2}{*}{\shortstack{SL}}

 & E-RAFT~\cite{Gehrig21threedv}
 & & 0.578 & 8.409 & 1.518 & 1.1767931182449056
 & 0.612 & 23.164 & \bnum{1.057} & 1.1161122798926002
 & \bnum{0.713} & 10.226 & \bnum{1.913} & 1.4688118
 & 0.589 & 8.878 & 1.303 & 1.335840906676725 \\

 & IDNet~\cite{Wu24icra}
 & & \bnum{0.546} & \bnum{2.07} & \bnum{1.35} & \novalue 
 & \bnum{0.603} & \bnum{4.556} & 1.161 & \novalue
 & 0.76 & \bnum{3.742} & 2.735 & \novalue
 & \bnum{0.548} & \bnum{2.545} & \bnum{1.022} & \novalue \\
 
 \midrule
 \multirow{7}{*}{\shortstack{MB/\\USL}}

 & RTEF~\cite{Brebion21tits}
 & & 3.91 & \novalue & 34.69 & \bnum{2.48}
 & 3.14 & \novalue & 34.08 & \bnum{1.42}
 & 4.00 & \novalue & 45.67 & \bnum{2.67}
 & 3.78 & \novalue & 37.99 & \bnum{2.82} \\

 & MultiCM~\cite{Shiba22eccv}
 & & 2.48 & 12.045 & 23.564 & 1.24194368901993
 & 3.862 & 28.613 & 43.961 & 1.1375111350019202
 & 2.724 & 12.624 & 30.53 & 1.4985924122973489
 & 2.347 & 11.815 & 20.987 & 1.4119389239425226 \\ 

 & BTEB~\cite{Paredes21cvpr}
 & 
 & 3.04 & \novalue & 25.41 & 1.33
 & 2.62 & \novalue & 25.80 & 1.03
 & 3.36 & \novalue & 36.34 & 1.24
 & 2.97 & \novalue & 25.53 & 1.33 \\

 & Paredes et al.~\cite{Paredes23iccv}
 & & 1.657 & 8.409 & 9.343 & \novalue
 & 2.724 & 23.164 & 26.649 & \novalue
 & 2.635 & 10.226 & 23.005 & \novalue
 & 1.686 & 8.878 & 9.977 & \novalue \\

 & EV-FlowNet~\cite{Zhu19cvpr}
 & & 3.04 & \novalue & 25.41 & 1.33
 & 2.62 & \novalue & 25.80 & 1.03
 & 3.36 & \novalue & 36.34 & 1.24
 & 2.97 & \novalue & 25.53 & 1.33 \\

 & MotionPriorCM~\cite{Hamann24eccv}
 & & 1.54 & 6.55 & 9.69 & 1.3275487422943115
 & 8.33 & 20.16 & 22.39 & 1.1288989782333374
 & 1.78 & 8.79 & 12.99 & 1.5562018156051636
 & 1.45 & 6.27 & 8.34 & 1.5112605094909668 \\

 & VSA-SM~\cite{You24arxiv}
 & 
 & 1.744 & 6.763 & 13.066 & \novalue 
 & 2.193 & 17.128 & 15.236 & \novalue 
 & 1.687 & 7.565 & 11.022 & \novalue 
 & 1.851 & 8.057 & 13.553 & \novalue \\

 & \textbf{Ours}
 & 
 & \bnum{1.147} & \bnum{4.89} & \bnum{5.809} & 1.625
 & \bnum{1.924} & \bnum{14.345} & \bnum{13.309} & 1.396
 & \bnum{1.504} & \bnum{6.931} & \bnum{10.512} & 1.923
 & \bnum{1.263} & \bnum{5.461} & \bnum{6.411} & 1.888 \\

\bottomrule
\end{tabular}
}
\caption{\label{tab:exp:dsec}
\emph{Optical flow evaluation}. 
Results on the DSEC optical flow benchmark \cite{Gehrig21threedv}.
Bold is the best in each category, except the column of $t_\text{inf}$, where only the shortest $t_\text{inf}$ is marked. 
Note that our model predicts both flow and intensity, while others only predicts the former.}
\vspace{0.5ex}
\end{table*}

\def\figWidth{0.16\linewidth}
\begin{figure*}[t]
	\centering
    {\footnotesize
    \setlength{\tabcolsep}{1pt}
	\begin{tabular}{
	>{\centering\arraybackslash}m{0.25cm} 
	>{\centering\arraybackslash}m{\figWidth}
	>{\centering\arraybackslash}m{\figWidth}
	>{\centering\arraybackslash}m{\figWidth}
	>{\centering\arraybackslash}m{\figWidth}
        >{\centering\arraybackslash}m{\figWidth}
        >{\centering\arraybackslash}m{\figWidth}
        }

    \rotatebox{90}{\makecell{zurich\_city\_14\_c}}
	&\gframe{\includegraphics[width=\linewidth]{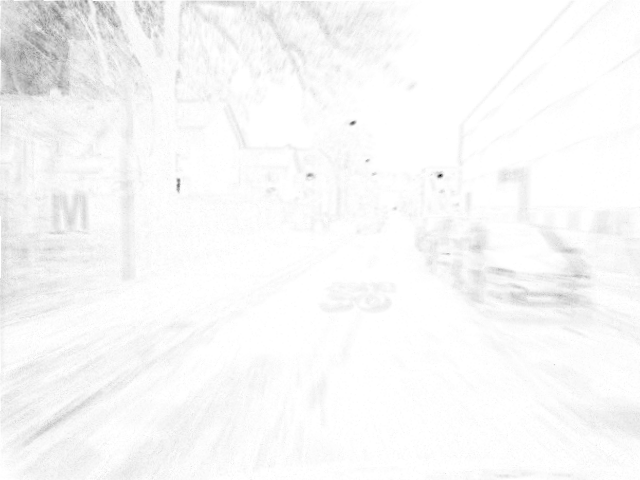}}
	&\gframe{\includegraphics[width=\linewidth]{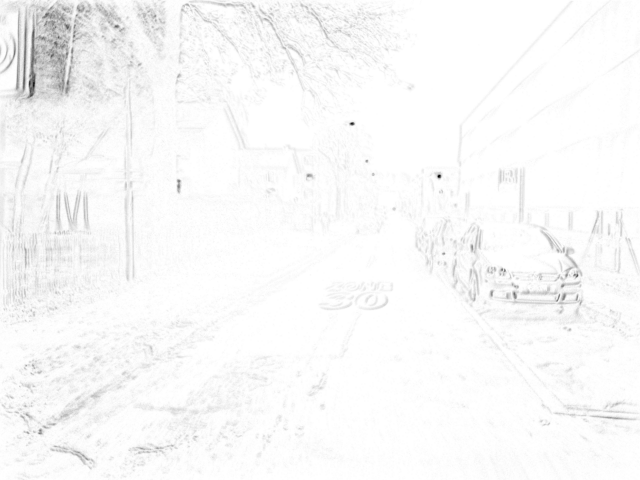}}
        &\includegraphics[width=\linewidth]{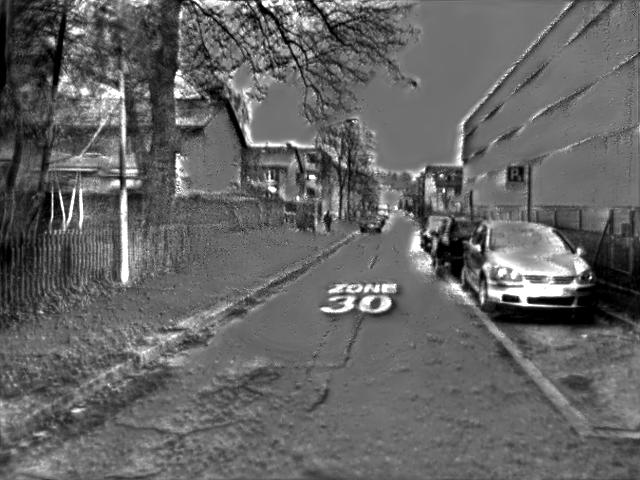}
	&\includegraphics[width=\linewidth]{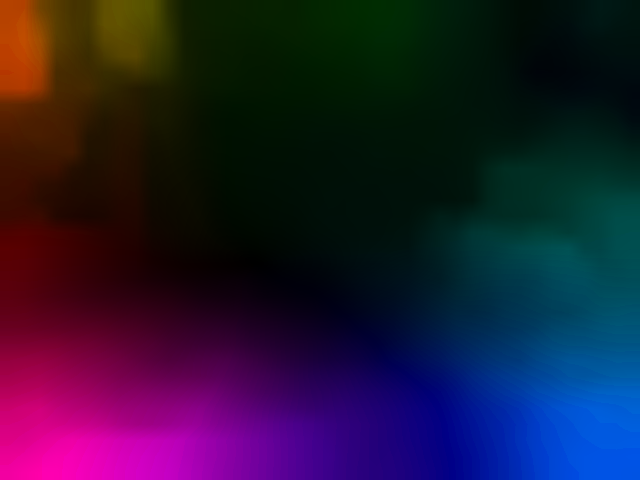}
        &\includegraphics[width=\linewidth]{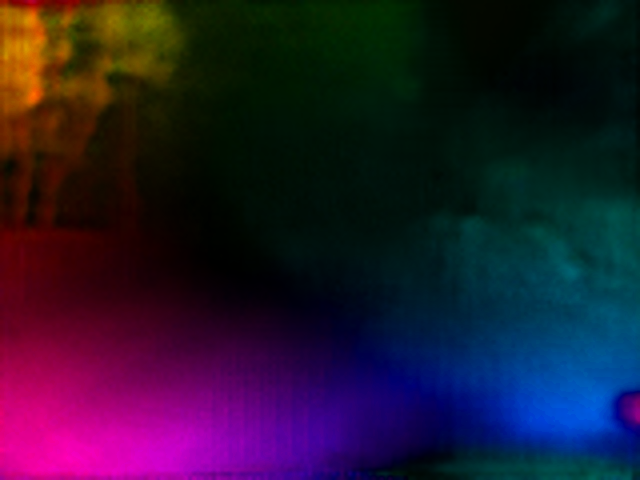}
        &\includegraphics[width=\linewidth]{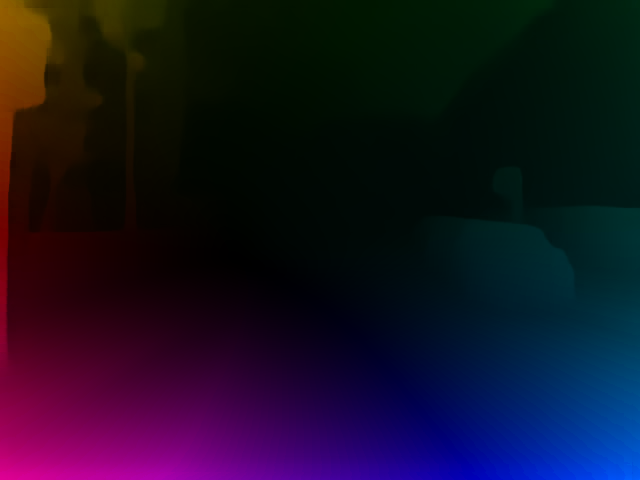}
	\\[-0.2ex]

    \rotatebox{90}{\makecell{thun\_01\_b}}
	&\gframe{\includegraphics[width=\linewidth]{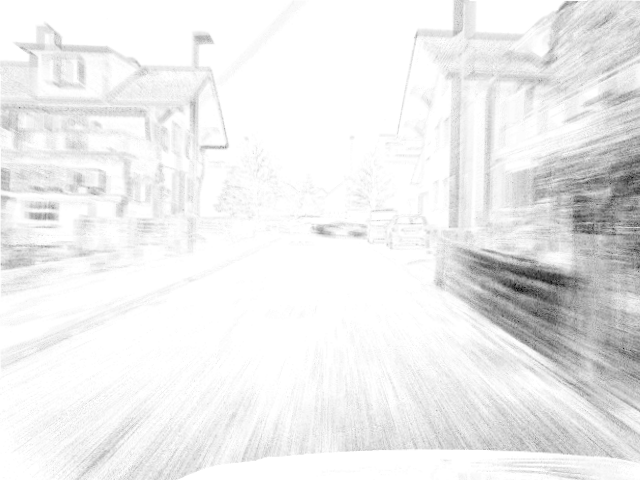}}
	&\gframe{\includegraphics[width=\linewidth]{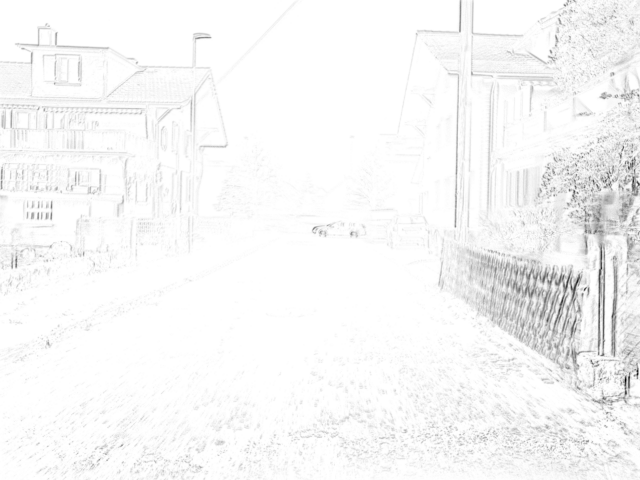}}
        &\includegraphics[width=\linewidth]{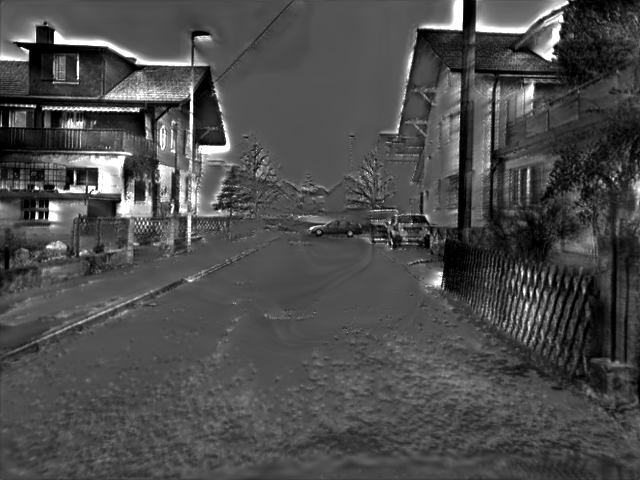}
	&\includegraphics[width=\linewidth]{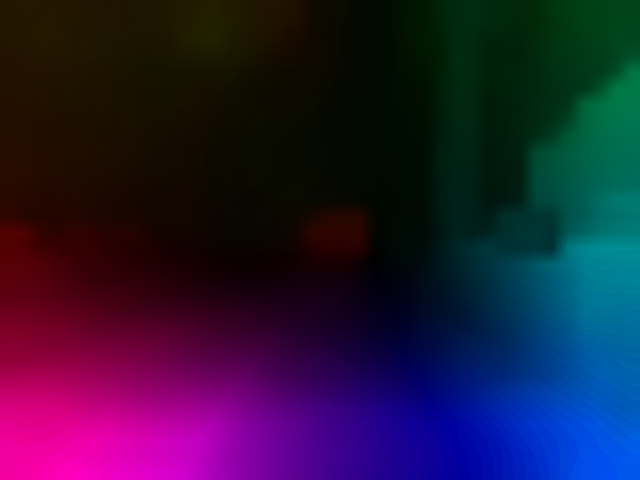}
        &\includegraphics[width=\linewidth]{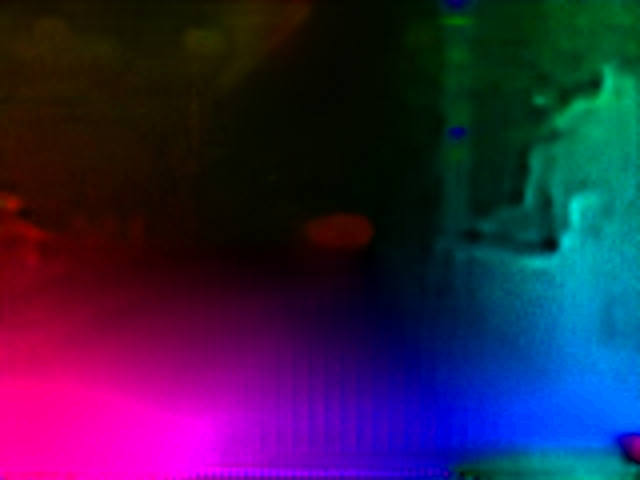}
        &\includegraphics[width=\linewidth]{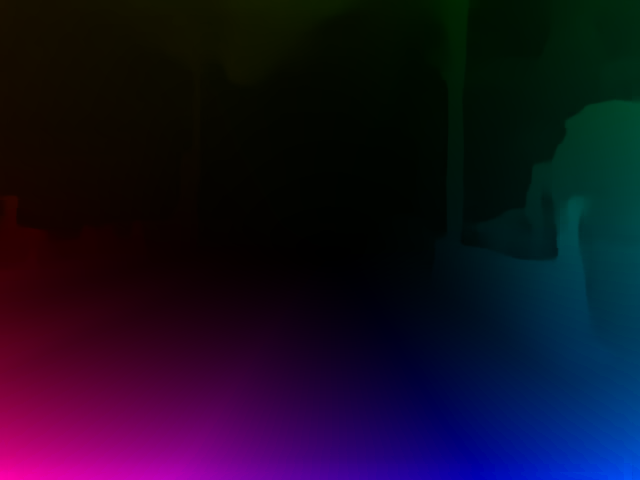}
	\\

        & (a) Events
        & (b) IWEs
        & (c) Intensity
        & (d) Flow (Ours)
        & (e) Flow ECCV'24 \cite{Hamann24eccv}
        & (f) Flow E-RAFT \cite{Gehrig21threedv}
        \\[-0.5ex]
	\end{tabular}
	}
    \caption{\emph{Qualitative comparisons on DSEC}. From left to right: (a) input events; (b) image of warped events (IWE) with our predicted flow; (c) our predicted intensity; (d) our predicted flow; (e)-(f) two baseline methods that predict optical flow (USL and SL, respectively).
    \label{fig:flow_array}}
\end{figure*}

\textbf{Accuracy}. 
\Cref{tab:exp:dsec} presents a comprehensive comparison between our method and baseline approaches, where the methods that require GT optical flow (SL) and those that do not (MB/USL) are compared respectively in two categories.
As mentioned in \cref{sec:related:flow}, SL methods achieve smaller errors because they are not only trained with GT, but also the GT of training and test sequences are from the same sensor (no distribution gap).
In the MB/USL category, our method significantly outperforms all others in the summarized metrics (``All'' columns) and the per-sequence values.
The only exception is that our value of \%Out on interlaken\_00\_b is slightly bigger than that of MotionPriorCM.
Overall, our method achieves great improvements of 20\%, 25\% and 26\% in terms of EPE, AE and \%Out, respectively.
Note that our method even outperforms E-RAFT (SL) in terms of AE, with a reduction of 39\%.
RTEF directly adopts FWL as the objective function to optimize, so it has the highest FWL scores on all sequences, followed by our method in the second place.
Note that a high FWL value may not always be good, as it can be indicative of event collapse \cite{Shiba24pami}. 
In addition to the overall metrics, our method shows the best performance on all individual sequences that are recorded in various scenarios (urban and highway) and various illumination conditions (daylight and night).
This demonstrates the robustness and versatility of the proposed approach.

\def\figWidth{0.135\linewidth}
\begin{figure*}[t]
	\centering
    {\footnotesize
    \setlength{\tabcolsep}{1pt}
	\begin{tabular}{
	>{\centering\arraybackslash}m{0.3cm} 
	>{\centering\arraybackslash}m{\figWidth} 
	>{\centering\arraybackslash}m{\figWidth}
	>{\centering\arraybackslash}m{\figWidth}
	>{\centering\arraybackslash}m{\figWidth}
        >{\centering\arraybackslash}m{\figWidth}
        >{\centering\arraybackslash}m{\figWidth}
        >{\centering\arraybackslash}m{\figWidth}
        }

	\rotatebox{90}{\makecell{boxes}}
	&\includegraphics[width=\linewidth]{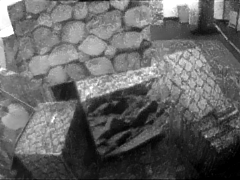}
	&\includegraphics[width=\linewidth]{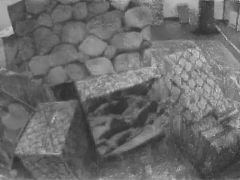}
        &\includegraphics[width=\linewidth]{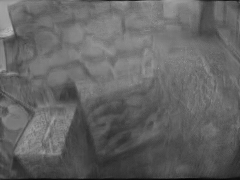}
        &\includegraphics[width=\linewidth]{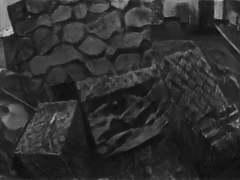}
	&\includegraphics[width=\linewidth]{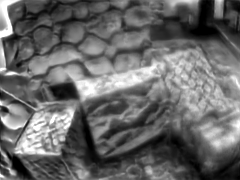}
        &\includegraphics[width=\linewidth]{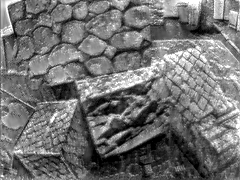}
        &\includegraphics[width=\linewidth]{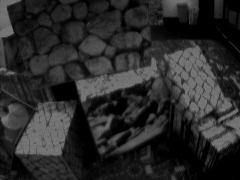}
	\\[-0.5ex]
    
        \rotatebox{90}{\makecell{rooftop\_05}}
	&\includegraphics[trim=10px 0px 10px 0px, clip, width=\linewidth]{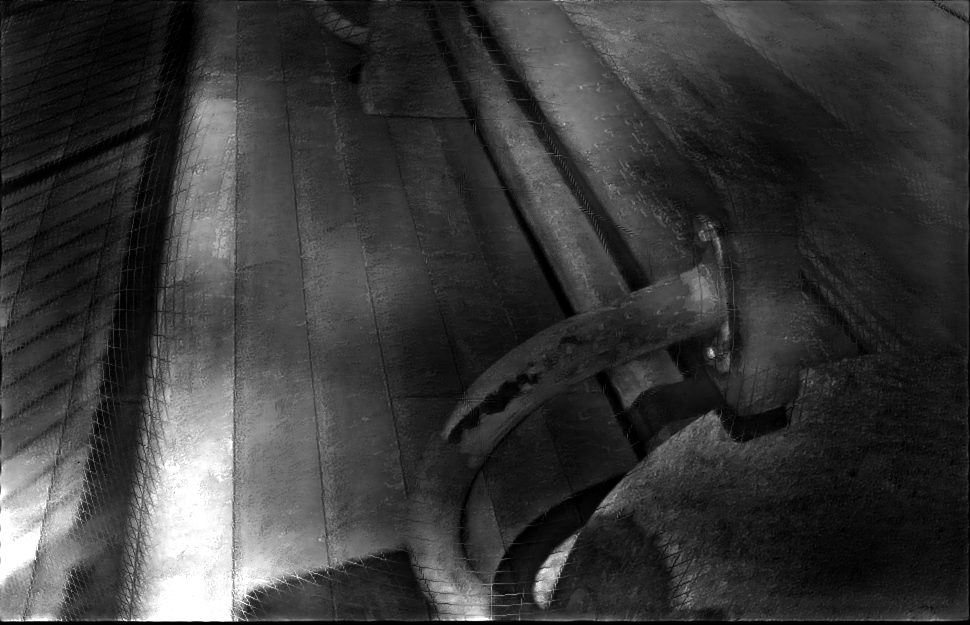}
	&\includegraphics[trim=10px 0px 10px 0px, clip, width=\linewidth]{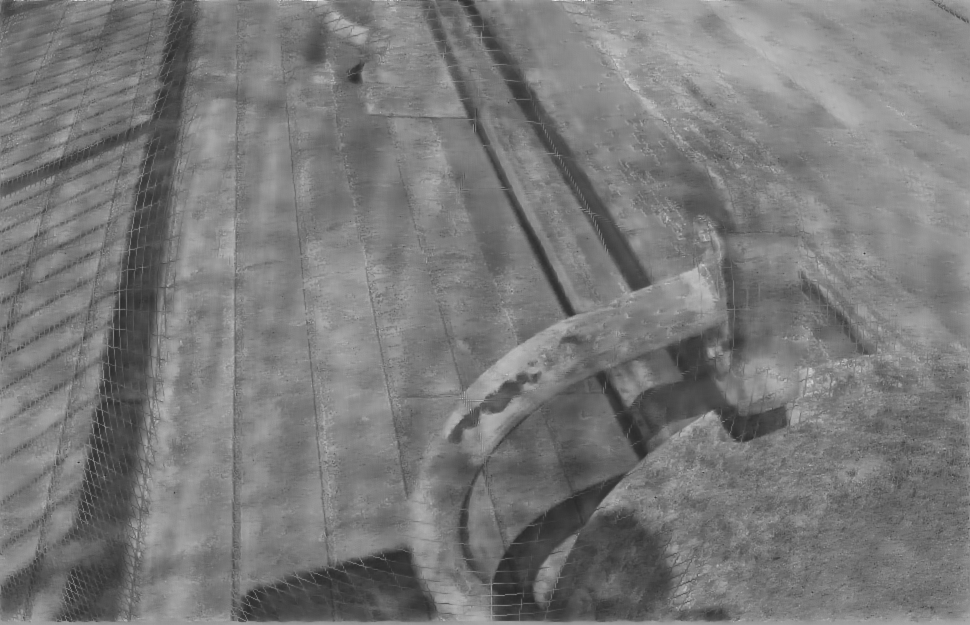}
        &\includegraphics[trim=10px 0px 10px 0px, clip, width=\linewidth]{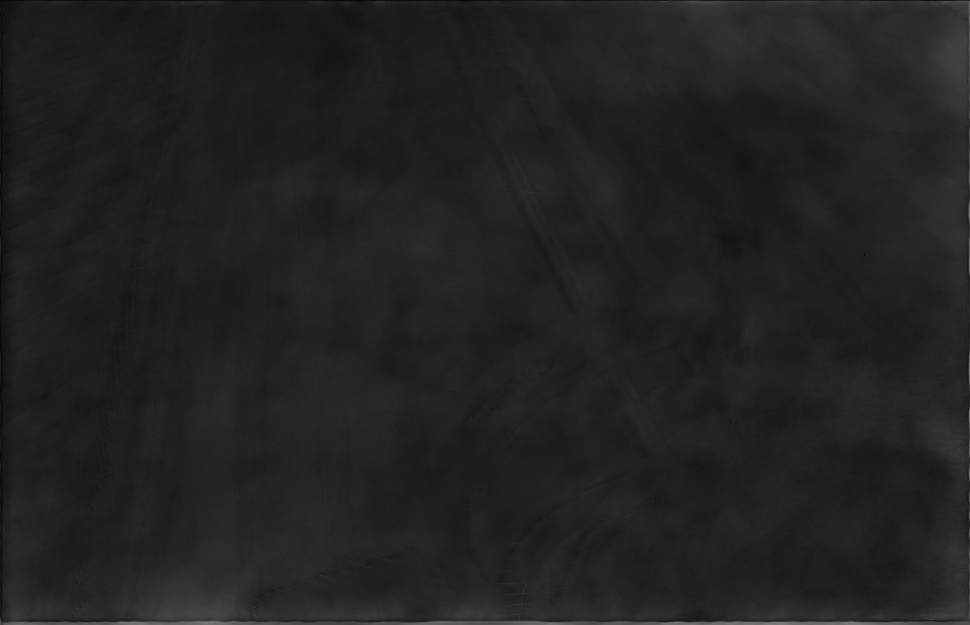}
        &\includegraphics[trim=10px 0px 10px 0px, clip, width=\linewidth]{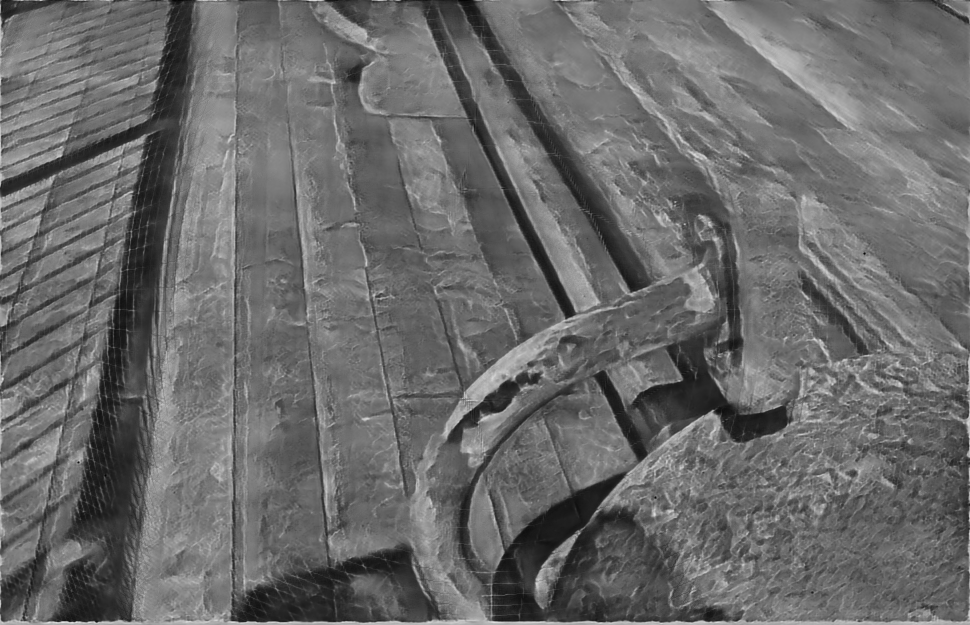}
        &\includegraphics[trim=10px 0px 10px 0px, clip, width=\linewidth]{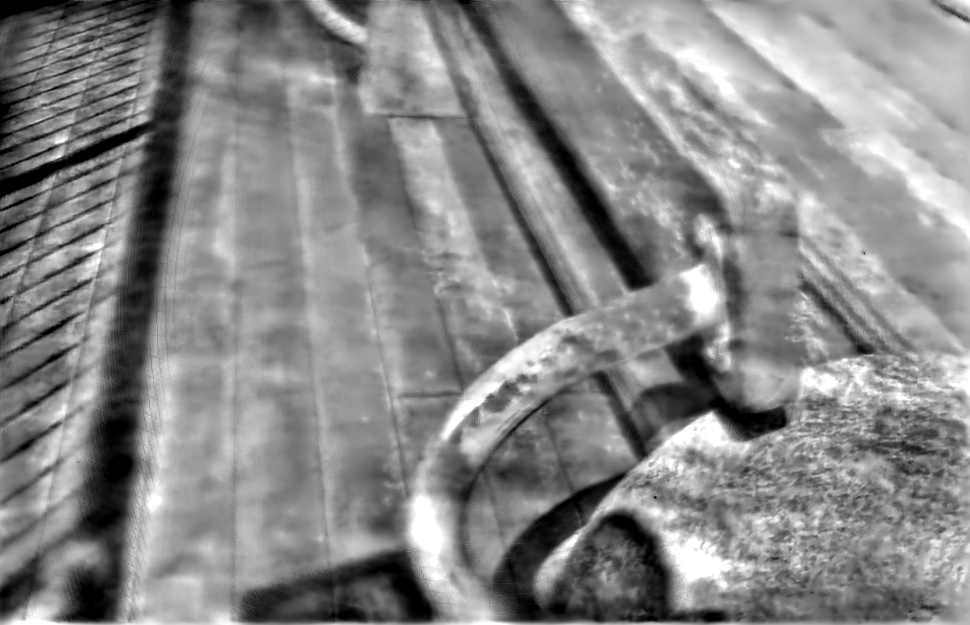}
	&\includegraphics[width=\linewidth]{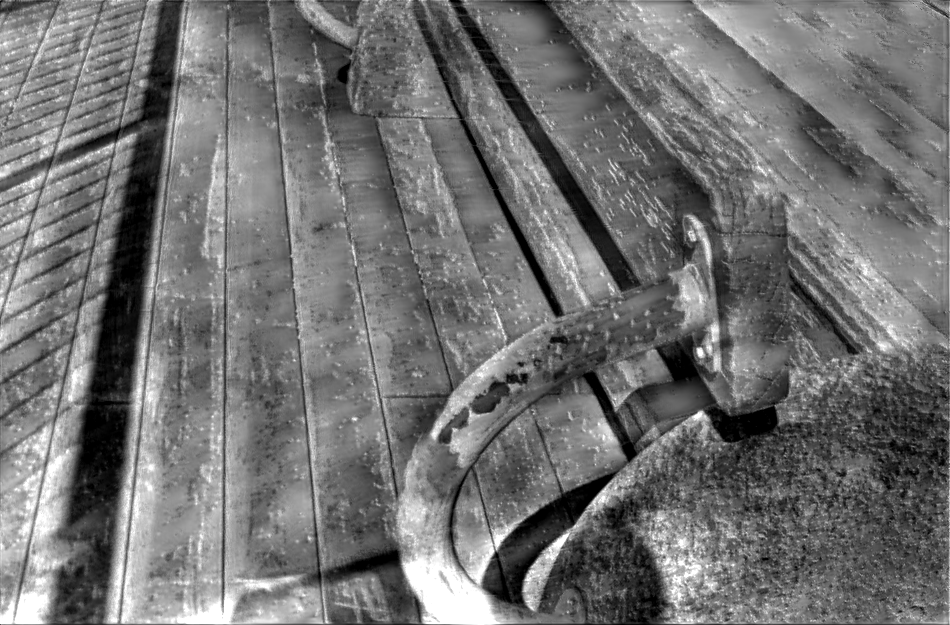}
        &\includegraphics[trim=10px 0px 10px 0px, clip, width=\linewidth]{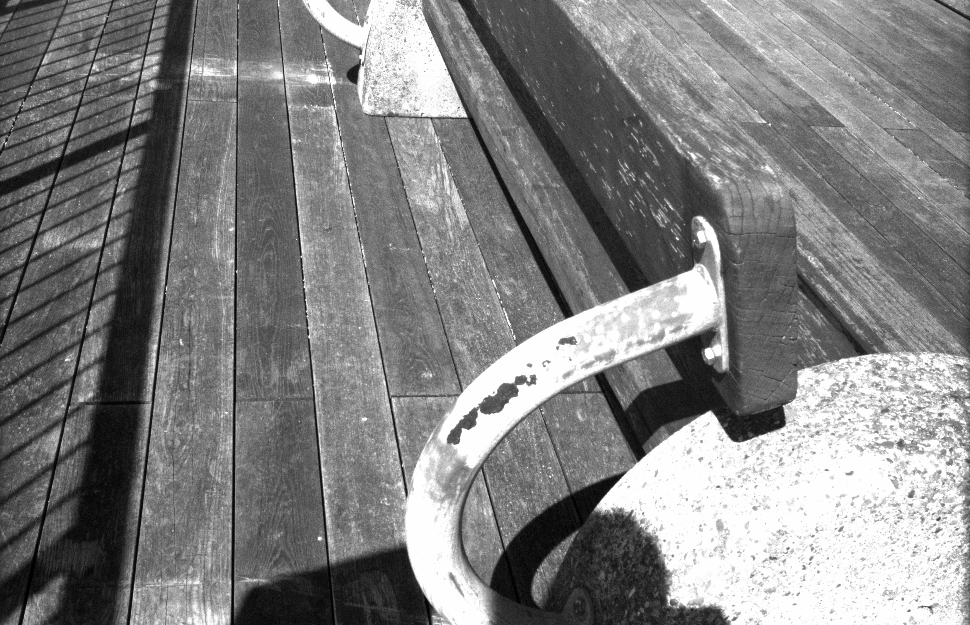}
	\\[-0.3ex]
	
        \rotatebox{90}{\makecell{selfie}}
	&\includegraphics[width=\linewidth]{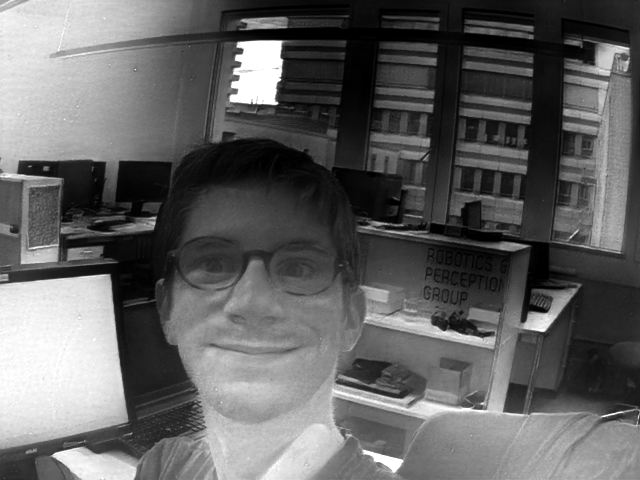}
	&\includegraphics[width=\linewidth]{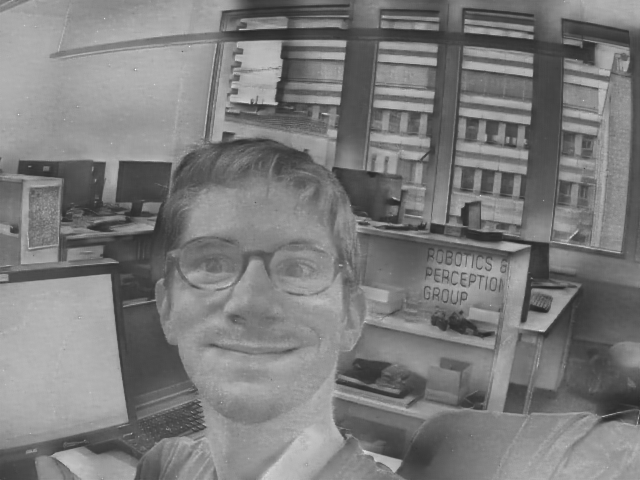}
        &\includegraphics[width=\linewidth]{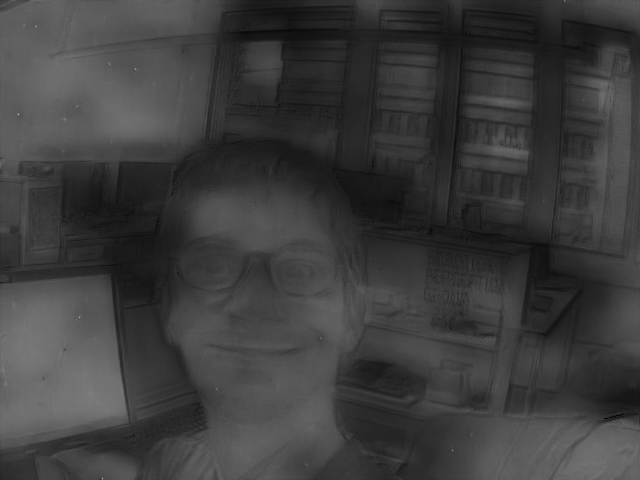}
	&\includegraphics[width=\linewidth]{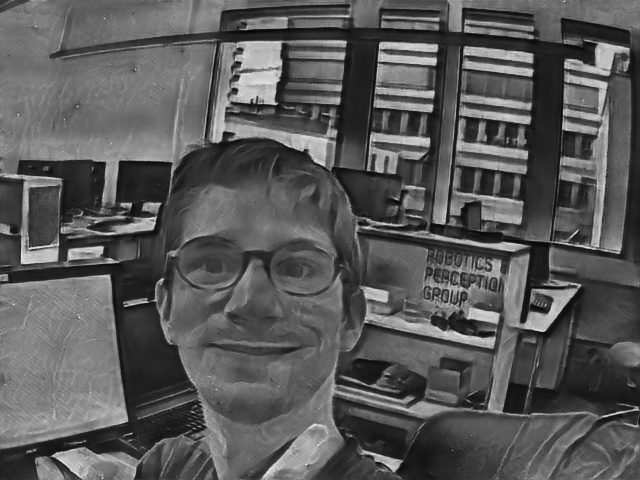}
        &\includegraphics[width=\linewidth]{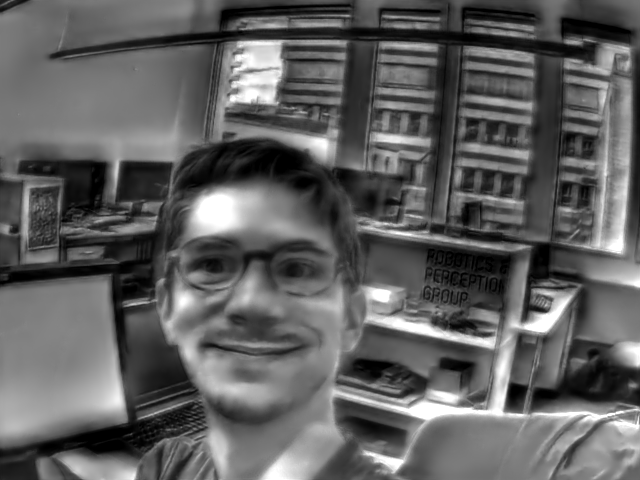}
        &\includegraphics[width=\linewidth]{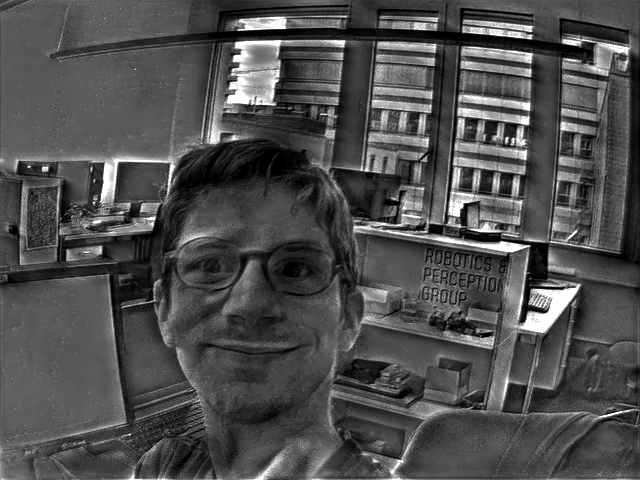}
        &\includegraphics[width=\linewidth]{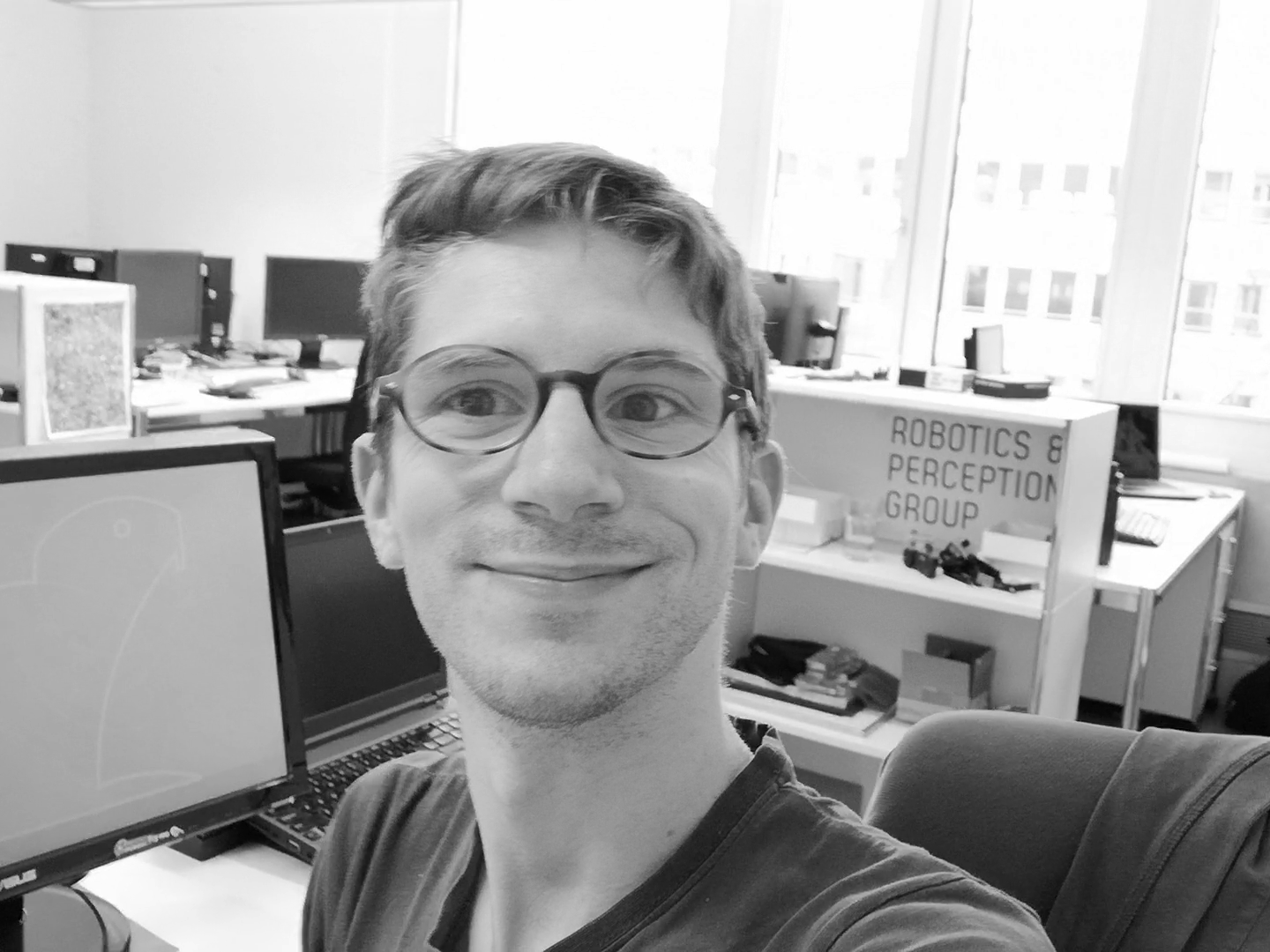}
	\\[-0.5ex]

        \rotatebox{90}{\makecell{poster (fast)}}
	&\includegraphics[width=\linewidth]{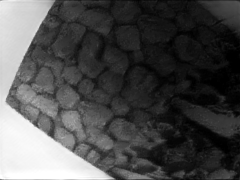}
	&\includegraphics[width=\linewidth]{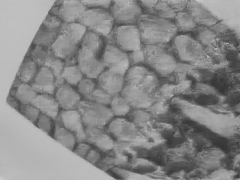}
        &\includegraphics[width=\linewidth]{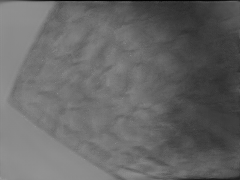}
	&\includegraphics[width=\linewidth]{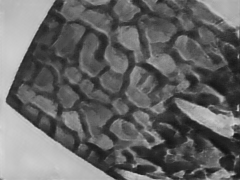}
        &\includegraphics[width=\linewidth]{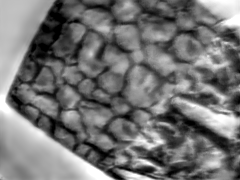}
        &\includegraphics[width=\linewidth]{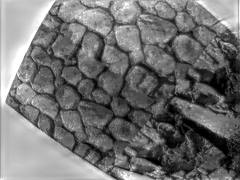}
        &\includegraphics[width=\linewidth]{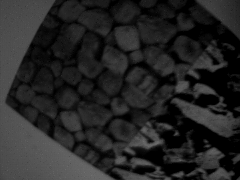}
	\\

        & E2VID \cite{Rebecq19pami}
        & FireNet \cite{Scheerlinck20wacv}
        & SPADE-E2VID \cite{Gantier21tip}
        & ET-Net \cite{Weng21cvpr}
        & BTEB \cite{Paredes21cvpr}
        & Ours
        & References
        \\[-.8ex]
	\end{tabular}
	}
    \caption{\emph{Qualitative image-intensity comparisons} on ECD, BS-ERGB, HDR and ECD-fast data.
    Best viewed when zoomed in.
    \label{fig:image_array}
    \vspace{-1ex}
    }
\end{figure*}

\Cref{fig:flow_array} presents a qualitative comparison on the DSEC dataset.
It is clear that the output flow maps (column d) of our method are more precise than those of the most recent USL method (MotionPriorCM, column e). 
Regarding scene appearance, our method not only generates sharp edge maps using the predicted flow (column b) --like flow-based methods--, but also simultaneously produces detailed intensity images (column c).

\textbf{Runtime}.
We also compare the inference time of some methods in \cref{tab:exp:dsec} (``$t_\text{inf}$'' column).
Our method shows the best efficiency despite predicting two quantities (flow and intensity).
Note that MotionPriorCM also uses a U-Net architecture, where the only difference from ours is the way of up-sampling features in the decoder (we use bilinear interpolation whereas \cite{Hamann24eccv} uses 2D transposed convolutions).

\subsection{Evaluation of Image Intensity Reconstruction}
\label{sec:experim:intensity}

The results of the benchmarks with and without GT reference images on the BS-ERGB and HDR datasets are presented in \cref{tab:exp:evreal}.
In addition, the inference times of all methods are reported in \cref{tab:intensity_inf_time}.

Among supervised-learning (SL) approaches, ET-Net achieves the best image quality in five out of six metrics, but it is the slowest by a margin.
Conversely, FireNet reports the shortest inference time, but its quality metrics are not as good as other methods.

In the unsupervised-learning (USL) category, our method reports comparable results in full-reference metrics as those of BTEB~\cite{Paredes21cvpr}.
However, our method significantly outperforms BTEB in all three metrics on the HDR dataset, with an improvement of around 50\%, where many SL methods are also surpassed.
In particular, we achieve the best MANIQA value compared to all baselines.
This implies that our method better unlocks the HDR properties of event cameras.
In terms of inference time, our model stands in the middle of the ranking; however, our model outputs both flow and intensity, while the others only output intensity.

\begin{table}[t]
\centering
\adjustbox{max width=\linewidth}{%
\setlength{\tabcolsep}{2pt}
\begin{tabular}{ll*{3}{S[table-format=2.3]}l*{3}{S[table-format=2.3]}}
\toprule
  &  & \multicolumn{3}{c}{BS-ERGB}
  &
  & \multicolumn{3}{c}{HDR} \\
 \cmidrule(l{1mm}r{1mm}){3-5}
 \cmidrule(l{1mm}r{1mm}){7-9}

 Type & Method
 & \text{MSE$\downarrow$} & \text{SSIM$\uparrow$} & \text{LPIPS$\downarrow$} 
 & ~
 & \text{BRISQUE$\downarrow$} & \text{NIQE$\downarrow$} & \text{MANIQA$\uparrow$} \\
 
\midrule
\multirow{4}{*}{\shortstack{SL}}

 & E2VID~\cite{Rebecq19pami}
 & 0.142 & 0.332 & 0.561
 &
 & \bnum{12.626} & 4.273 & 0.298 \\ 

 & FireNet~\cite{Scheerlinck20wacv}
 & 0.099 & 0.336 & 0.531
 &
 & 18.566 & 3.846 & 0.297 \\

 & SPADE-E2VID~\cite{Gantier21tip}
 & 0.091 & 0.346 & 0.631
 &
 & 24.511 & 7.174 & 0.277 \\

 & ET-Net~\cite{Weng21cvpr}
 & \bnum{0.073} & \bnum{0.374} & \bnum{0.441}
 &
 & 19.195 & \bnum{3.448} & \bnum{0.324} \\

\midrule
\multirow{2}{*}{\shortstack{USL}}

 & BTEB~\cite{Paredes21cvpr} %
 & \bnum{0.088} & \bnum{0.356} & 0.624
 &
 & 51.469 & 6.239 & 0.182 \\

 & \textbf{Ours}
 & 0.100 & 0.312 & \bnum{0.555} %
 & 
 & \bnum{25.031} & \bnum{3.776} & \bnum{0.397} \\

\bottomrule
\end{tabular}
}
\caption{
\emph{Image-intensity quality assessment}. 
Full-reference evaluation results on the BS-ERGB \cite{Tulyakov22cvpr} dataset (left), 
and non-reference evaluation results on the HDR \cite{Rebecq19pami} dataset (right).
Bold is the best in each category.
The between-frame event slicing is used for BS-ERGB while the fixed-duration slicing ($\Delta t = 100$ms) is used for HDR, where no frame is available.
\label{tab:exp:evreal}
}
\end{table}

\begin{table}[t]
\centering    
    \begin{adjustbox}{max width=\linewidth}
    \setlength{\tabcolsep}{4pt}
    \begin{tabular}{l*{6}{S[table-format=4.3]}}
    \toprule
                & \text{E2VID} & \text{FireNet} & \text{SPADE-E2VID} & \text{ET-Net} & \text{BTEB} & \text{Ours} \\
    Resolution  & \text{(2019)} & \text{(2020)} & \text{(2021)} & \text{(2021)} & \text{(2021)} & \text{(2024)} \\
    \midrule
    $640 \times 480$ & 10.948 & 4.940 & 36.070 & 173.563 & 10.594 & 15.111 \\
    $1280 \times 720$ & 31.041 & 14.667 & 105.869 & 1606.334 & 29.894 & 40.778 \\
    \bottomrule
    \end{tabular}
    \end{adjustbox}
    \caption{
    \label{tab:intensity_inf_time} 
    \emph{Runtime evaluation} [ms] of event-based intensity estimation methods at VGA and HD resolutions.
    Note that our model infers both flow and intensity, while others only infer intensity.
    \vspace{-1ex}
    }
\end{table}

An important remark to make is that the above metrics have limitations: as analyzed in \cite{Zhang22pami}, different metrics often lead to divergent conclusions, where discrepancies occur.
For example, some methods achieve better MSE just by darkening the predictions (as GT images are dark).
Our method reports better LPIPS, but worse MSE and SSIM values than BTEB; 
E2VID reports better BRISQUE than ET-Net, but it performs worse in terms of NIQE and MANIQA. 
The numbers do not seem to entirely reflect visual quality from a human perspective, as we present in the qualitative comparisons of \cref{fig:image_array}.
From the images, it is clear that our approach is able to recover fine details (sharp image) of scene appearance despite the HDR illumination (selfie) and fast motion (poster fast).
\Cref{fig:image_array} also illustrates the shortcomings of the above quantitative metrics. 
All images from SPADE-E2VID are visually blurred, but it reports better MSE and SSIM than FireNet and our method, whose images are markedly sharper and cleaner.

\subsection{Ablation Study}
\label{sec:experim:ablation}

We conduct an ablation study to illustrate the effects of some loss terms and the superiority of joint flow-intensity estimation over flow-only estimation.
The quantitative and qualitative results are presented in \cref{tab:exp:ablation,fig:exp:ablation}.
Please refer to the supplementary for the ablation studies on loss weights and camera contrast threshold $C$.

\textbf{TC Loss.}
First, we disable the TC loss ($\lambda_5 = 0$).
Although flow accuracy does not decrease dramatically (around 10\%), intensity accuracy is reduced significantly (\cref{tab:exp:ablation}, 1st row).
This is also revealed in \cref{fig:exp:ablation} (column a), where texture at valid pixels is recovered, while invalid pixels are filled with markedly incorrect values due to the lack of constraints.
This confirms our claim in \cref{sec:intro}, that better estimation for both visual quantities is achieved by leveraging their synergies.

\textbf{TV regularization.}
Next, we disable the TV regularization of both flow and intensity ($\lambda_3=\lambda_4=0$).
Contrary to the previous test, intensity accuracy remains while flow accuracy drops considerably (\cref{tab:exp:ablation}, 2nd row).
The same conclusion can be drawn from \cref{fig:exp:ablation} (column b), where the estimated flow shows artifacts, mostly at invalid pixels.
As a consequence, the output image intensity at such pixels is not as smooth as that of our main model (column c).

\textbf{Flow-only Estimation.}
Finally, we train the network to predict only optical flow ($\lambda_1=\lambda_4=\lambda_5 = 0$, thus changing the number of output channels from three to two).
In this case, the loss function basically reduces to the one used by recent purely CMax-based optical flow estimation methods \cite{Hamann24eccv,Shiba22eccv}.
As reported in the third row of \cref{tab:exp:ablation}, the flow accuracy is slightly better than that of w/o TV reg., but still falls far behind our main model, with respective gaps of 43\%, 33\%, 53\% in terms of EPE, AE and \%Out.
This again confirms the advantages of our proposed joint estimation method over the separate ones.

\subsection{Single Network vs.~Dual Network}
To illustrate the superiority of joint learning with a single network, 
we compare it to a dual network (two U-Nets, one for optical flow prediction and the other for intensity reconstruction), with configuration identical to ours (loss function, loss weights and training settings).
\Cref{tab:exp:ablation} show that the estimation results of the dual network (4th row) are considerably worse than those of the single U-Net model (last row).
Our joint learning scheme enables one single network to learn the motion and appearance, as well as their synergies, thus achieving better performance than separately.

\def\figWidth{0.24\linewidth}
\begin{figure}[t]
	\centering
    {\footnotesize
    \setlength{\tabcolsep}{1pt}
	\begin{tabular}{
	>{\centering\arraybackslash}m{\figWidth} 
	>{\centering\arraybackslash}m{\figWidth}
	>{\centering\arraybackslash}m{\figWidth}
	>{\centering\arraybackslash}m{\figWidth}
        }

	\includegraphics[width=\linewidth]{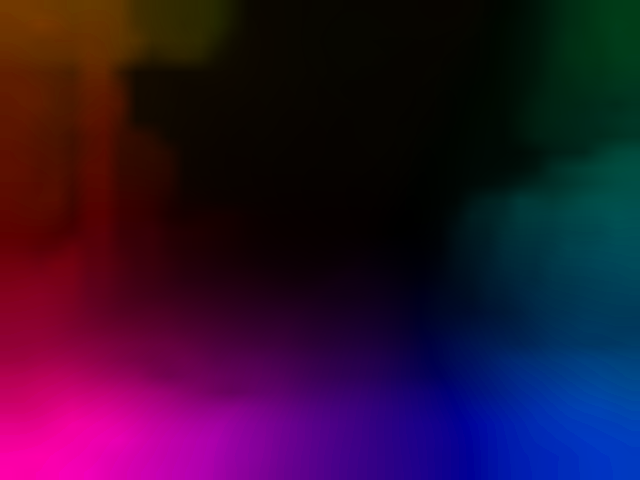}
        &\includegraphics[width=\linewidth]{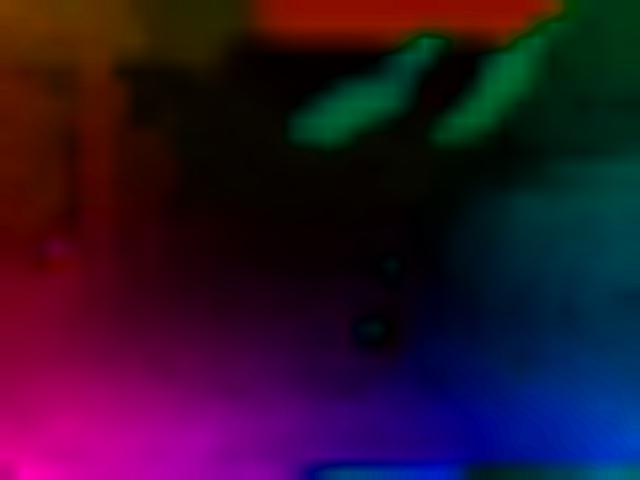}
	&\includegraphics[width=\linewidth]{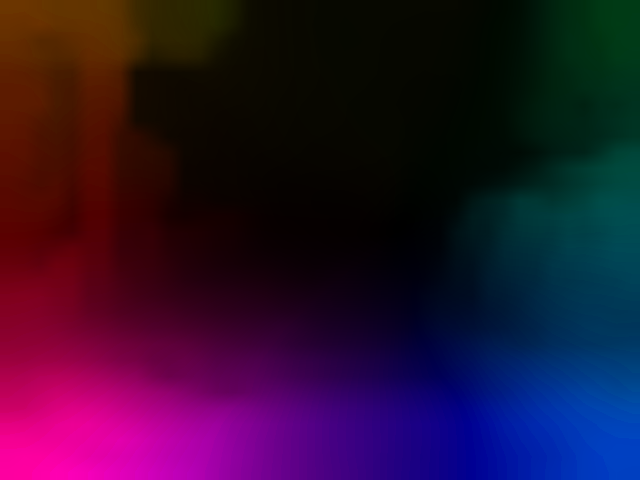}
        &\includegraphics[width=\linewidth]{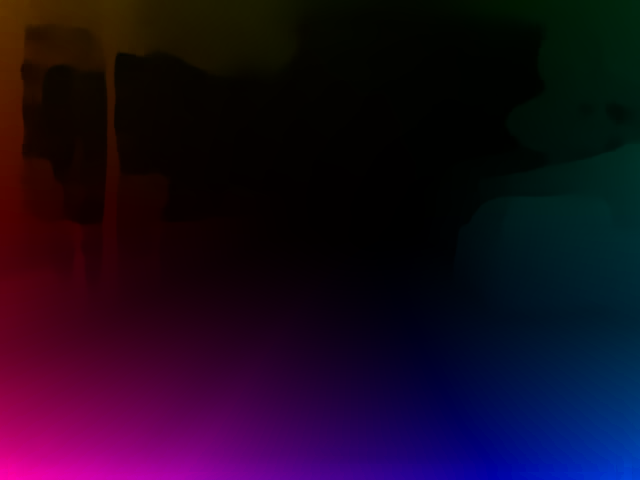}
	\\[-0.3ex]
		
	\includegraphics[width=\linewidth]{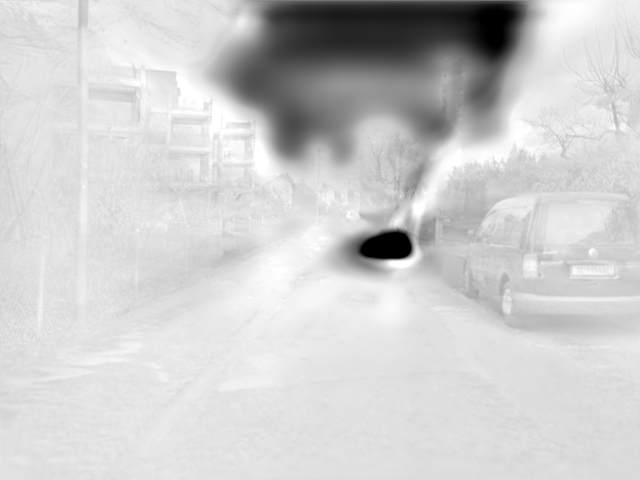}
        &\includegraphics[width=\linewidth]{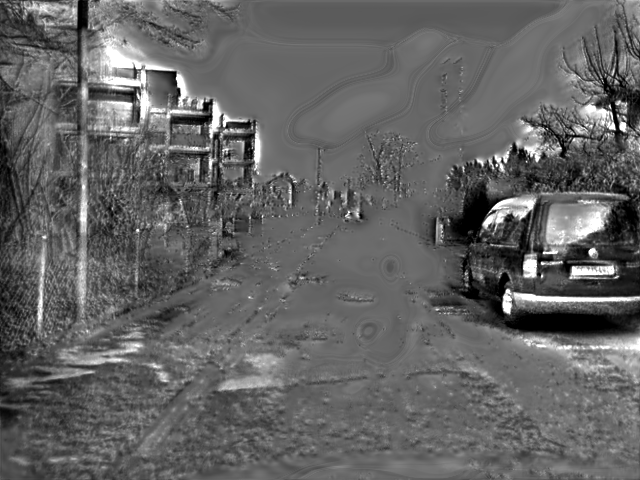}
	&\includegraphics[width=\linewidth]{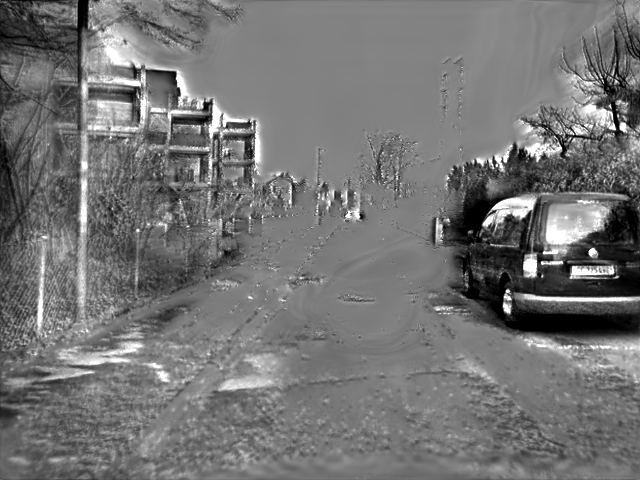}
        &\includegraphics[width=\linewidth]{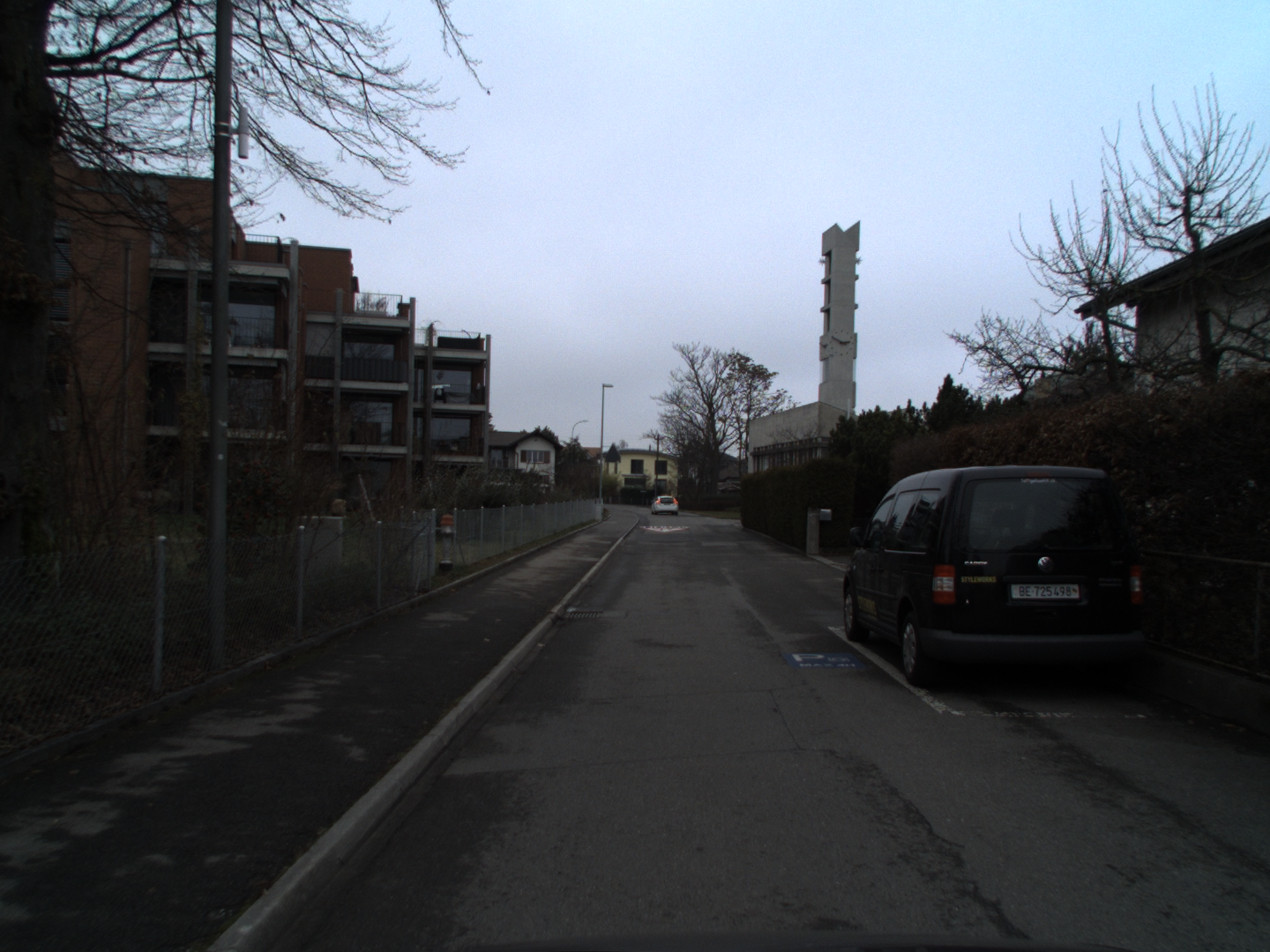}
	\\

        (a) w/o TC loss
        & (b) w/o TV reg.
        & (c) Ours
        & (d) References
        \\[-.8ex]
	\end{tabular}
	}
    \caption{
    \emph{Qualitative results of ablation study.}
    The above are results from the thun\_01\_b sequence of DSEC.
    The reference optical flow is generated using E-RAFT \cite{Gehrig21threedv}.
    \label{fig:exp:ablation}
    }
\end{figure}

\begin{table}[t!]
\centering
\adjustbox{max width=\linewidth}{%
\setlength{\tabcolsep}{2pt}
\begin{tabular}{l*{3}{S[table-format=2.3]}l*{3}{S[table-format=2.3]}}
\toprule
  & \multicolumn{3}{c}{Flow}
  &
  & \multicolumn{3}{c}{Intensity} \\
 \cmidrule(l{1mm}r{1mm}){2-4}
 \cmidrule(l{1mm}r{1mm}){6-8}

 Ablation
 & \text{EPE$\downarrow$} & \text{AE$\downarrow$} & \text{$\%$Out$\downarrow$}
 & ~
 & \text{MSE$\downarrow$} & \text{SSIM$\uparrow$} & \text{LPIPS$\downarrow$} \\
 
\midrule

 \text{w/o TC loss}
 & 2.006 & 7.915 & 13.964
 &
 & 0.160 & 0.230 & 0.637 \\ 

 \text{w/o TV reg.}
 & 3.303 & 14.934 & 24.95
 &
 & \bnum{0.098} & 0.295 & \bnum{0.548} \\

 \text{Flow-only}
 & 3.134 & 9.609 & 23.557
 &
 & \novalue & \novalue & \novalue \\

 \text{Dual network}
 & 3.30 & 10.15 & 22.85
 &
 & 0.11 & 0.29 & 0.59 \\

 \text{\textbf{Ours}}
 & \bnum{1.781} & \bnum{6.439} & \bnum{11.241}
 &
 & \bnum{0.100} & \bnum{0.312} & 0.555 \\

\bottomrule
\end{tabular}
}
\caption{
\emph{Quantitative results of an ablation study.}
Flow evaluation is performed on the DSEC dataset and intensity evaluation is done on the BS-ERGB dataset. 
\label{tab:exp:ablation}
}
\vspace{-1ex}
\end{table}

\section{Limitations}
\label{sec:limitations}

The event camera output depends on scene texture and camera motion.
In regions where no events are produced, it is thus difficult to recover motion parameters and/or scene appearance.
In this case, regularization is used to fill in those pixels by encouraging spatial smoothness. 
However, this may cause inaccuracies.
This issue could be overcome by combining two visual modalities (events and frames),
not without considering data fusion challenges, 
and/or by adding recurrent connections (at the expense of increasing the initialization time and inertia of the system \cite{Rebecq19pami}).

Events triggered by flickering lights or hot pixels do not satisfy the brightness constancy assumption, which is the basis of all involved MB/USL methods. 
As outliers, they can undermine estimation accuracy. 
Learning-based methods have some capacity to deal with such outliers. 
However outliers would be more sensibly treated in pre-processing by some denoising method, or more recently, jointly \cite{Shiba25iccv}.

\section{Conclusion}
\label{sec:conclusion}
We have presented the first unsupervised joint learning framework for optical flow and image intensity estimation with event cameras.
This enabled us to train a single and lightweight network for predicting both visual quantities simultaneously, exploiting their synergies.
It is designed to match the natural sensing principle of event cameras, that is, that motion and appearance are intertwined and, therefore, shall be jointly estimated. 
A comprehensive and well-behaved loss function is proposed by combining the event generation model, photometric consistency, event alignment and regularization.
The experiments demonstrate that our method is accurate and efficient:
(\emph{i}) it achieves the highest accuracy among MB/USL optical flow methods;
(\emph{ii}) it shows competitive intensity quality, especially in HDR conditions;
and
(\emph{iii}) it reports very short inference time while predicting both flow and image intensity.

\section*{Acknowledgments}
Funded by the Deutsche Forschungsgemeinschaft (DFG, German Research Foundation) under Germany’s Excellence Strategy – EXC 2002/1 ``Science of Intelligence'' – project number 390523135.

\ifarxiv
\section*{Supplementary Material}
\else
\clearpage
\setcounter{page}{1}
\maketitlesupplementary
\fi

\section{Additional Ablation Study}
\label{sec:suppl:ablation}
As an expansion of \cref{sec:experim:ablation}, we present the results of the ablation studies on loss weights and contrast threshold $C$.

\subsection{Loss Weights}
\label{sec:suppl:ablation:weights}
In addition to disabling some loss terms to show their effects (\cref{sec:experim:ablation}), an ablation study on the weight of CMax loss $\lambda_2$ is performed to show how the ratio between CMax and PhE influences model performance.
The results are reported in the upper half of \cref{tab:suppl:ablation}.
It turns out that increasing or decreasing $\lambda_2$ does not lead to a further improvement in  performance.
Therefore, the choice of $\lambda_1$ and $\lambda_2$ to train our main model yields a sensible combination of the CMax and PhE loss terms.

\subsection{Contrast Threshold}
The contrast threshold $C$ can vary across event cameras and change even within the same dataset \cite{Stoffregen20eccv}.
Hence, it is worth analyzing the influence of $C$ on model performance.

The works of \cite{Guo24epba,Guo24eccv} have shown that the PhE and its linearized version are insensitive to the value of $C$, due to the PhE being calculated using thousands or millions of events instead of a few. 
The user just needs to set a mean value for $C$, and then the optimizer seeks a balanced motion and brightness to best explain the events.
This finding has also been verified by the results in \cref{fig:image_array} in our main paper and \cref{fig:suppl_comp,fig:suppl_results,fig:suppl_zoom_in} in this supplementary: 
our model was trained only on the DSEC dataset (Prophesee Gen3) with $C = 0.2$, but nevertheless it is able to predict precise optical flow and image intensity on several other datasets, such as ECD (DAVIS240C), HDR (Samsung DVS Gen3) and BS-ERGB (Prophesee Gen4).

Furthermore, we also train models with different $C$ values, and report the results in the lower half of \cref{tab:suppl:ablation}.
The results agree with the statements above; the accuracy remains approximately constant for different $C$ values.

\begin{table}[ht]
\makeatletter\def\@captype{table}
\centering
\adjustbox{max width=\linewidth}{%
\setlength{\tabcolsep}{2pt}
\begin{tabular}{cc*{3}{S[table-format=2.3]}l*{3}{S[table-format=2.3]}}
\toprule
  & & \multicolumn{3}{c}{Flow}
  &
  & \multicolumn{3}{c}{Intensity} \\
 \cmidrule(l{1mm}r{1mm}){3-5}
 \cmidrule(l{1mm}r{1mm}){7-9}

 Ablation & Value
 & \text{EPE$\downarrow$} & \text{AE$\downarrow$} & \text{$\%$Out$\downarrow$}
 & ~
 & \text{MSE$\downarrow$} & \text{SSIM$\uparrow$} & \text{LPIPS$\downarrow$} \\
 
\midrule
\multirow{2}{*}{\shortstack{$\lambda_2$}}

 & 0.2
 & 2.777 & 8.123 & 18.929
 &
 & 0.099 & 0.307 & 0.571 \\

 & 5.0
 & 2.34 & 9.726 & 16.417
 &
 & 0.100 & 0.309 & 0.558 \\

\midrule
\multirow{2}{*}{\shortstack{$C$}}

 & 0.1
 & 1.893 & 6.801 & 12.335
 &
 & 0.111 & 0.304 & \bnum{0.549} \\

 & 0.4
 & 1.884 & \bnum{6.422} & 12.319
 &
 & 0.101 & \bnum{0.322} & 0.557 \\

\midrule

 \multicolumn{2}{c}{\textbf{Main ($\lambda_2 = 1.0, C = 0.2$)}}
 & \bnum{1.781} & 6.439 & \bnum{11.241}
 &
 & \bnum{0.100} & 0.312 & 0.555 \\

\bottomrule
\end{tabular}
}
\caption{
\footnotesize{Results of ablation studies on $\lambda_2$ and $C$.}
\label{tab:suppl:ablation}
}
\end{table}

\section{DSEC: Training Sequence Selection}
\label{sec:suppl:dsec_seq}

As mentioned in \cref{sec:limitations}, all MB/USL methods rely on the brightness constancy assumption to estimate optical flow or image intensity.
Events triggered by flickering lights or by hot pixels would undermine the estimation accuracy.
To this end, the sequences in the training split of the DSEC dataset \cite{Gehrig21threedv} are screened before being used for training, according to the data quality.
Here we present the list of the selected training sequences in \cref{tab:dsec_seqs}.

\begin{table}[h]
\centering
\begin{adjustbox}{max width=1.0\linewidth}
\setlength{\tabcolsep}{3pt}
\renewcommand{\arraystretch}{1.1}
\begin{tabular}{llll}
\toprule
zurich\_city\_02\_a & zurich\_city\_02\_b & zurich\_city\_02\_c & zurich\_city\_02\_d \\
zurich\_city\_02\_e & zurich\_city\_03\_a & zurich\_city\_04\_a & zurich\_city\_04\_b \\ zurich\_city\_04\_c & zurich\_city\_04\_d & zurich\_city\_04\_e & zurich\_city\_04\_f \\ zurich\_city\_05\_a & zurich\_city\_06\_a & zurich\_city\_07\_a & zurich\_city\_08\_a \\ zurich\_city\_11\_a & zurich\_city\_11\_b & interlaken\_00\_c   & interlaken\_00\_d   \\ interlaken\_00\_e   & interlaken\_00\_f   & interlaken\_00\_g   & thun\_00\_a         \\
\bottomrule
\end{tabular}
\end{adjustbox}
\caption{
Sequences from the DSEC training split that were used for training our model.
\label{tab:dsec_seqs}
}
\end{table}

\section{BS-ERGB: Evaluation Sequence Cropping}
\label{sec:suppl:bsergb_seq}

Here we describe the removal of the last seconds of the may29\_rooftop sequences, for the intensity evaluation on the BS-ERGB \cite{Tulyakov22cvpr} dataset (970 $\times$ 625 px resolution), as presented in \cref{tab:bsergb_seqs}. 
We perform the cropping because the camera is not moving in the last seconds (mentioned in \cref{sec:experim:setup}) of those sequences, thus the recorded event data is pure noise.

\begin{table}[h]
\centering
    \begin{adjustbox}{max width=0.8\linewidth}
    \setlength{\tabcolsep}{4pt}
    \begin{tabular}{lcc}
    \toprule
    Sequence & Start Time [s] & End Time [s]\\
    \midrule
    may29\_rooftop\_handheld\_01 & 0.0 & 24.0 \\
    may29\_rooftop\_handheld\_02 & 0.0 & 17.0 \\
    may29\_rooftop\_handheld\_03 & 0.0 & 14.0 \\
    may29\_rooftop\_handheld\_05 & 0.0 & 9.5 \\
    \bottomrule
    \end{tabular}
    \end{adjustbox}
    \caption{\label{tab:bsergb_seqs} 
    Details of the cropping of the may29\_rooftop sequences in the BS-ERGB dataset.
    }
\end{table}

\section{Additional Qualitative Results}
\label{sec:suppl:add_results}
In this section, we present additional qualitative results of our model on the high-resolution BS-ERGB \cite{Tulyakov22cvpr} datasets in \cref{fig:suppl_results}, including (a) input events, (b) image of warped events (IWE) with
the predicted flow, (c) optical flow, (d) image intensity and (e) reference images.
A qualitative comparison between our method and other baselines on the same dataset is also presented in \cref{fig:suppl_comp}, where some regions of interest (ROIs) are highlighted for further zoomed-in visualization in \cref{fig:suppl_zoom_in}.

\def\figWidth{0.19\linewidth}
\begin{figure*}[ht]
	\centering
    {\footnotesize
    \setlength{\tabcolsep}{1pt}
	\begin{tabular}{
        >{\centering\arraybackslash}m{0.3cm} 
	>{\centering\arraybackslash}m{\figWidth} 
	>{\centering\arraybackslash}m{\figWidth}
	>{\centering\arraybackslash}m{\figWidth}
	>{\centering\arraybackslash}m{\figWidth}
        >{\centering\arraybackslash}m{\figWidth}
        }

        \rotatebox{90}{\makecell{may29\_01}}
	&\gframe{\includegraphics[width=\linewidth]{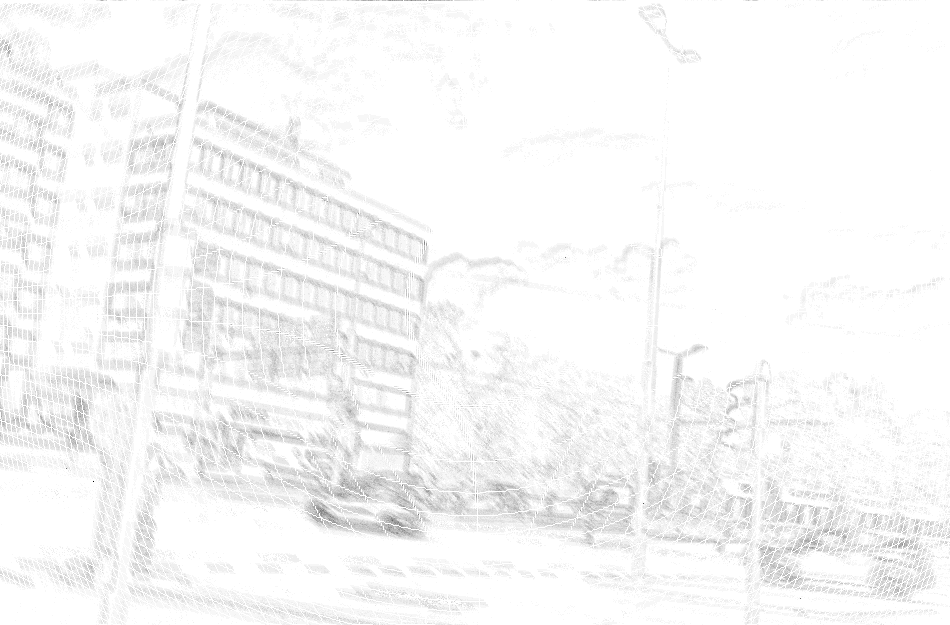}}
	&\gframe{\includegraphics[width=\linewidth]{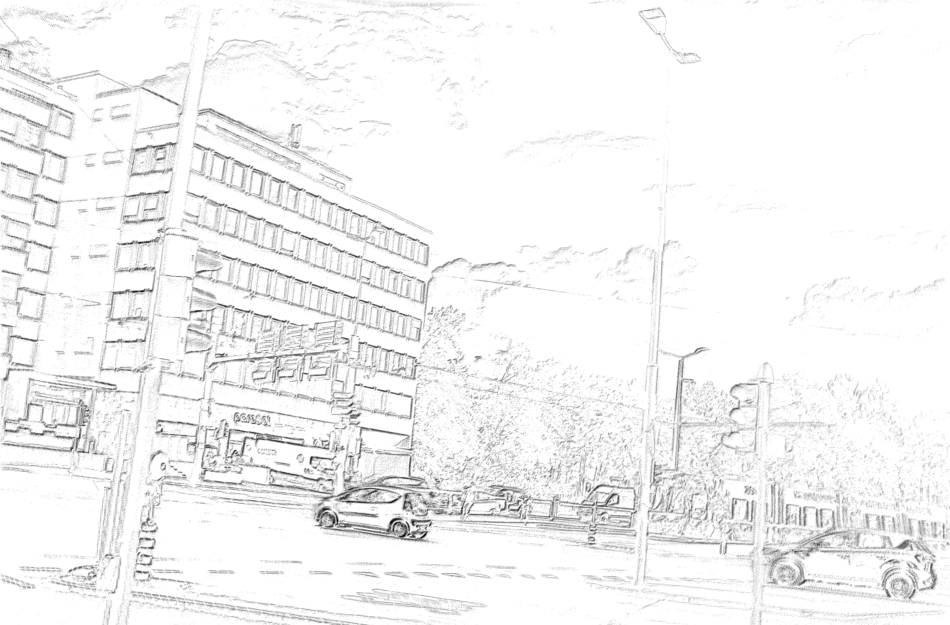}}
        &\includegraphics[width=\linewidth]{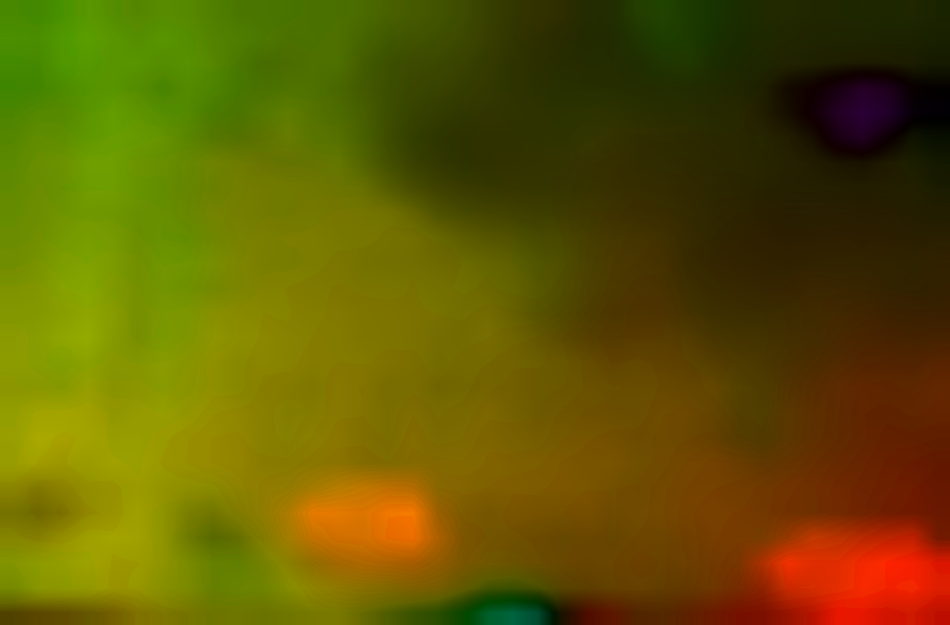}
	&\includegraphics[width=\linewidth]{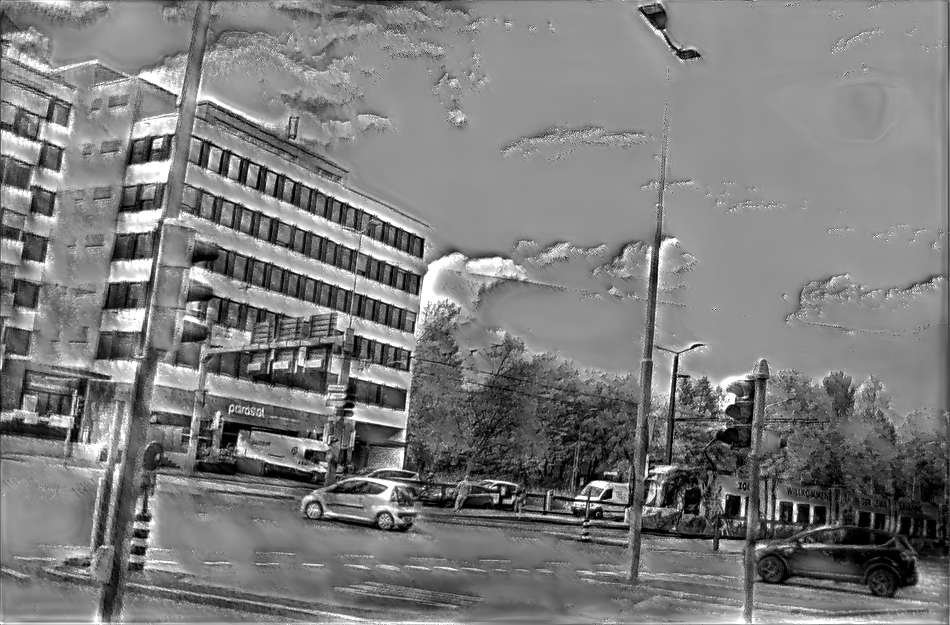}
        &\includegraphics[trim=10px 0px 10px 0px, clip, width=\linewidth]{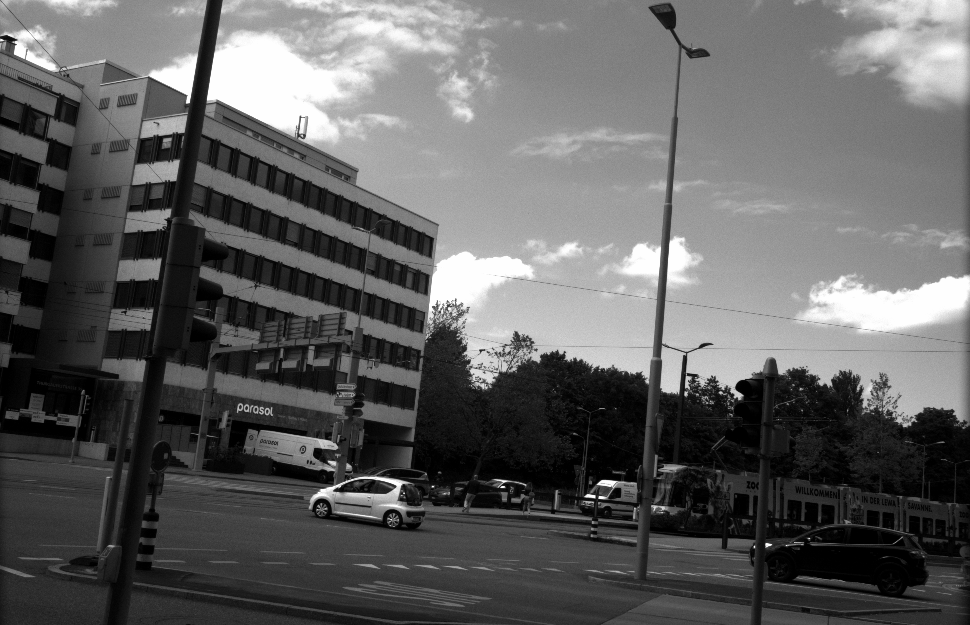}
	\\[-0.2ex]

        \rotatebox{90}{\makecell{may29\_03}}
        &\gframe{\includegraphics[width=\linewidth]{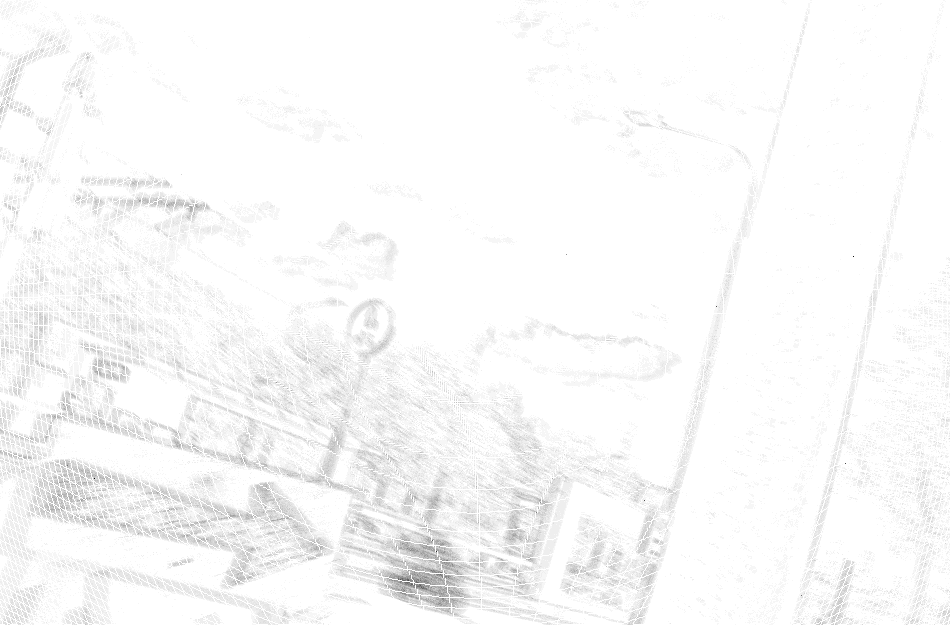}}
	&\gframe{\includegraphics[width=\linewidth]{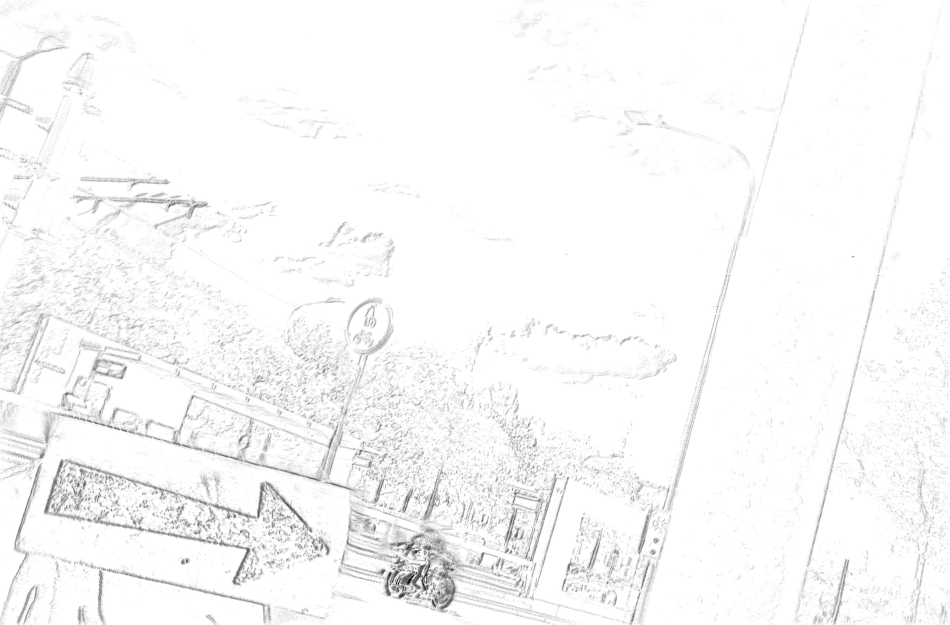}}
        &\includegraphics[width=\linewidth]{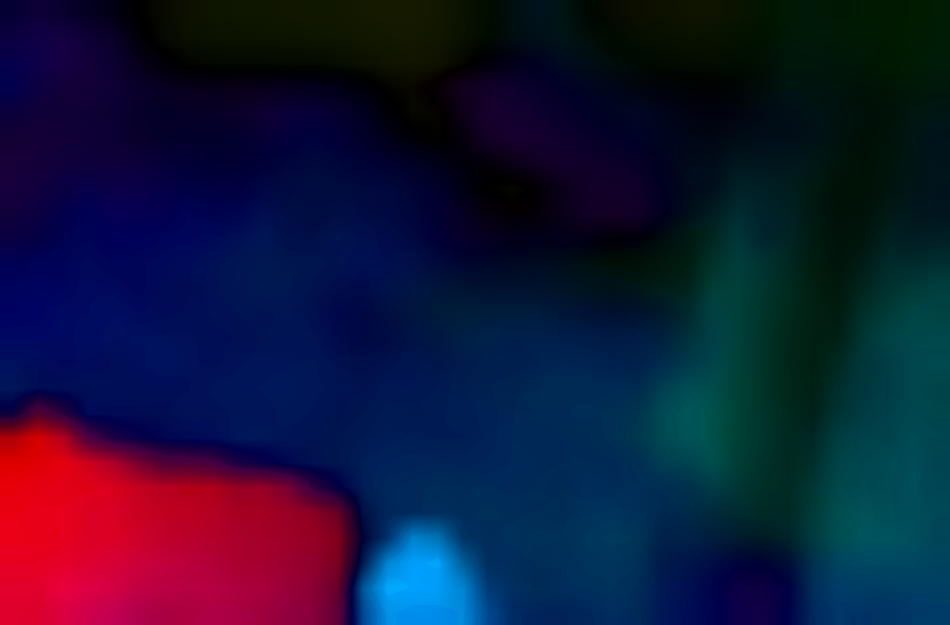}
	&\includegraphics[width=\linewidth]{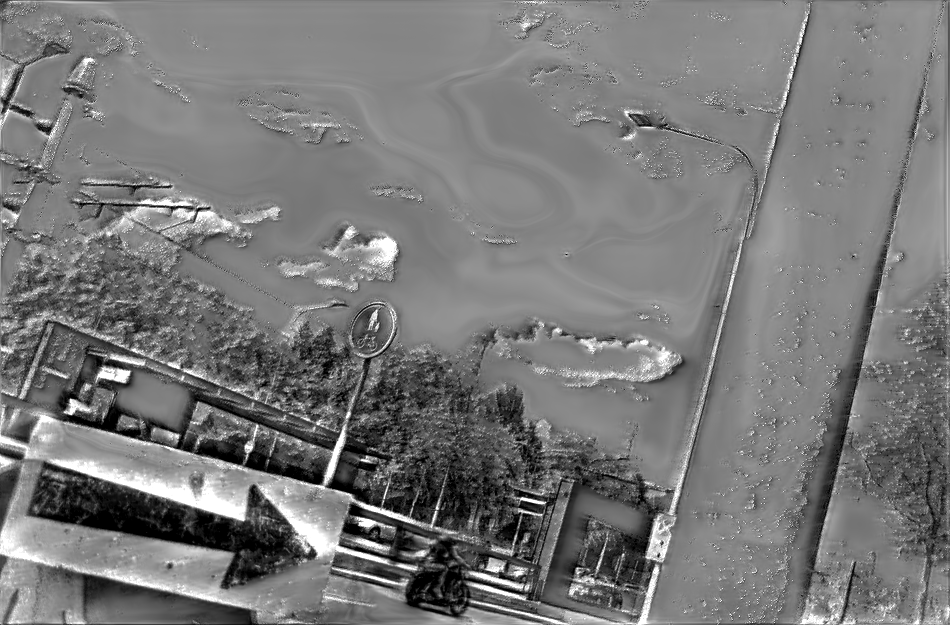}
        &\includegraphics[trim=10px 0px 10px 0px, clip, width=\linewidth]{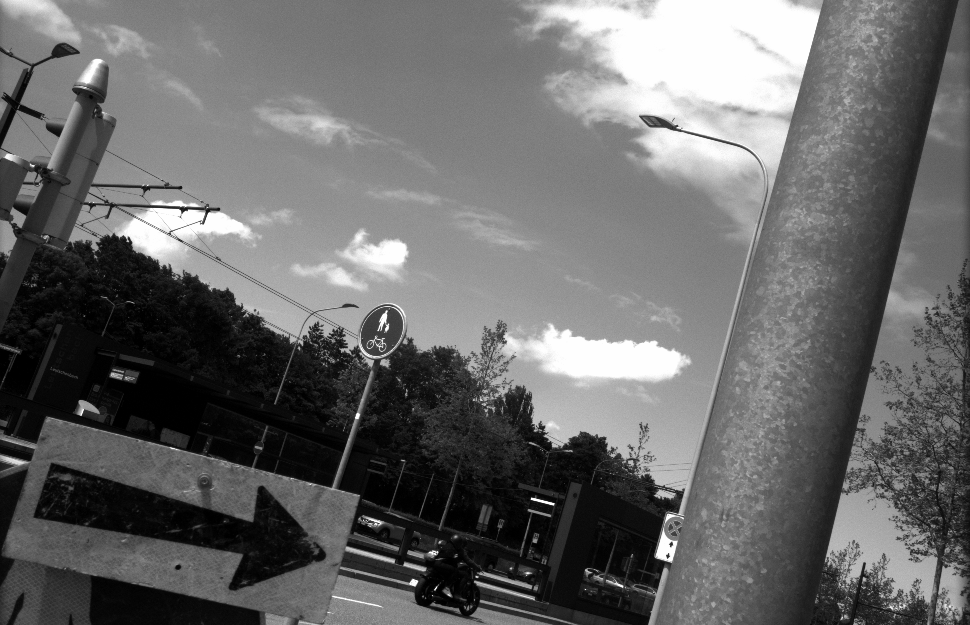}
	\\[-0.2ex]

        \rotatebox{90}{\makecell{rooftop\_01}}
        &\gframe{\includegraphics[width=\linewidth]{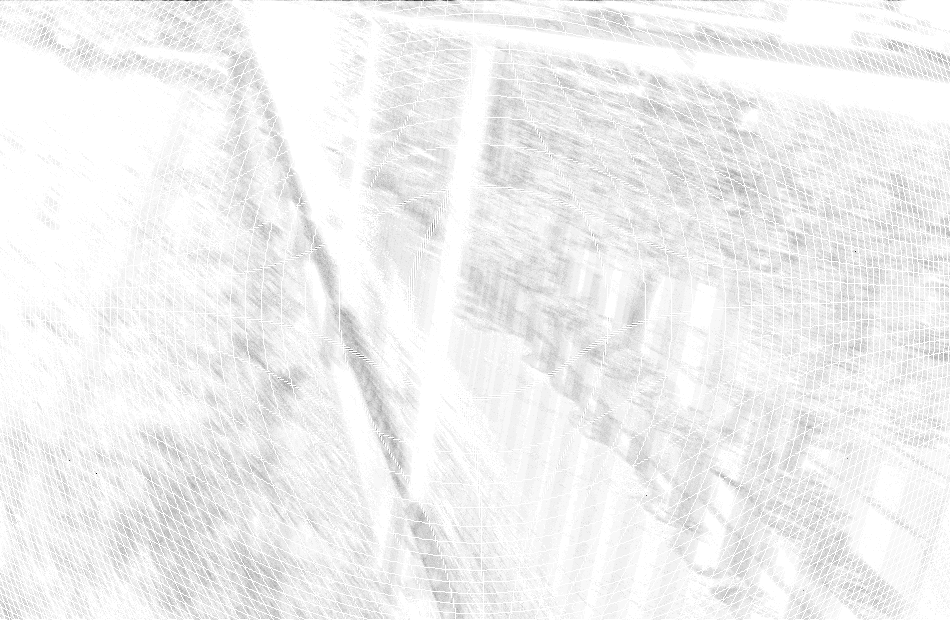}}
	&\gframe{\includegraphics[width=\linewidth]{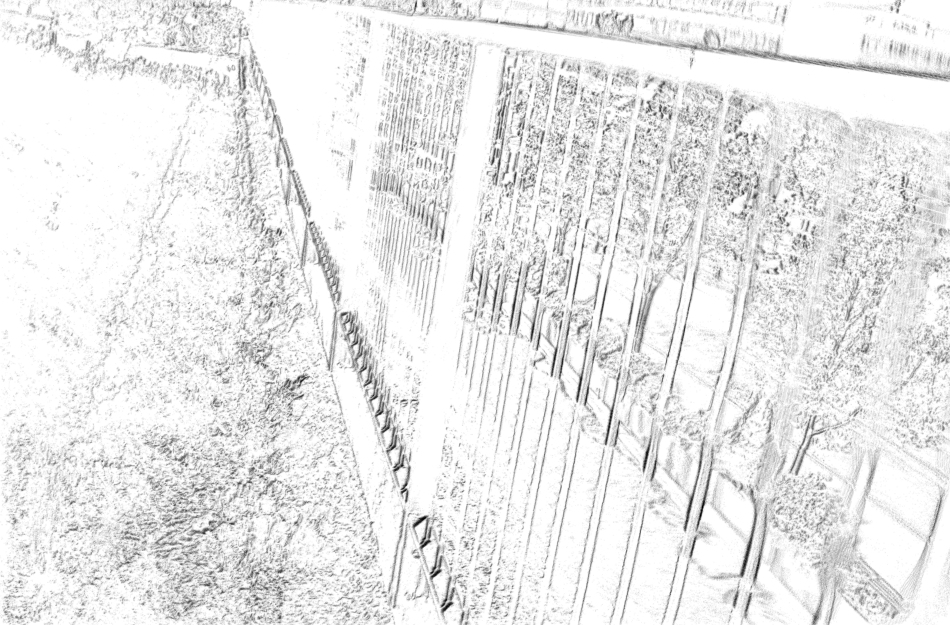}}
        &\includegraphics[width=\linewidth]{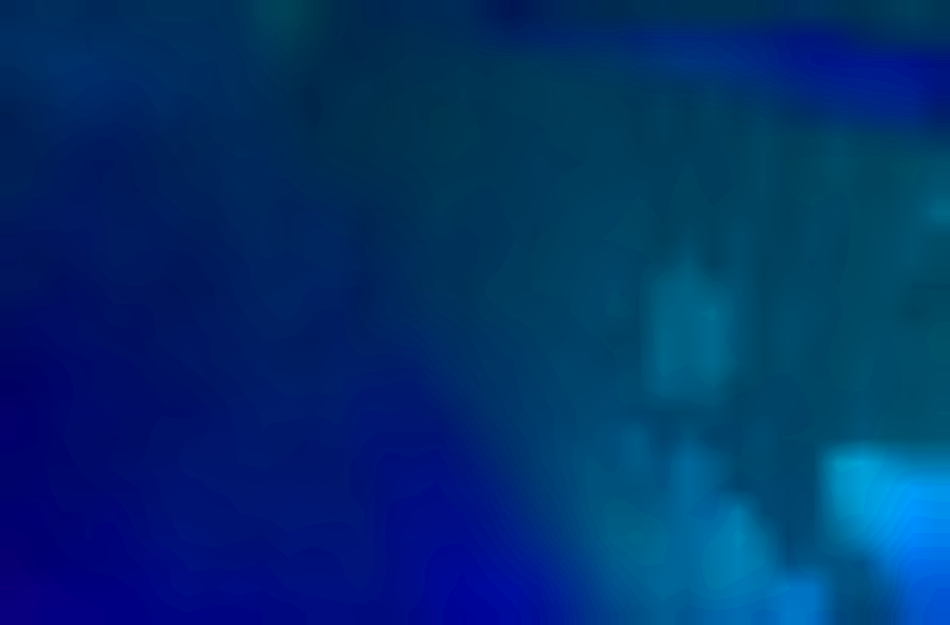}
	&\includegraphics[width=\linewidth]{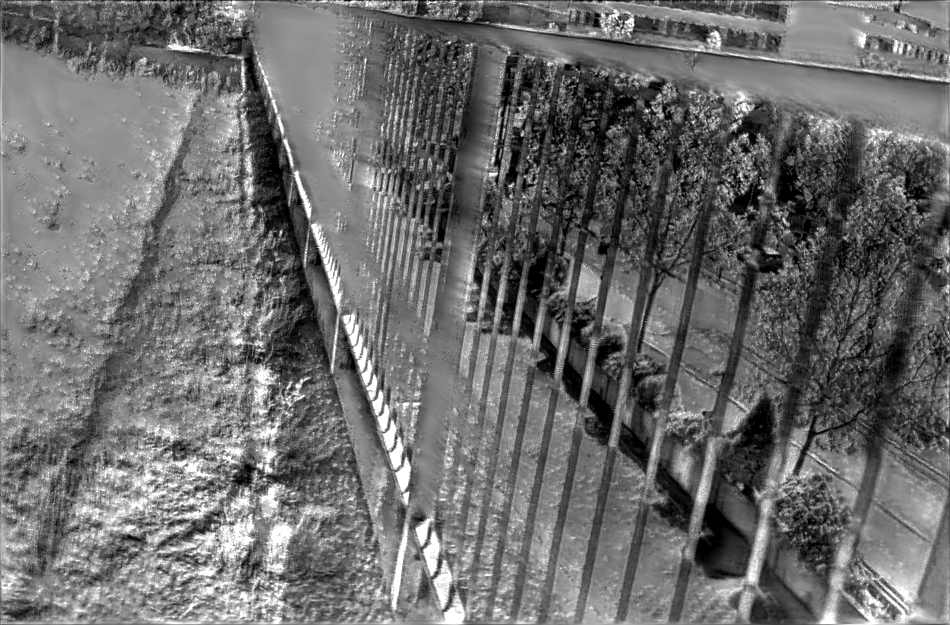}
        &\includegraphics[trim=10px 0px 10px 0px, clip, width=\linewidth]{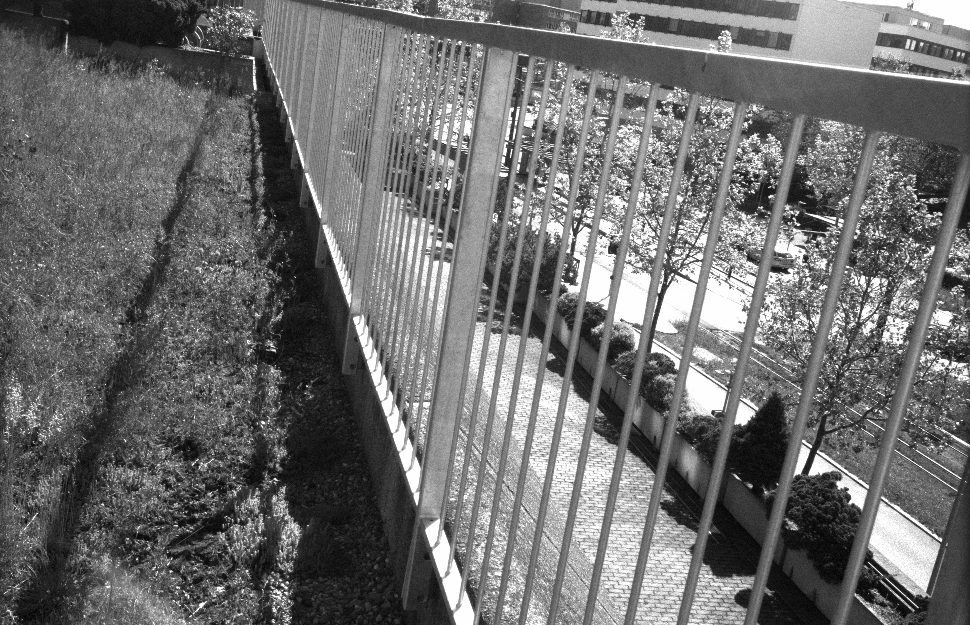}
	\\[-0.2ex]

        \rotatebox{90}{\makecell{rooftop\_02}}
        &\gframe{\includegraphics[width=\linewidth]{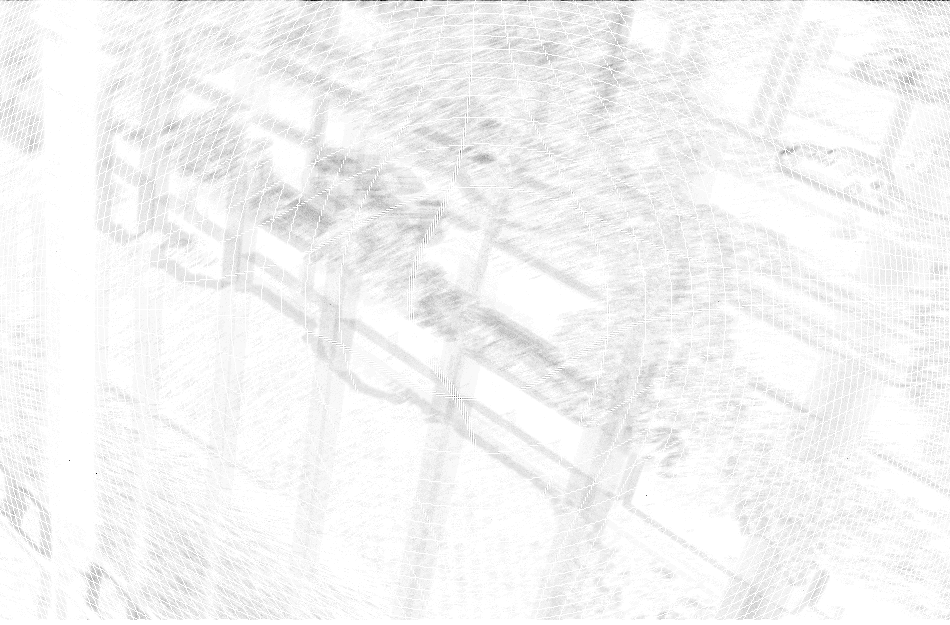}}
	&\gframe{\includegraphics[width=\linewidth]{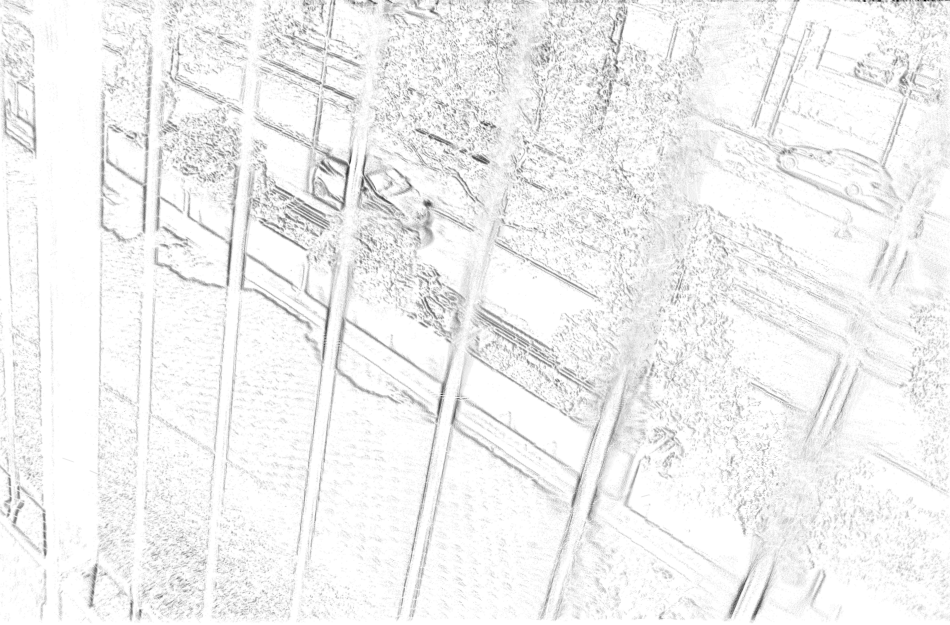}}
        &\includegraphics[width=\linewidth]{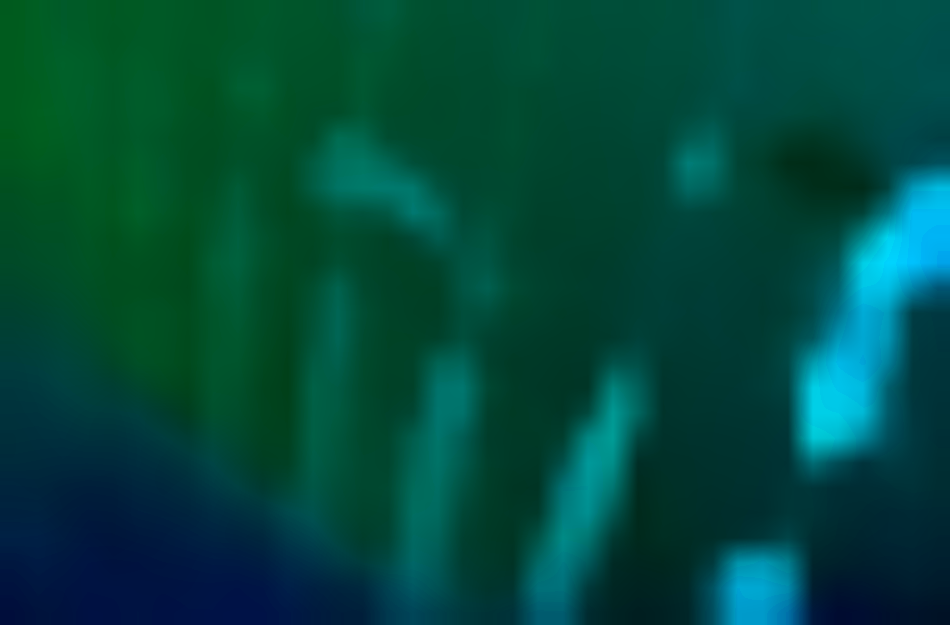}
	&\includegraphics[width=\linewidth]{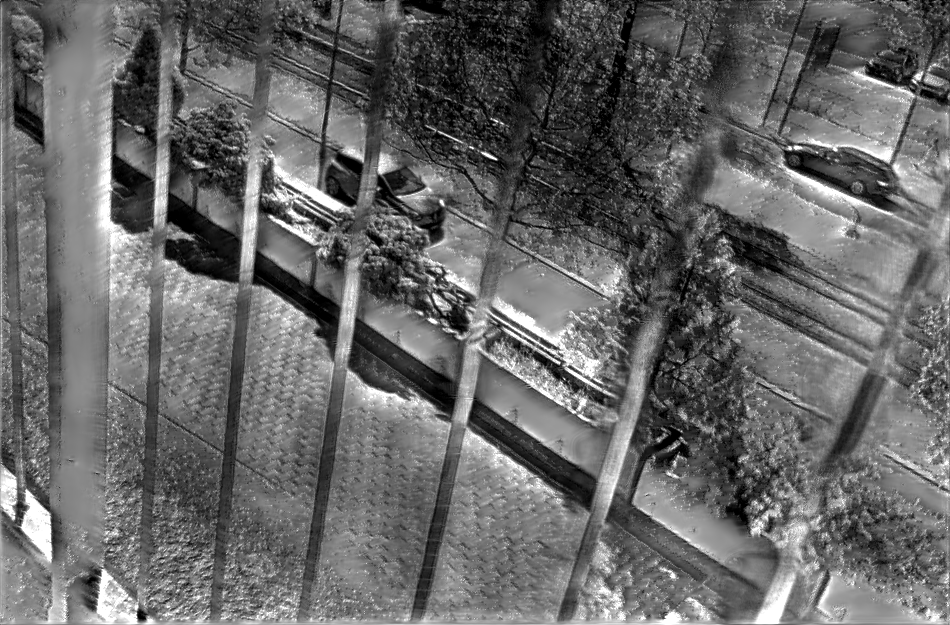}
        &\includegraphics[trim=10px 0px 10px 0px, clip, width=\linewidth]{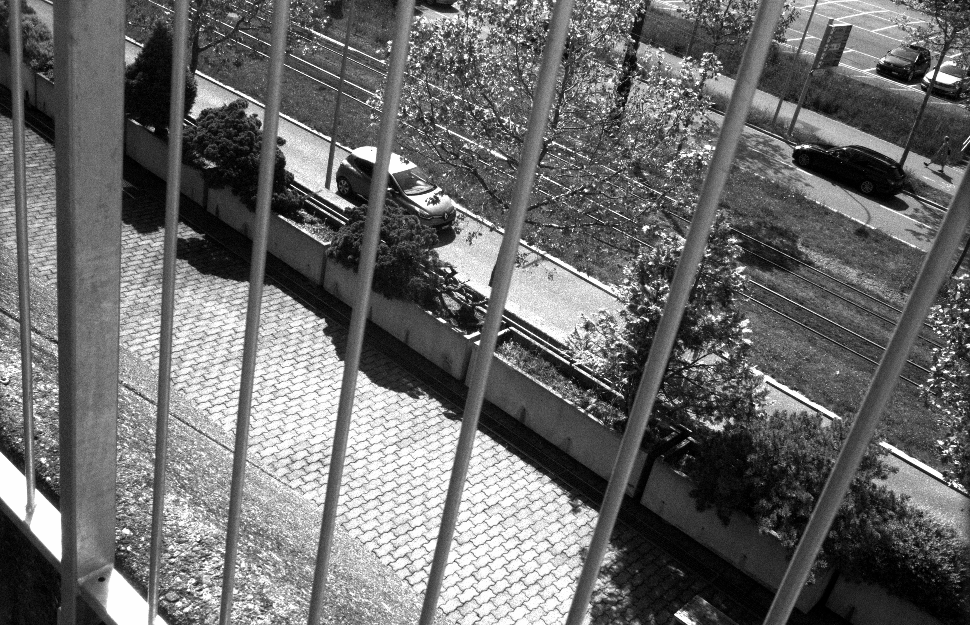}
	\\[-0.2ex]

        \rotatebox{90}{\makecell{rooftop\_02}}
        &\gframe{\includegraphics[width=\linewidth]{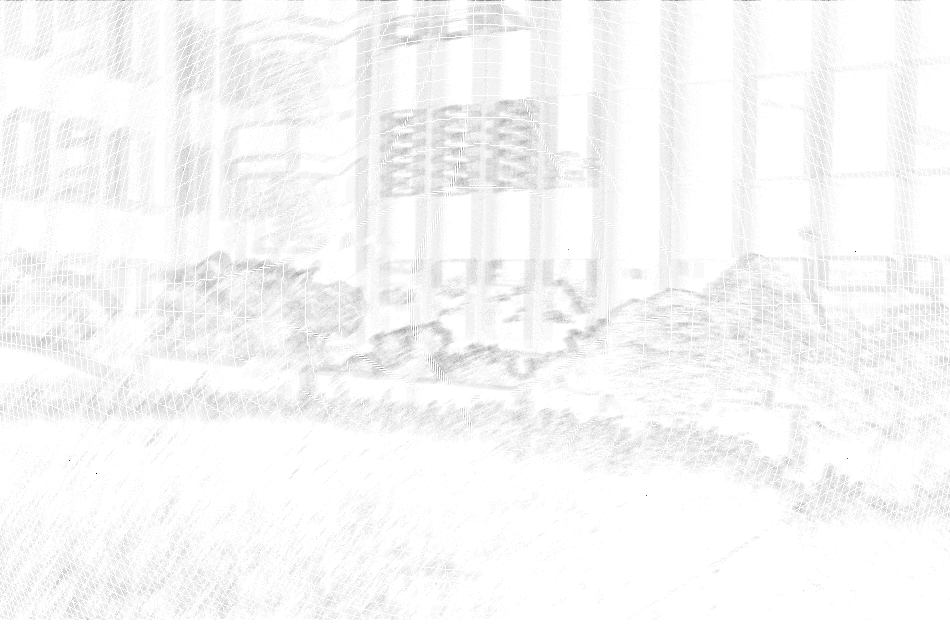}}
	&\gframe{\includegraphics[width=\linewidth]{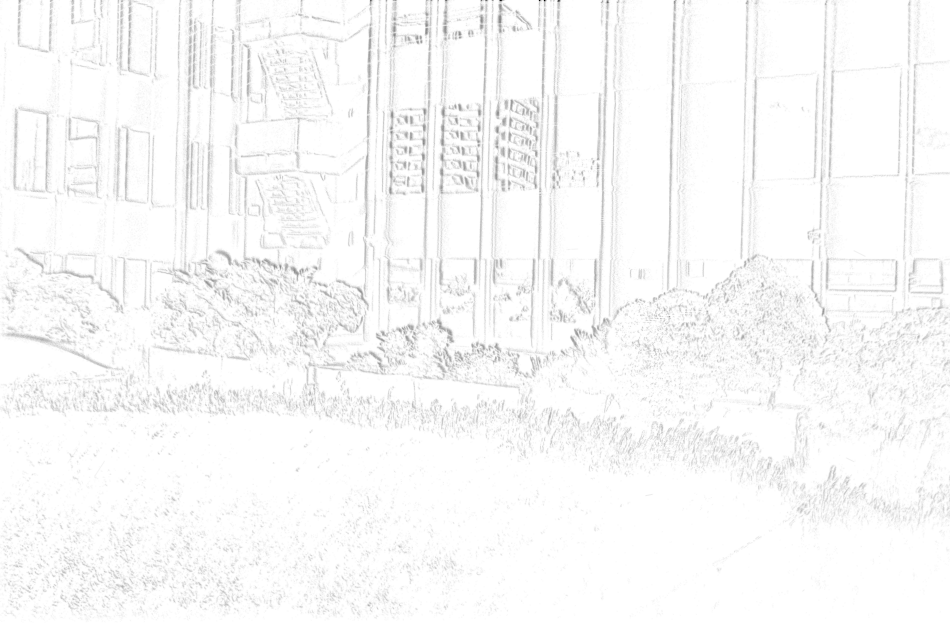}}
        &\includegraphics[width=\linewidth]{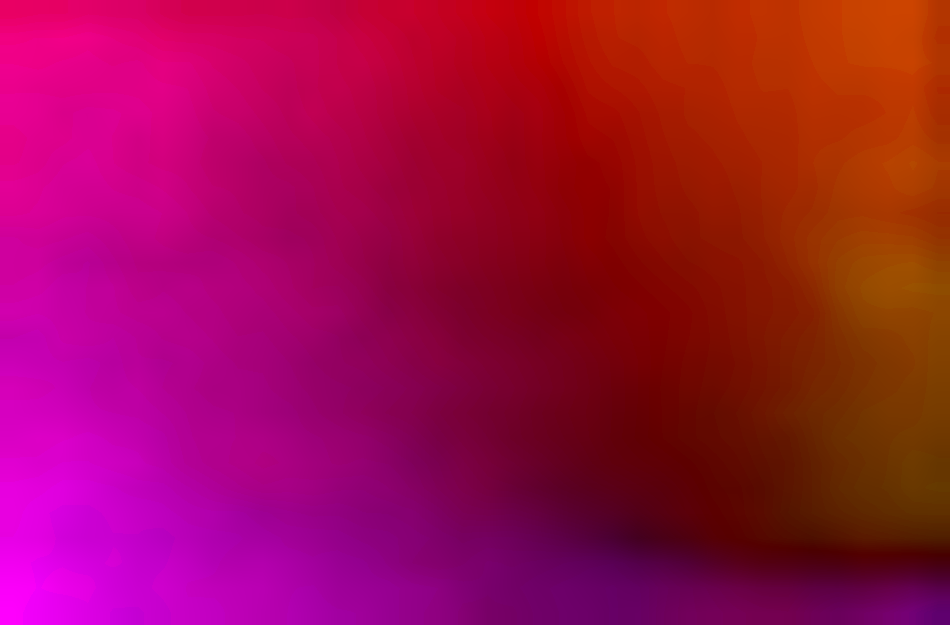}
	&\includegraphics[width=\linewidth]{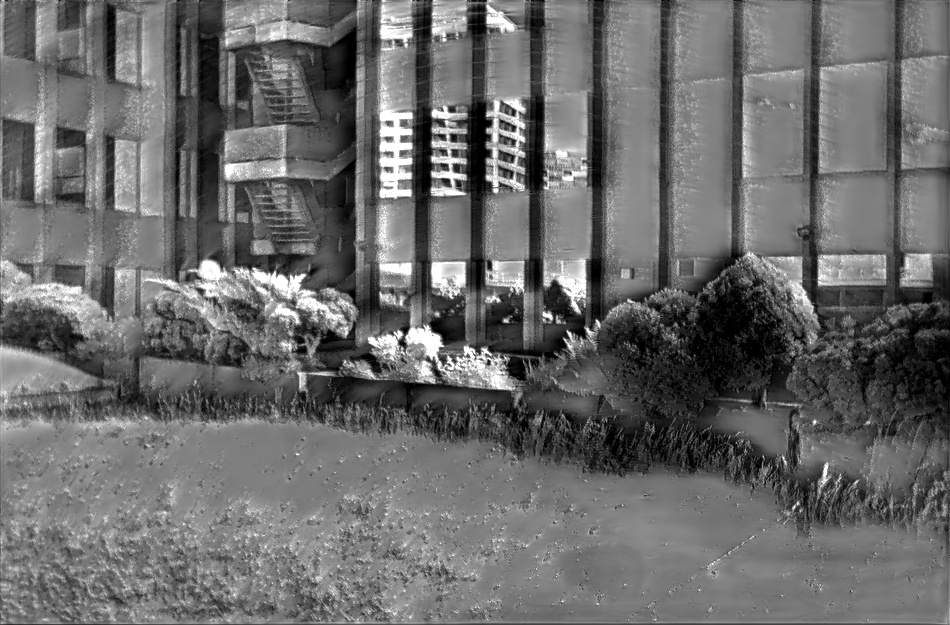}
        &\includegraphics[trim=10px 0px 10px 0px, clip, width=\linewidth]{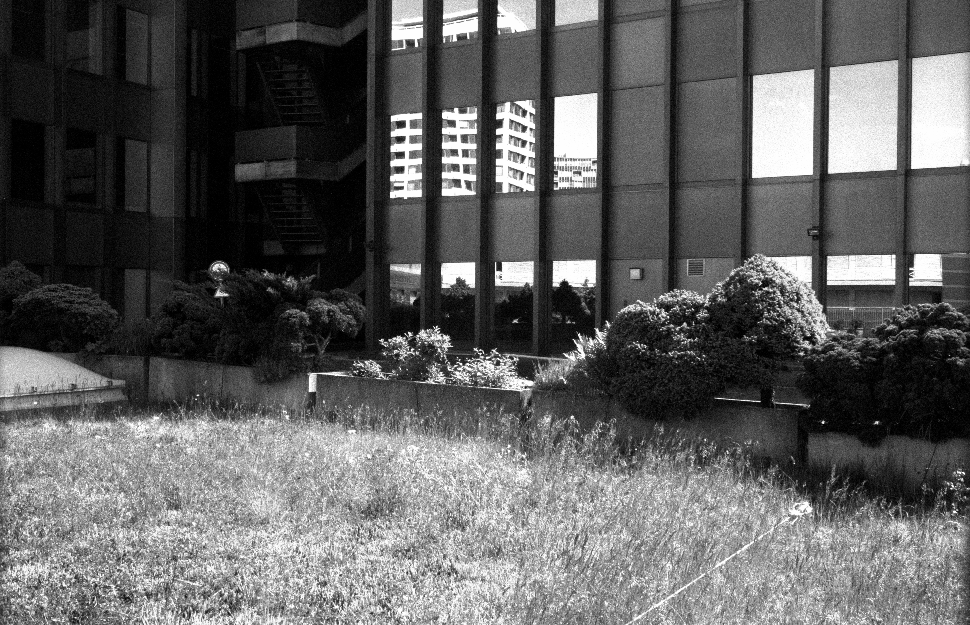}
	\\[-0.2ex]

        \rotatebox{90}{\makecell{rooftop\_03}}
        &\gframe{\includegraphics[width=\linewidth]{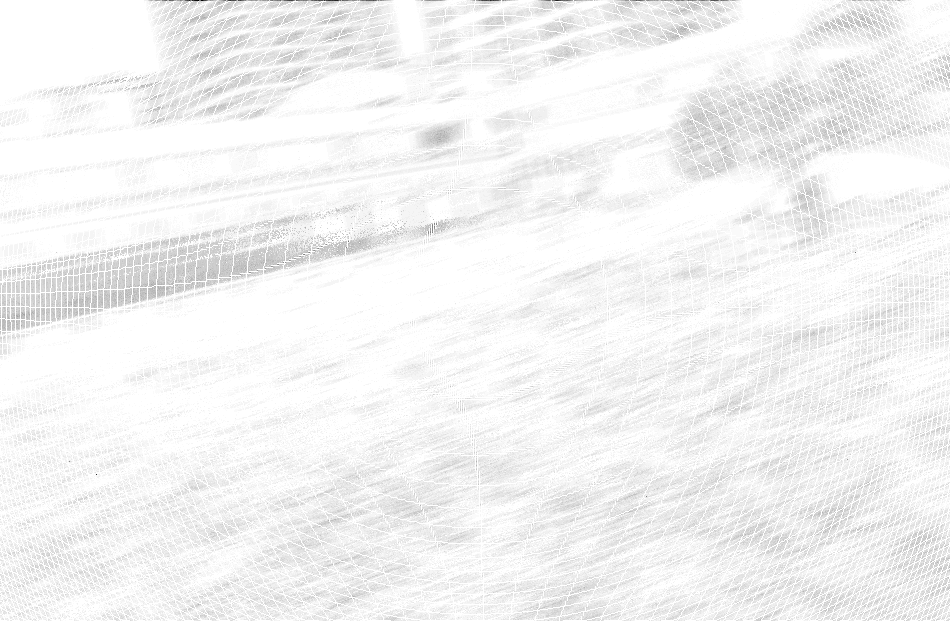}}
	&\gframe{\includegraphics[width=\linewidth]{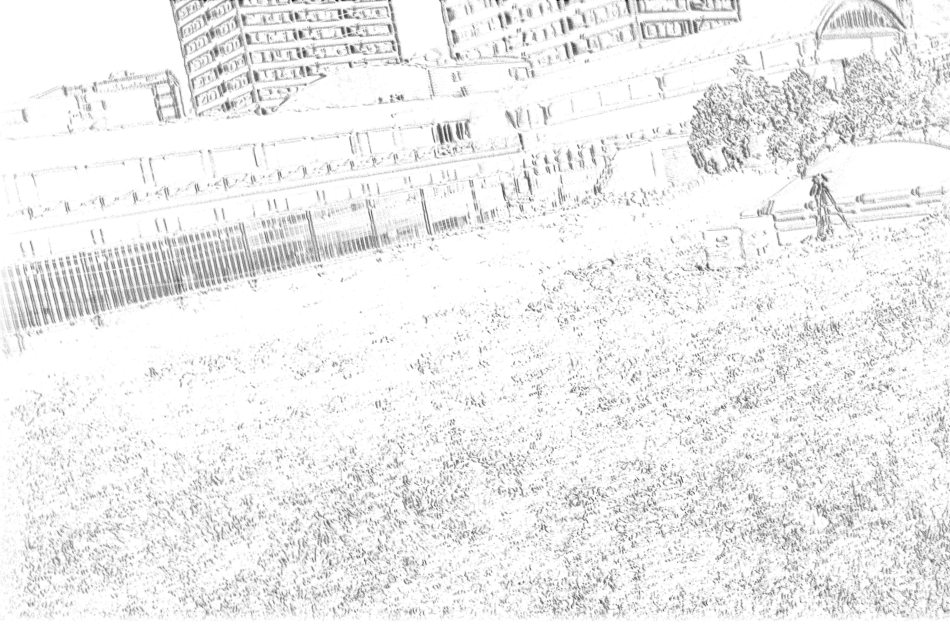}}
        &\includegraphics[width=\linewidth]{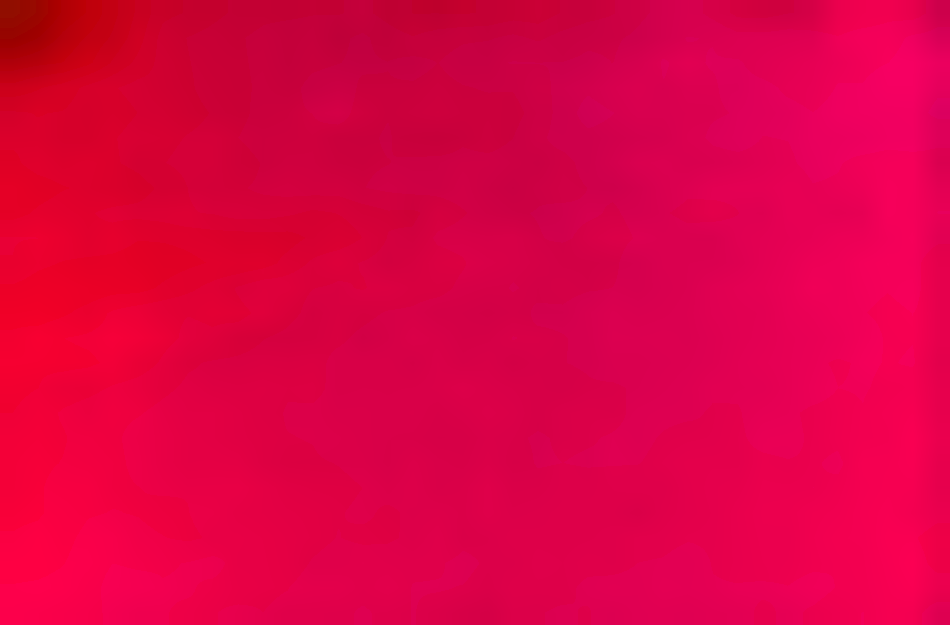}
	&\includegraphics[width=\linewidth]{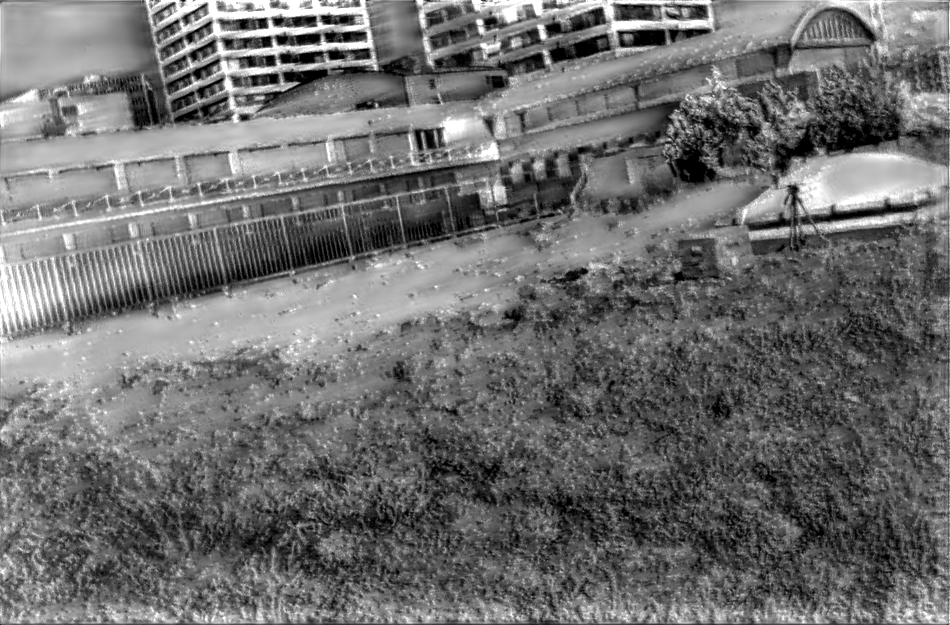}
        &\includegraphics[trim=10px 0px 10px 0px, clip, width=\linewidth]{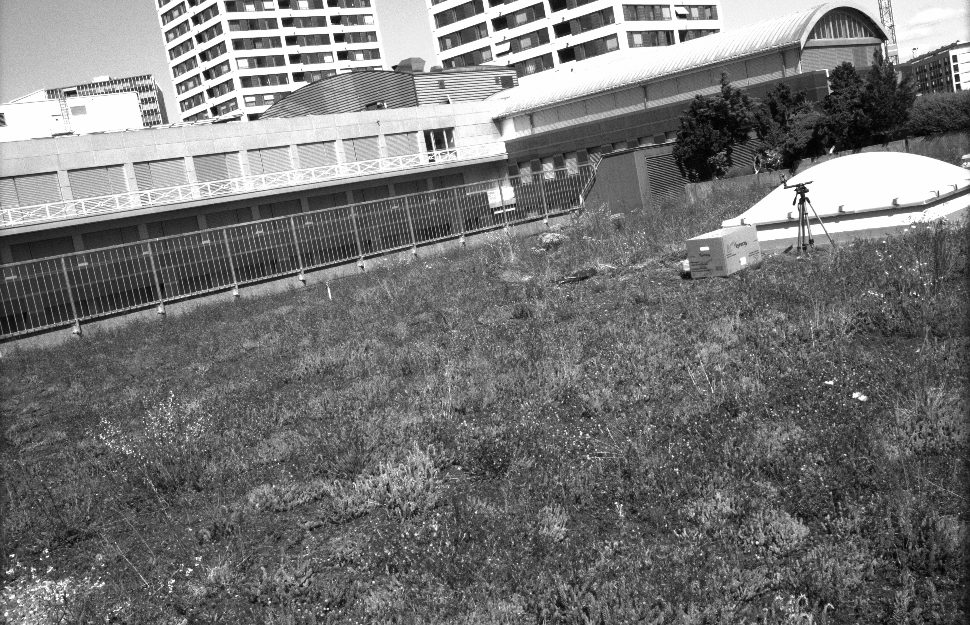}
	\\[-0.2ex]

        \rotatebox{90}{\makecell{rooftop\_03}}
        &\gframe{\includegraphics[width=\linewidth]{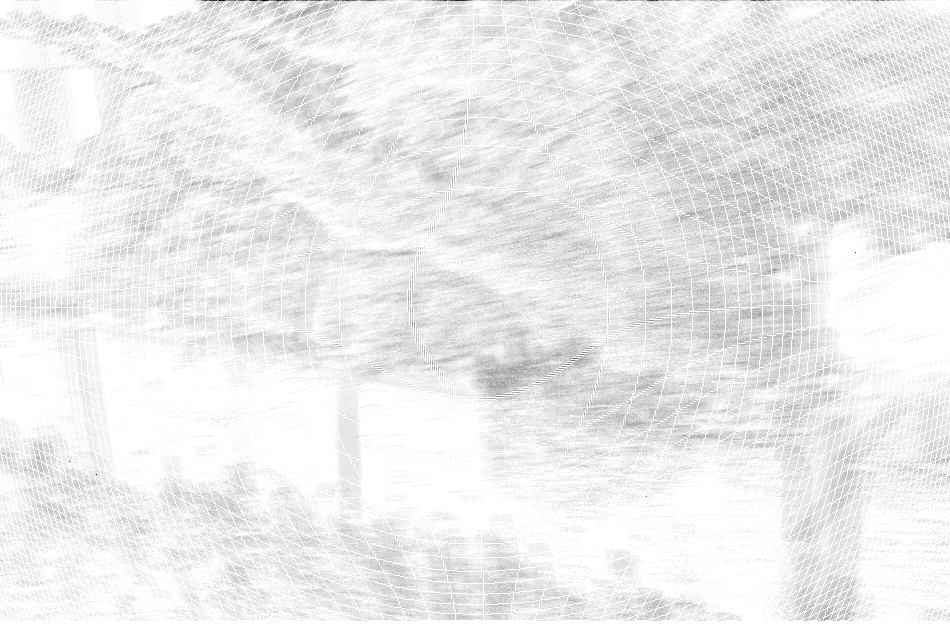}}
	&\gframe{\includegraphics[width=\linewidth]{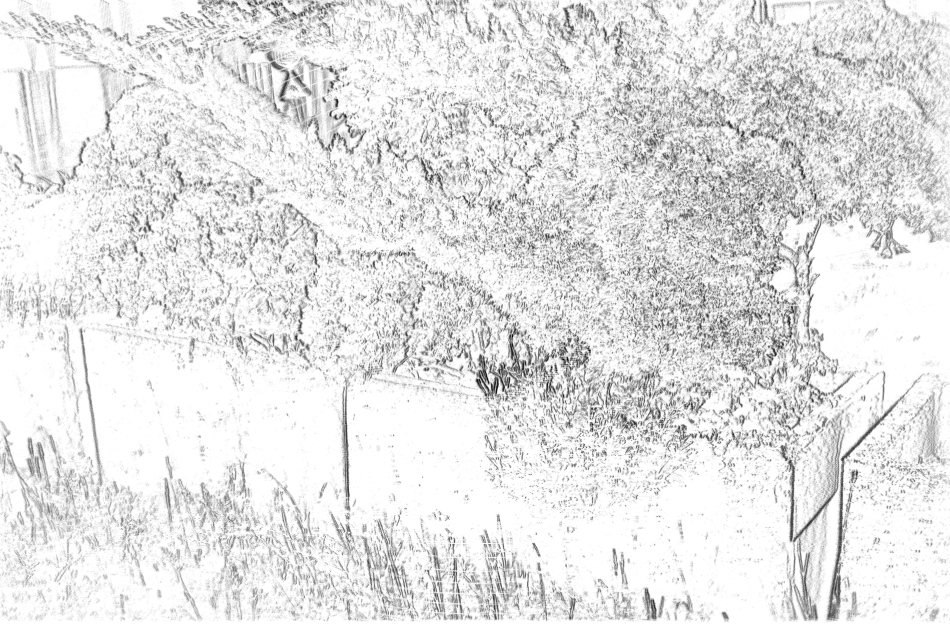}}
        &\includegraphics[width=\linewidth]{images/suppl/may29_rooftop_handheld_03/008/flow.png}
	&\includegraphics[width=\linewidth]{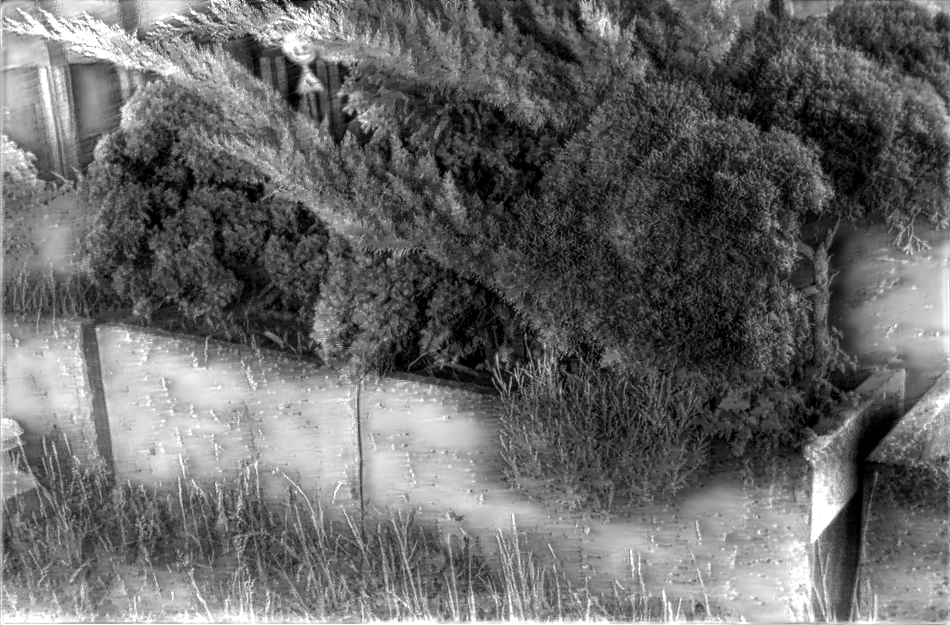}
        &\includegraphics[trim=10px 0px 10px 0px, clip, width=\linewidth]{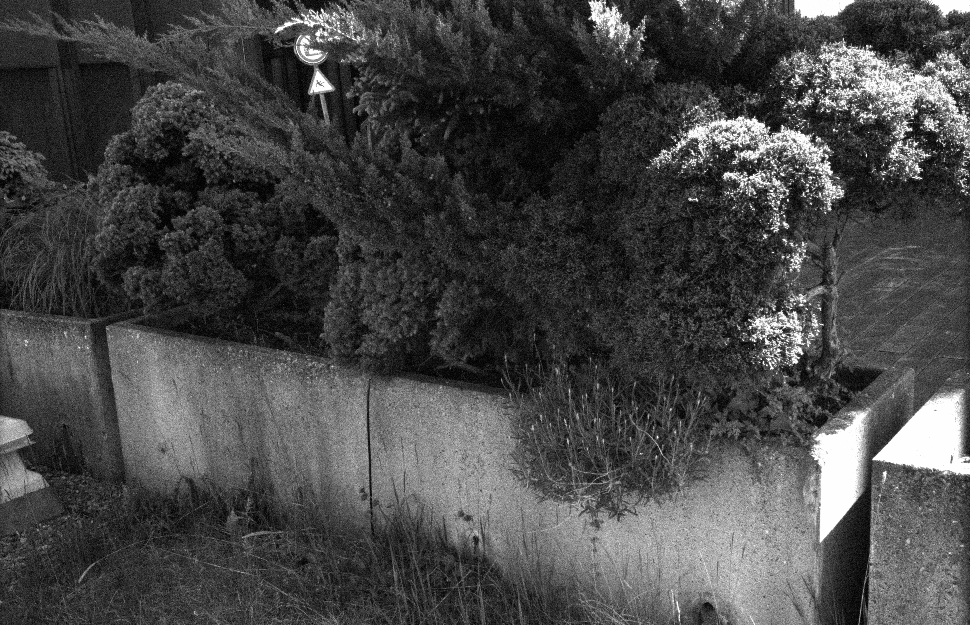}
	\\[-0.2ex]

        \rotatebox{90}{\makecell{rooftop\_05}}
        &\gframe{\includegraphics[width=\linewidth]{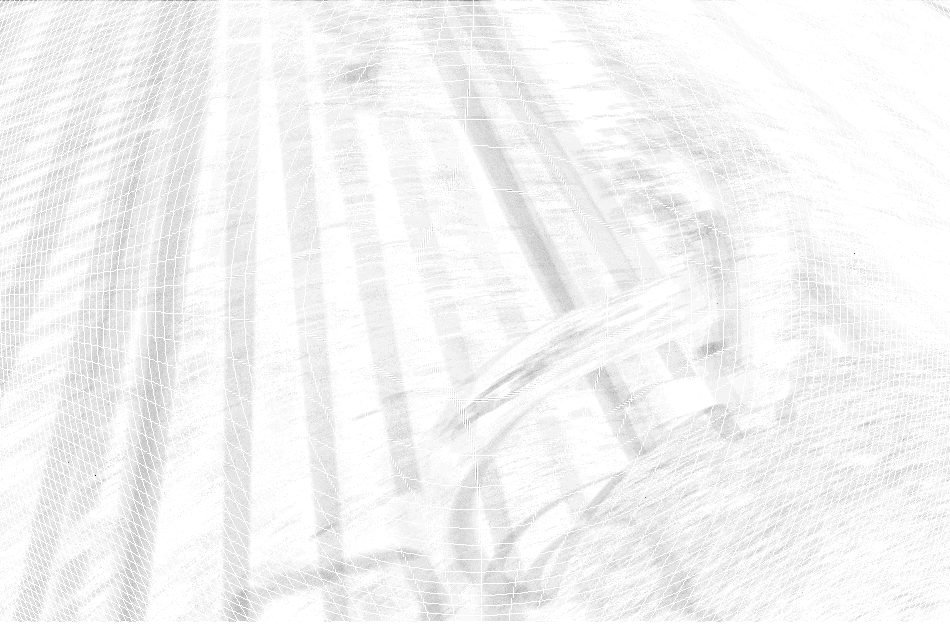}}
	&\gframe{\includegraphics[width=\linewidth]{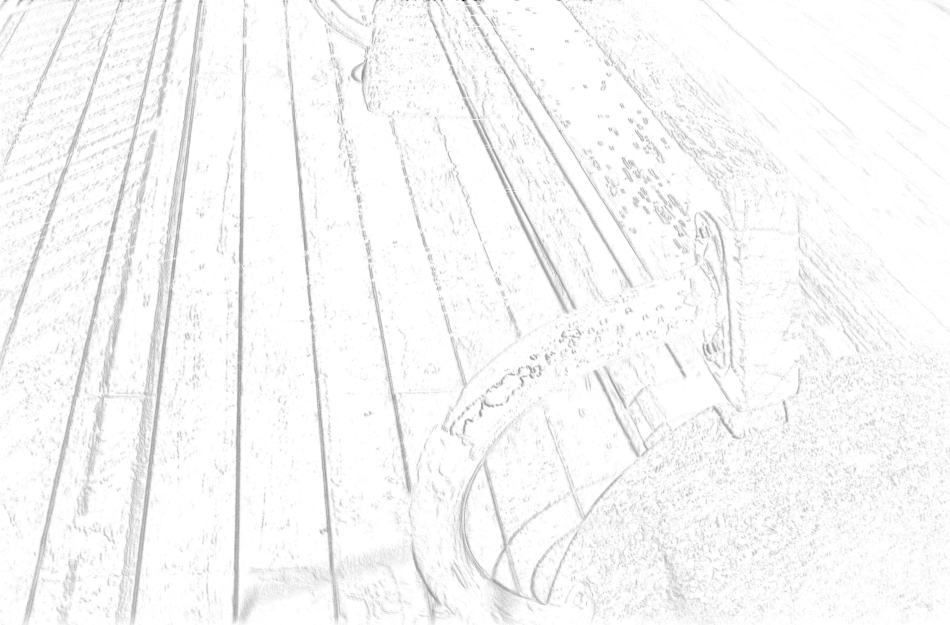}}
        &\includegraphics[width=\linewidth]{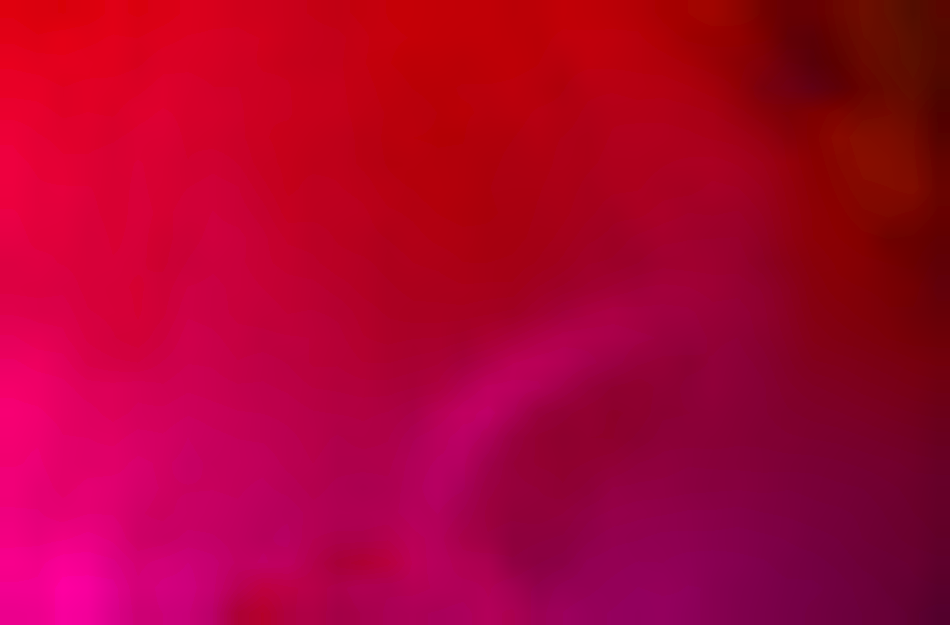}
	&\includegraphics[width=\linewidth]{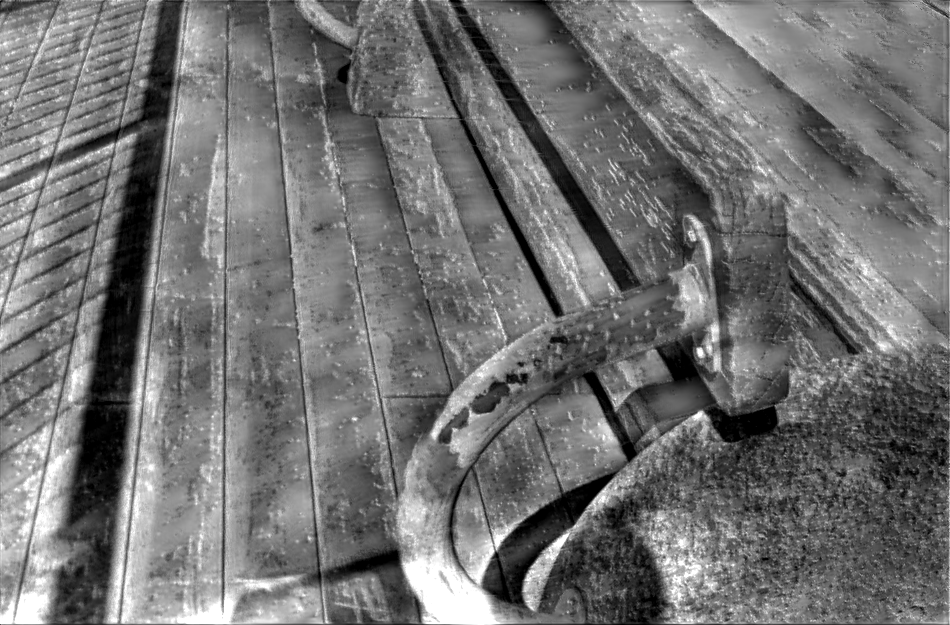}
        &\includegraphics[trim=10px 0px 10px 0px, clip, width=\linewidth]{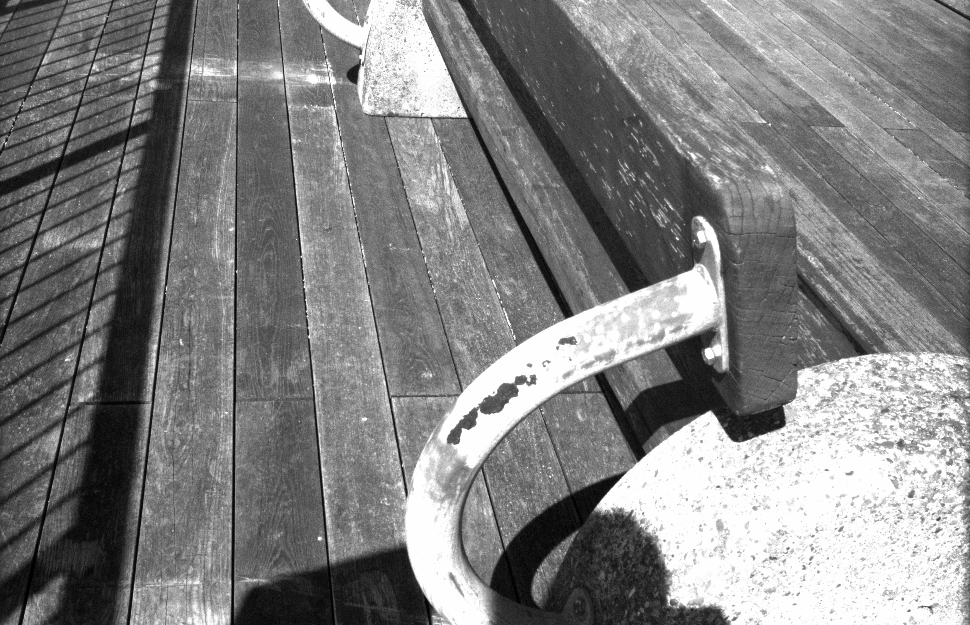}
	\\[-0.2ex]

        & (a) Events
        & (b) IWEs
        & (c) Flow
        & (d) Intensity
        & (e) Reference image
        \\
	\end{tabular}
	}
    \caption{\emph{Additional qualitative results on the BS-ERGB dataset}. From left to right: (a) input events; (b) image of warped events (IWE) with our predicted flow; (c) our predicted flow; (d) our predicted intensity; (e) reference image.
    \label{fig:suppl_results}}
\end{figure*}

\def\figWidth{0.135\linewidth}
\begin{figure*}[ht]
	\centering
    {\footnotesize
    \setlength{\tabcolsep}{1pt}
	\begin{tabular}{
        >{\centering\arraybackslash}m{0.3cm} 
	>{\centering\arraybackslash}m{\figWidth} 
	>{\centering\arraybackslash}m{\figWidth}
	>{\centering\arraybackslash}m{\figWidth}
	>{\centering\arraybackslash}m{\figWidth}
        >{\centering\arraybackslash}m{\figWidth}
        >{\centering\arraybackslash}m{\figWidth}
        >{\centering\arraybackslash}m{\figWidth}
        }

        \rotatebox{90}{\makecell{may29\_01}}
	&\includegraphics[trim=10px 0px 10px 0px, clip, width=\linewidth]{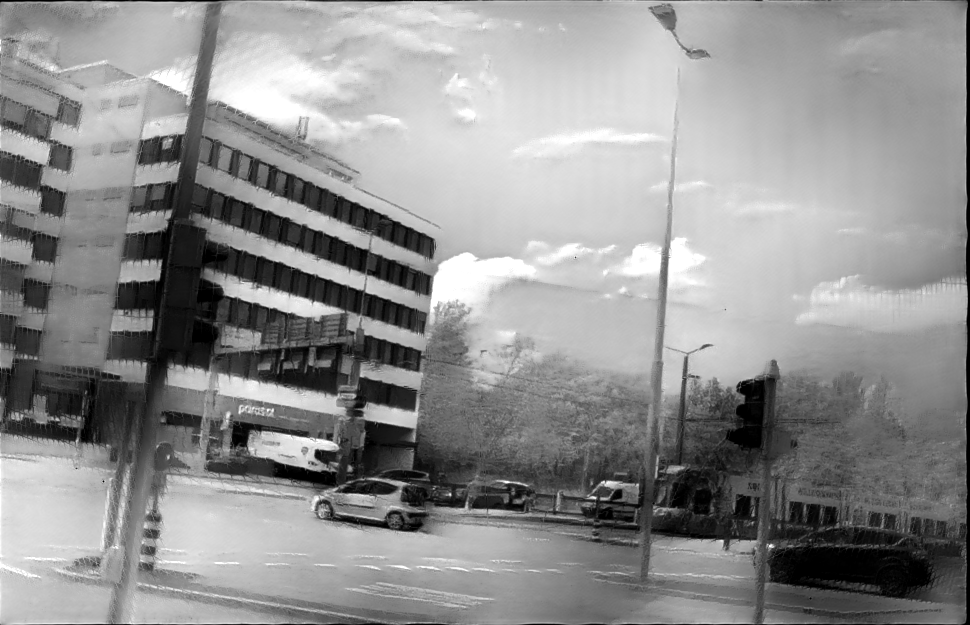}
	&\includegraphics[trim=10px 0px 10px 0px, clip, width=\linewidth]{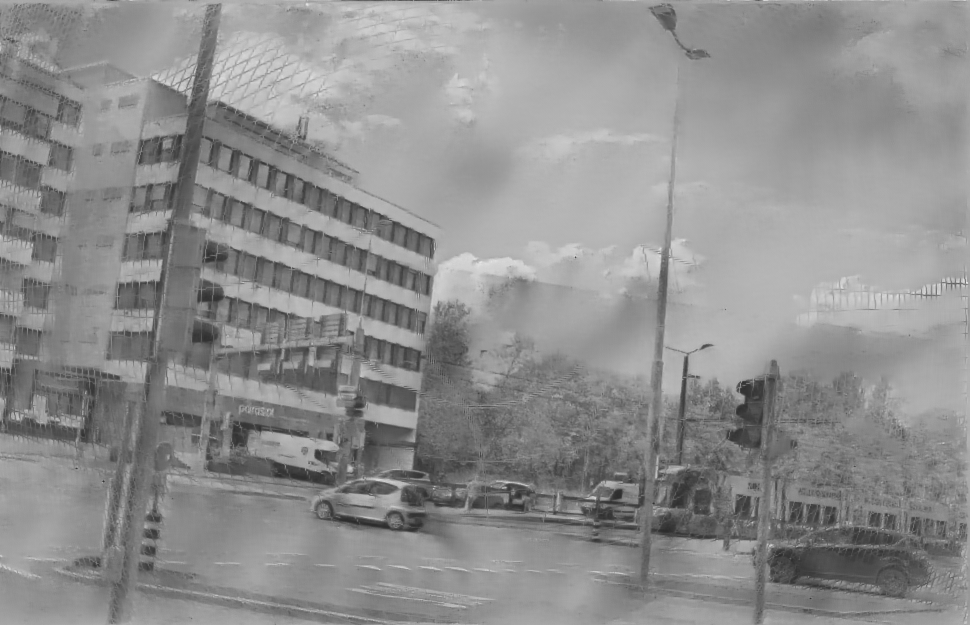}
        &\includegraphics[trim=10px 0px 10px 0px, clip, width=\linewidth]{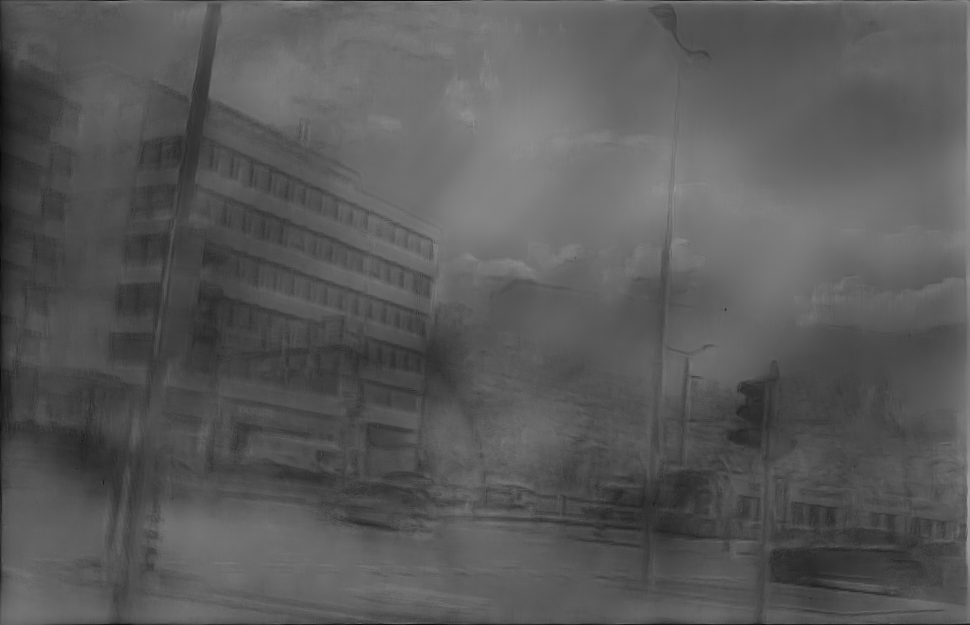}
        &\includegraphics[trim=10px 0px 10px 0px, clip, width=\linewidth]{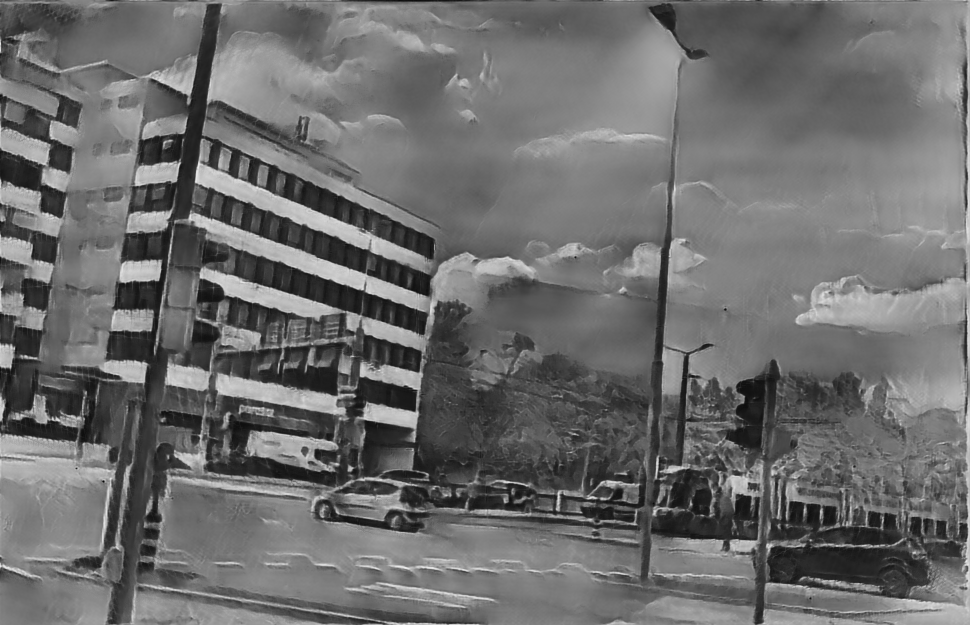}
        &\includegraphics[trim=10px 0px 10px 0px, clip, width=\linewidth]{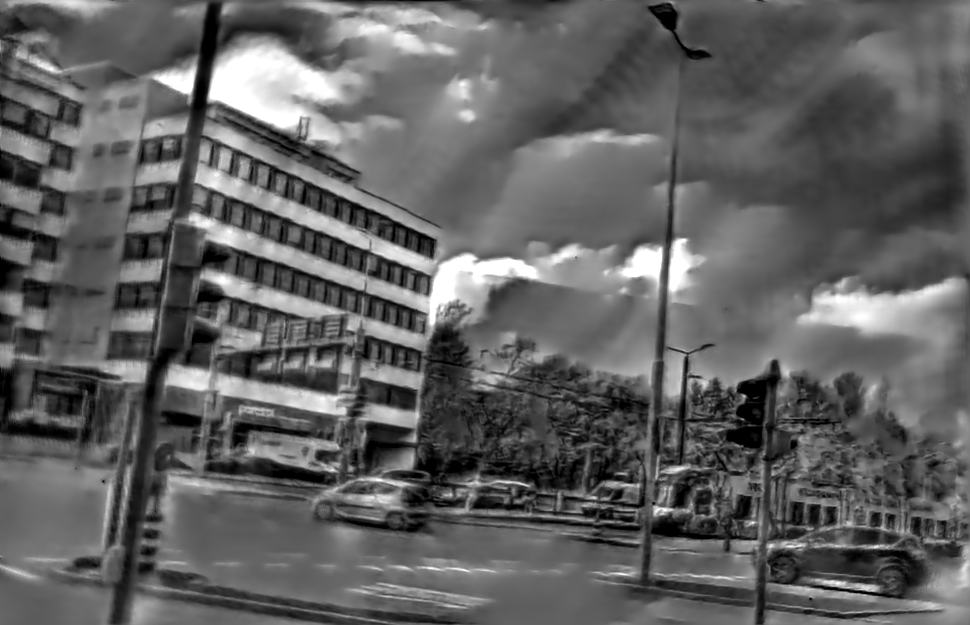}
	&\includegraphics[width=\linewidth]{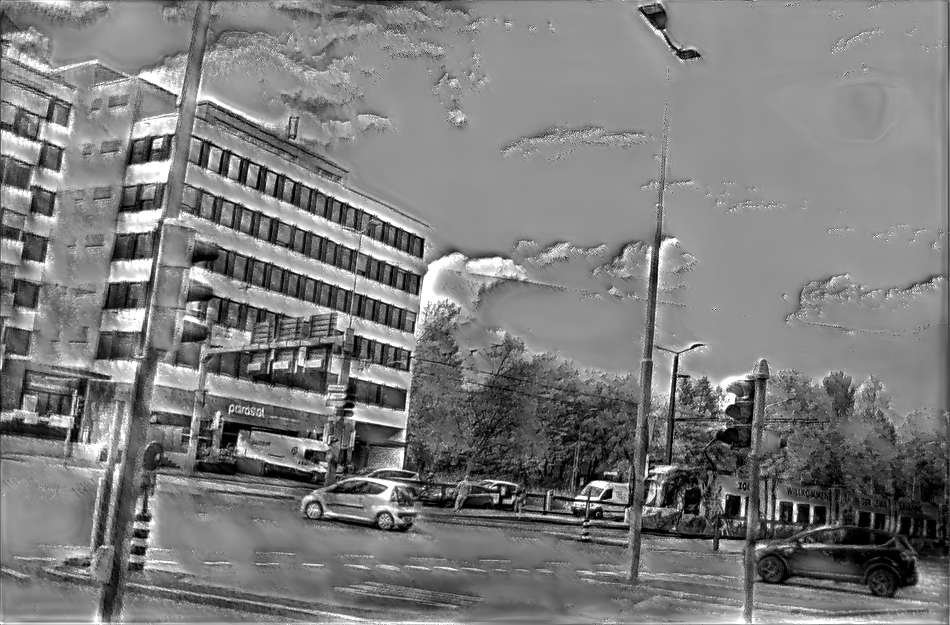}
        &\includegraphics[trim=10px 0px 10px 0px, clip, width=\linewidth]{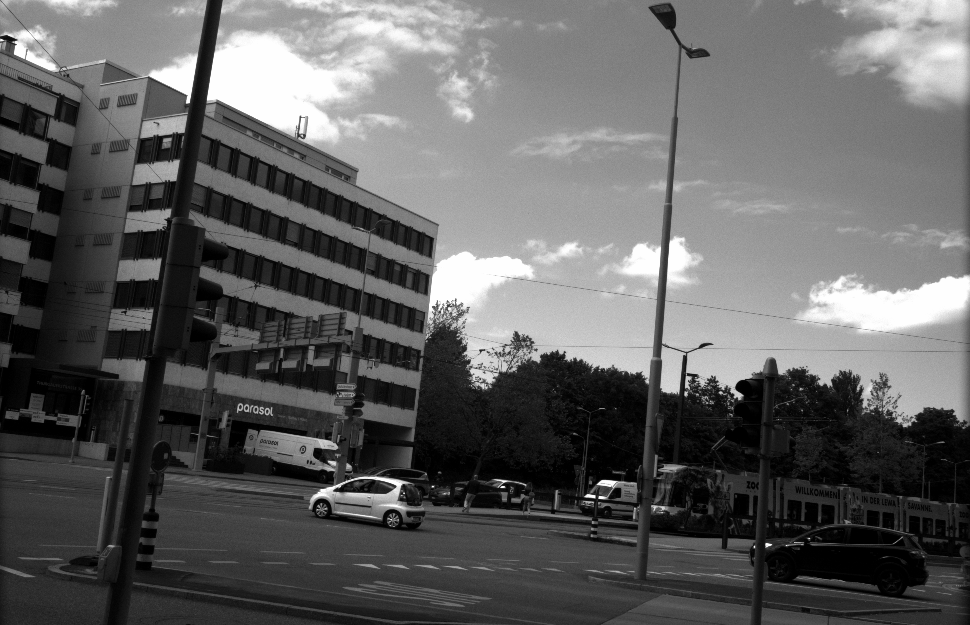}
	\\[-0.2ex]

        \rotatebox{90}{\makecell{may29\_03}}
	&\includegraphics[trim=10px 0px 10px 0px, clip, width=\linewidth]{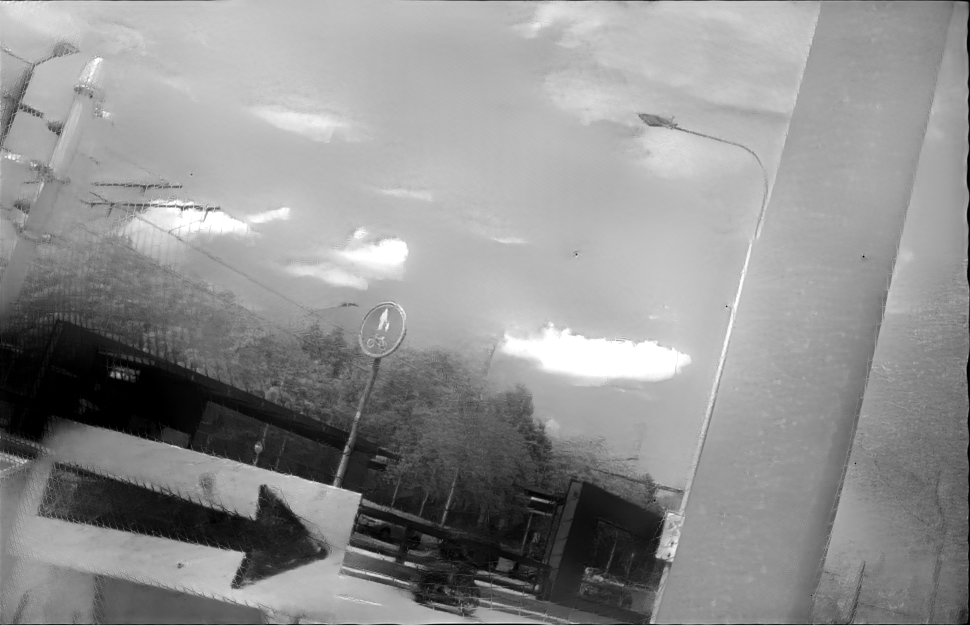}
	&\includegraphics[trim=10px 0px 10px 0px, clip, width=\linewidth]{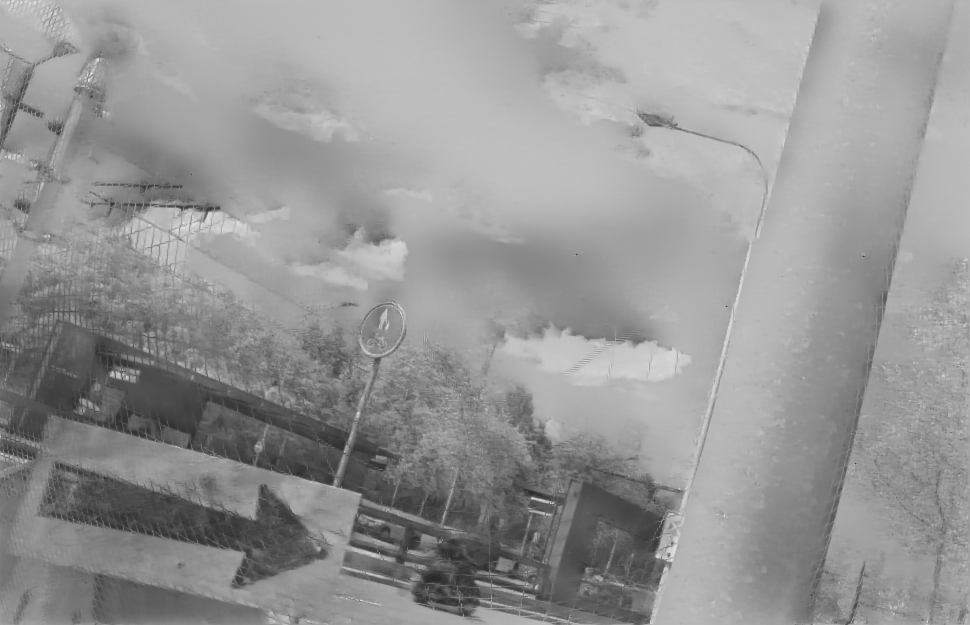}
        &\includegraphics[trim=10px 0px 10px 0px, clip, width=\linewidth]{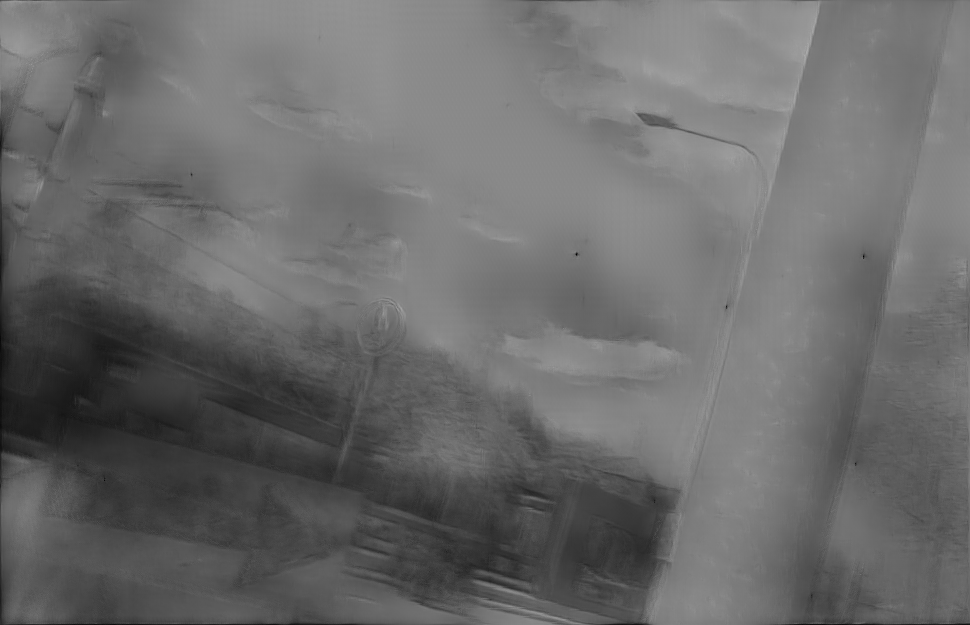}
        &\includegraphics[trim=10px 0px 10px 0px, clip, width=\linewidth]{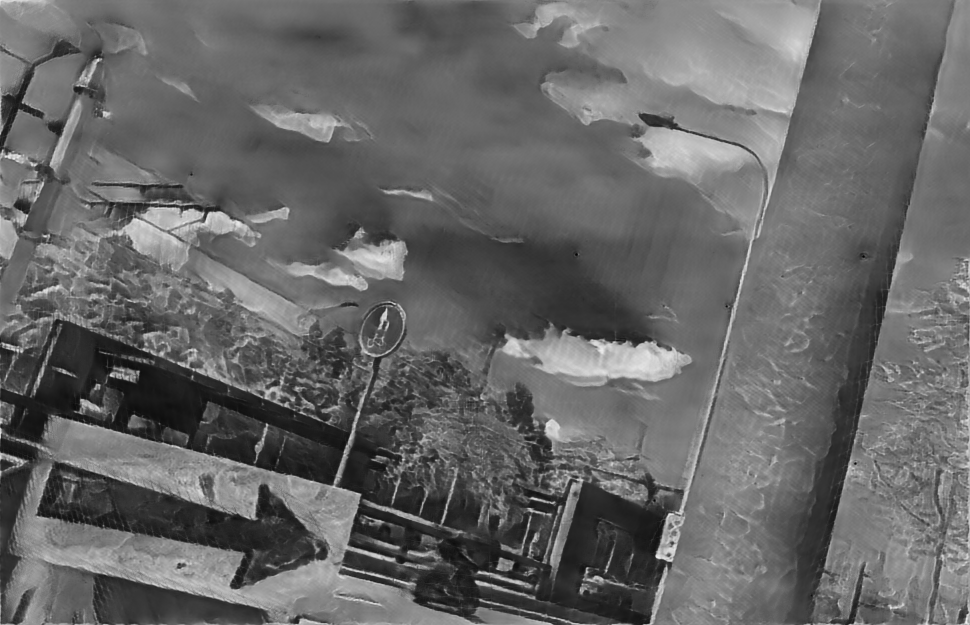}
        &\includegraphics[trim=10px 0px 10px 0px, clip, width=\linewidth]{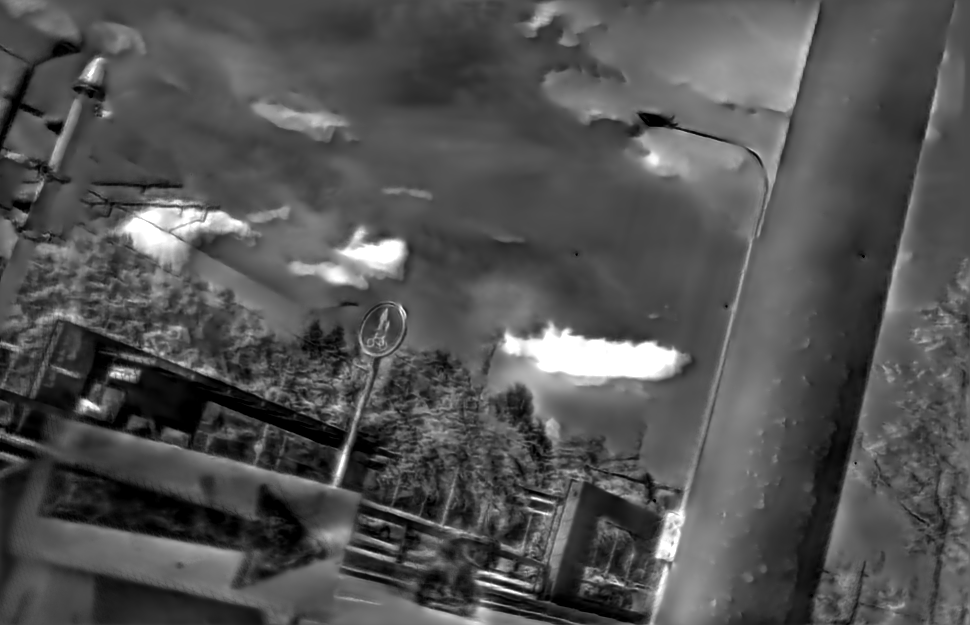}
	&\includegraphics[width=\linewidth]{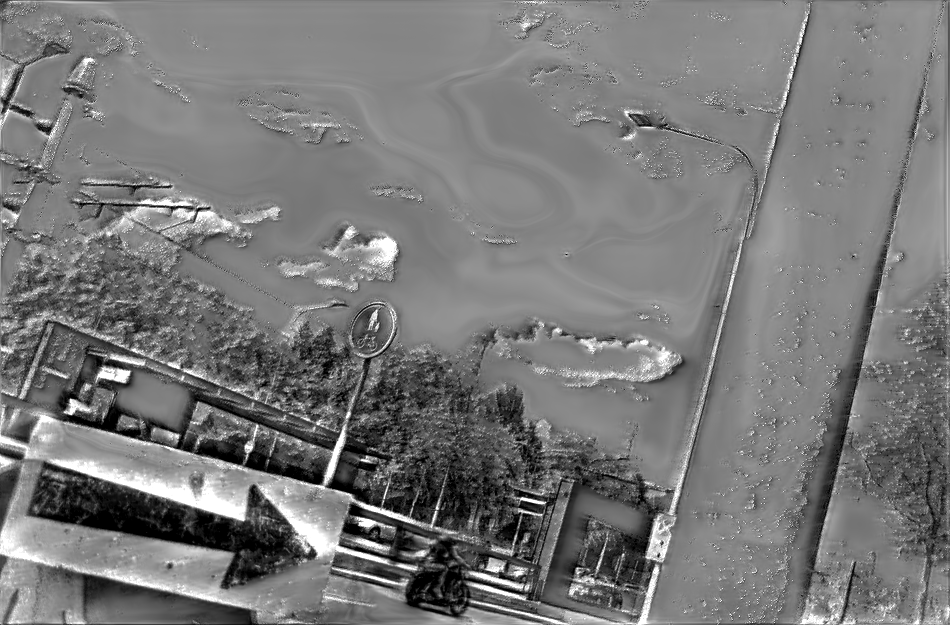}
        &\includegraphics[trim=10px 0px 10px 0px, clip, width=\linewidth]{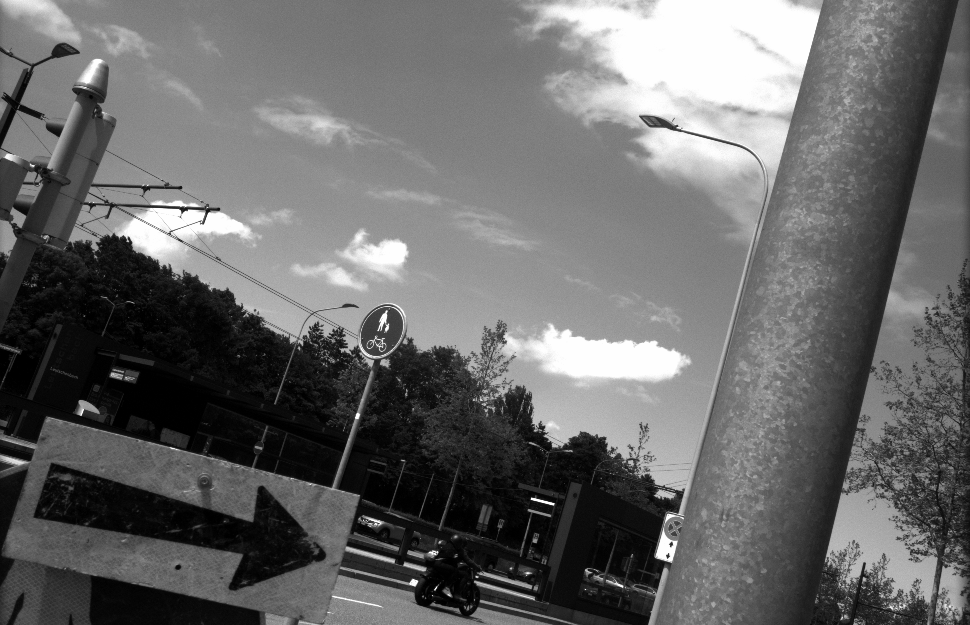}
	\\[-0.2ex]

        \rotatebox{90}{\makecell{rooftop\_01}}
	&\includegraphics[trim=10px 0px 10px 0px, clip, width=\linewidth]{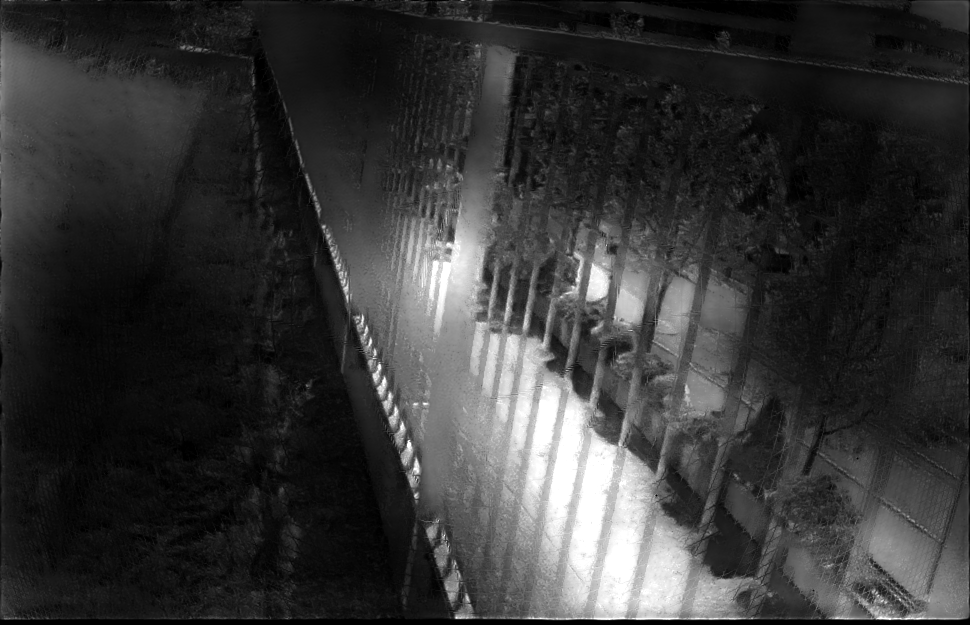}
	&\includegraphics[trim=10px 0px 10px 0px, clip, width=\linewidth]{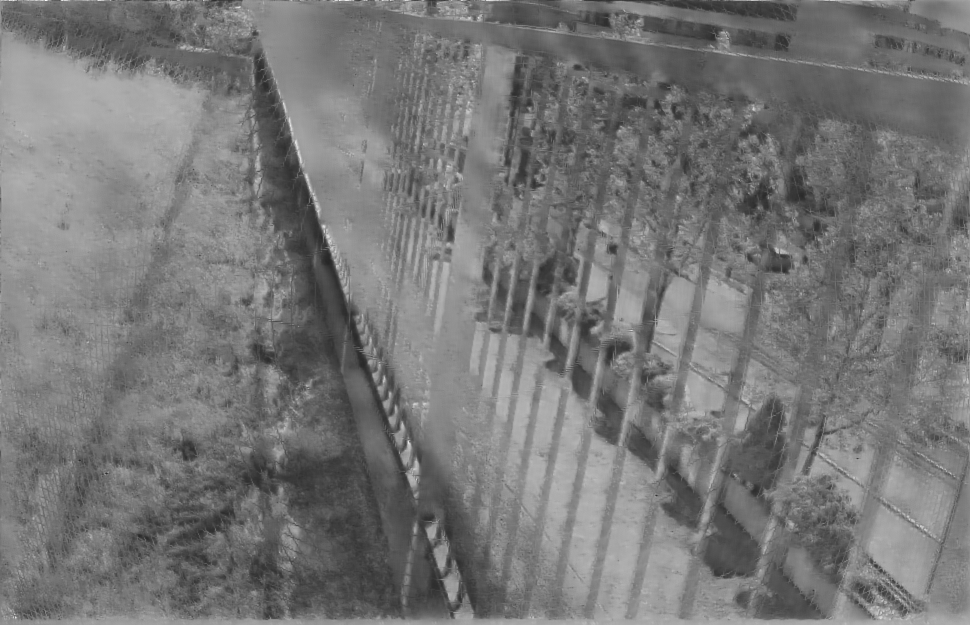}
        &\includegraphics[trim=10px 0px 10px 0px, clip, width=\linewidth]{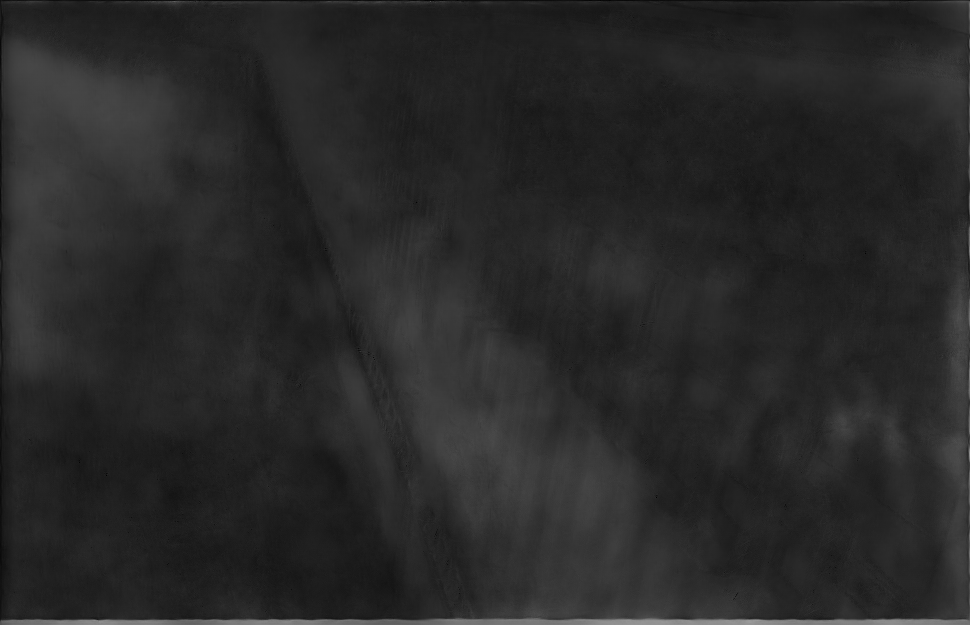}
        &\includegraphics[trim=10px 0px 10px 0px, clip, width=\linewidth]{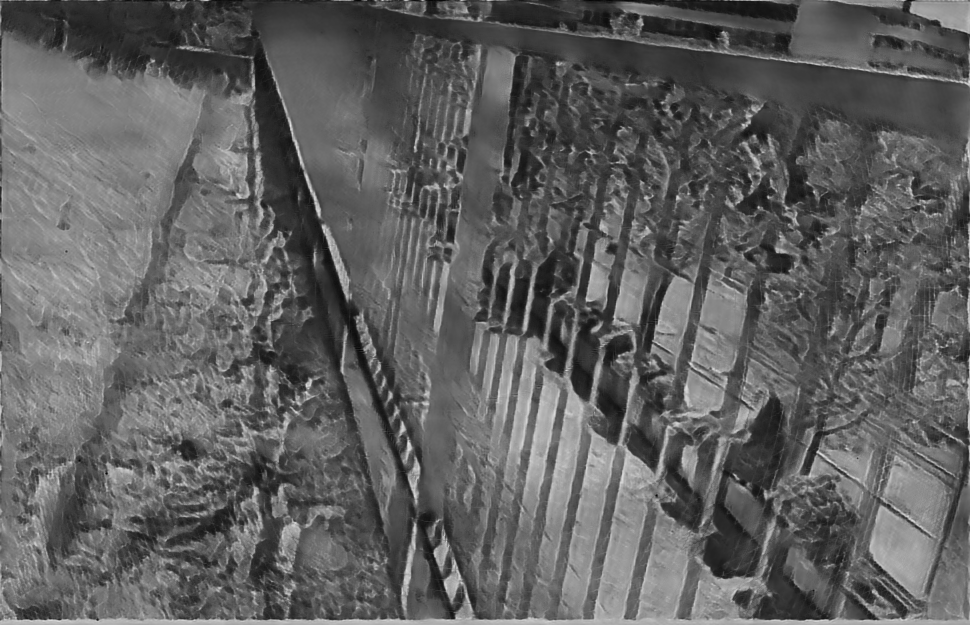}
        &\includegraphics[trim=10px 0px 10px 0px, clip, width=\linewidth]{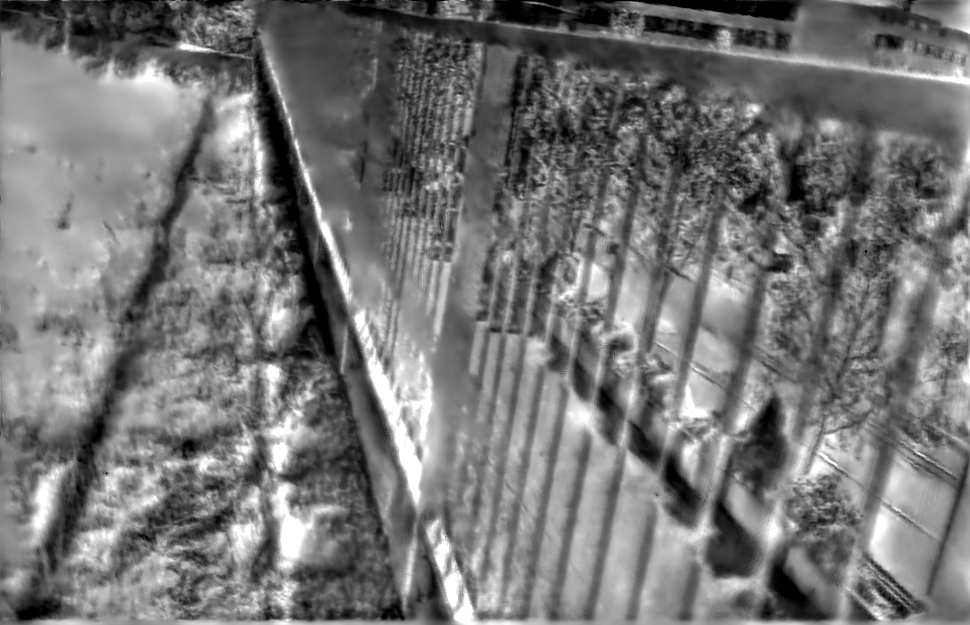}
	&\includegraphics[width=\linewidth]{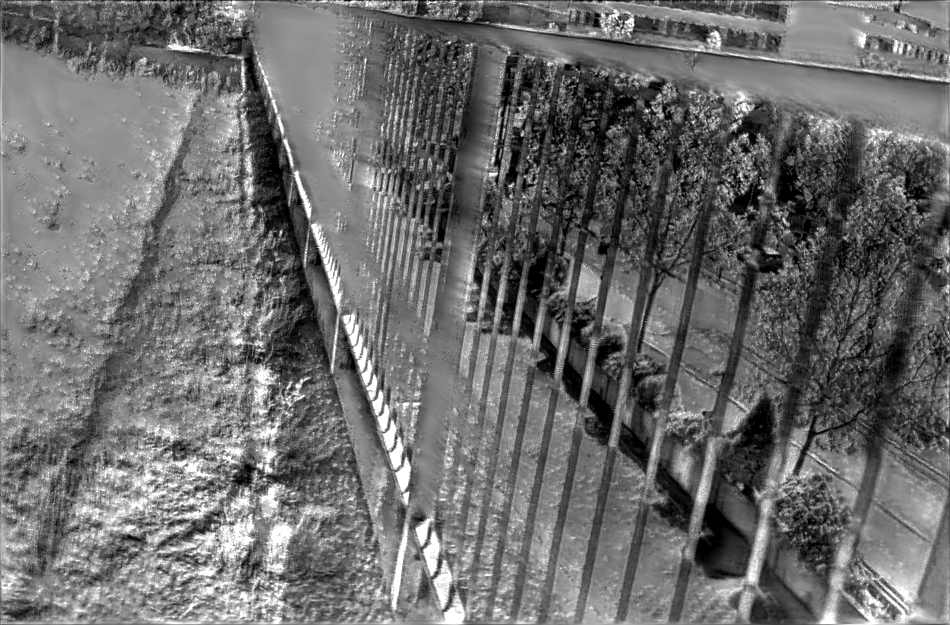}
        &\includegraphics[trim=10px 0px 10px 0px, clip, width=\linewidth]{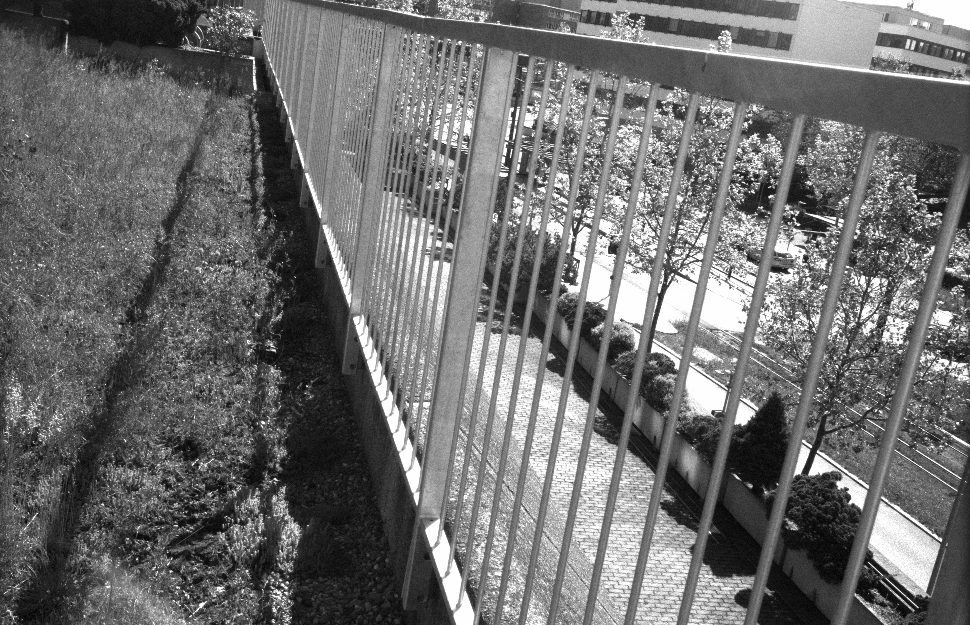}
	\\[-0.2ex]

        \rotatebox{90}{\makecell{rooftop\_02}}
	&\includegraphics[trim=10px 0px 10px 0px, clip, width=\linewidth]{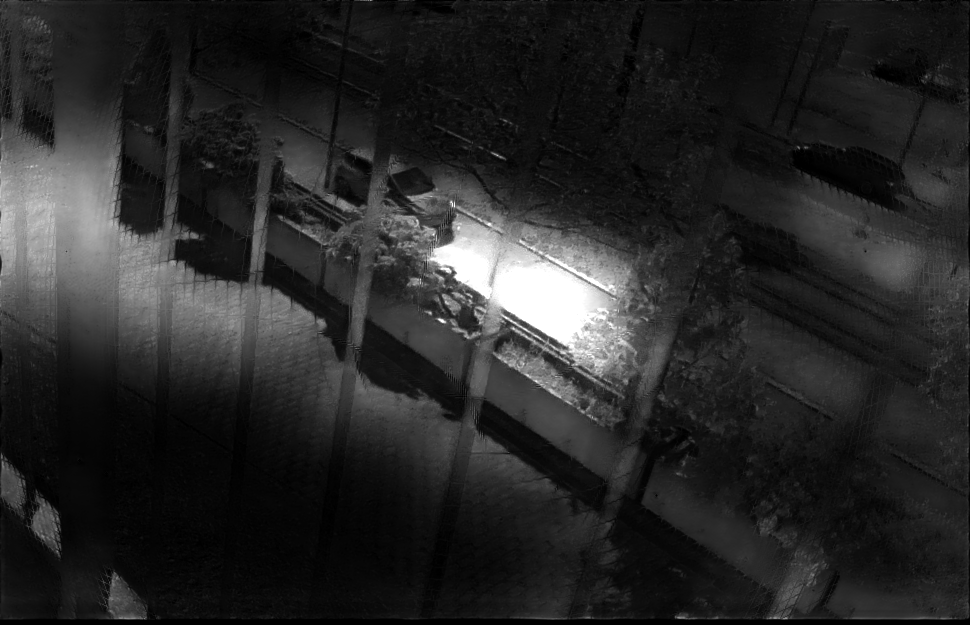}
	&\includegraphics[trim=10px 0px 10px 0px, clip, width=\linewidth]{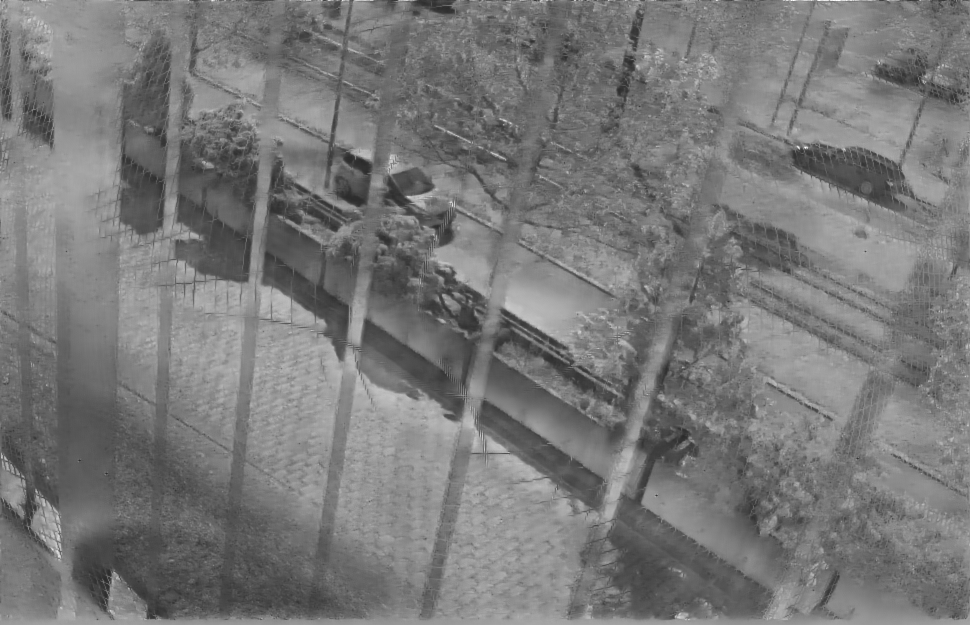}
        &\includegraphics[trim=10px 0px 10px 0px, clip, width=\linewidth]{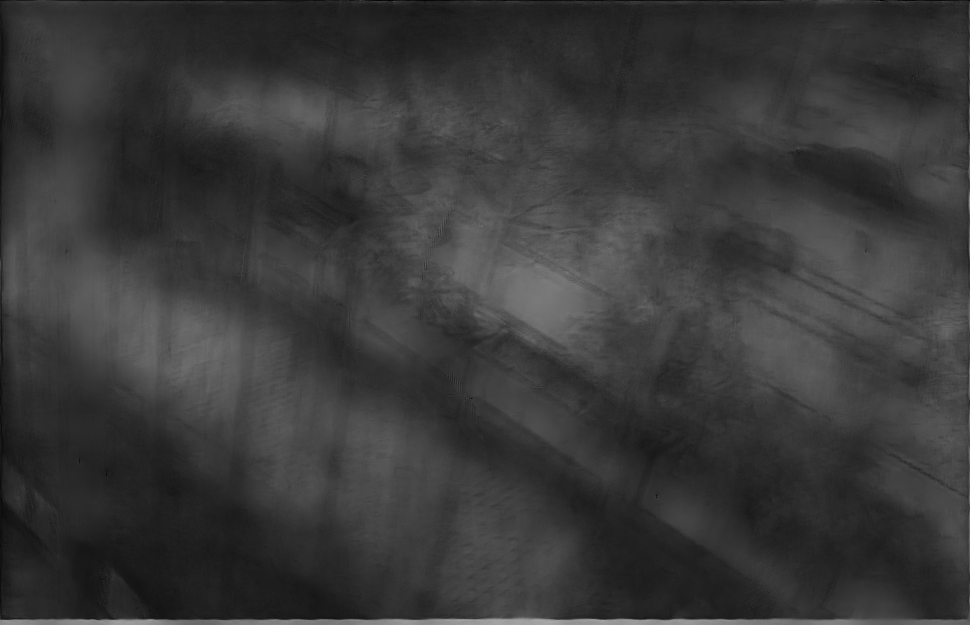}
        &\includegraphics[trim=10px 0px 10px 0px, clip, width=\linewidth]{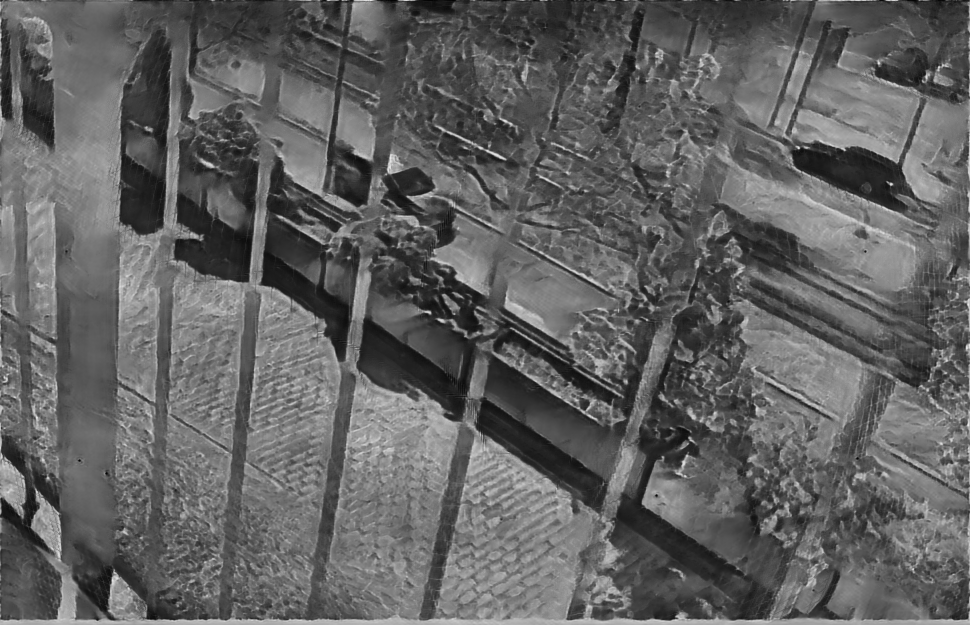}
        &\includegraphics[trim=10px 0px 10px 0px, clip, width=\linewidth]{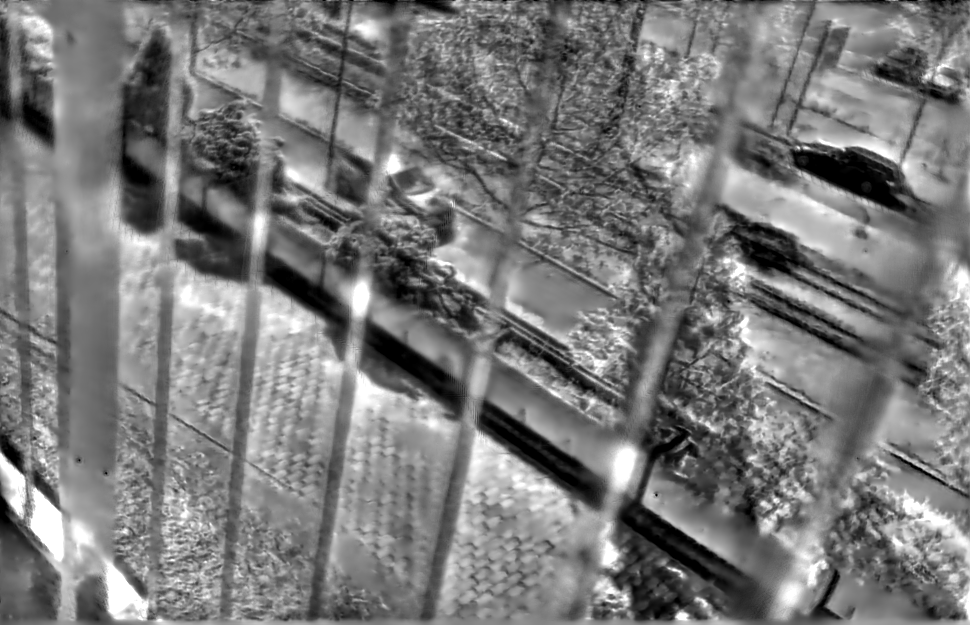}
	&\includegraphics[width=\linewidth]{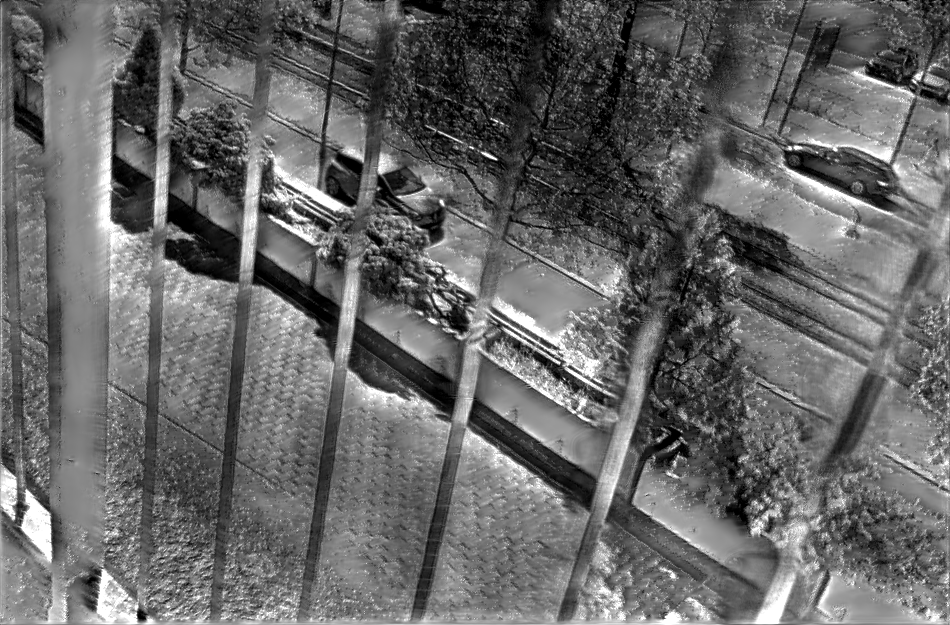}
        &\includegraphics[trim=10px 0px 10px 0px, clip, width=\linewidth]{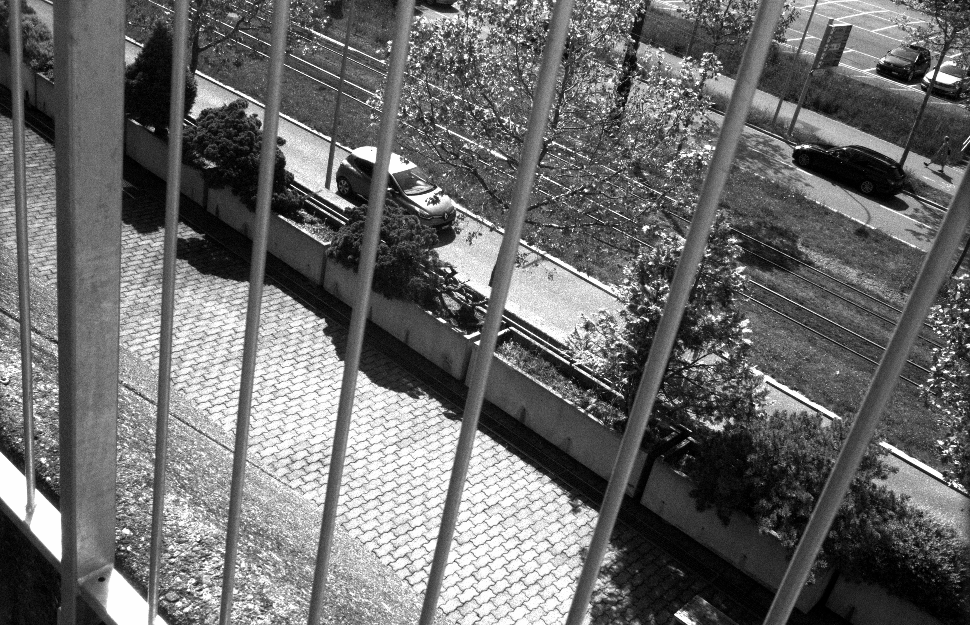}
	\\[-0.2ex]

        \rotatebox{90}{\makecell{rooftop\_03}}
	&\includegraphics[trim=10px 0px 10px 0px, clip, width=\linewidth]{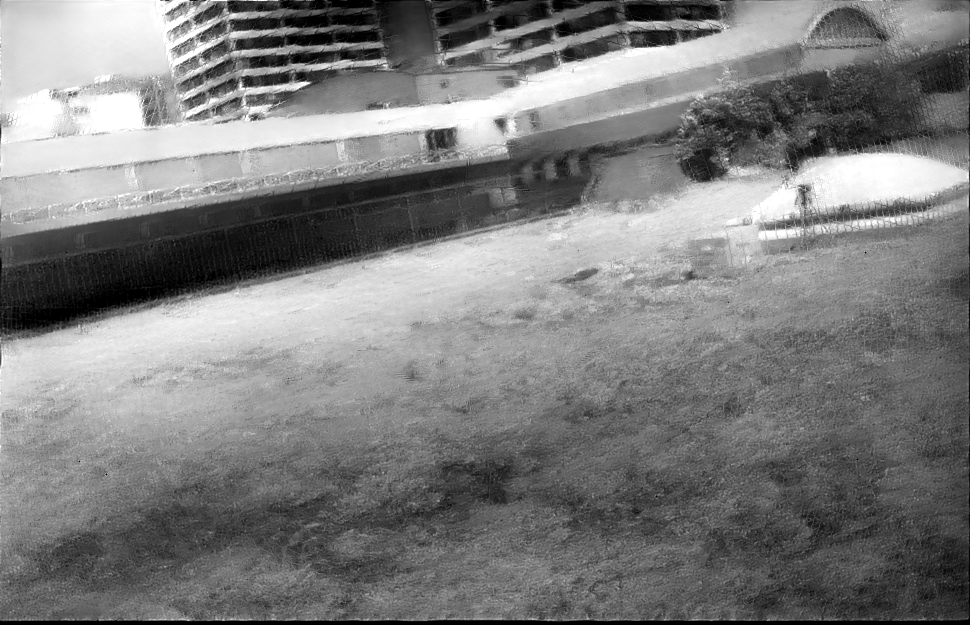}
	&\includegraphics[trim=10px 0px 10px 0px, clip, width=\linewidth]{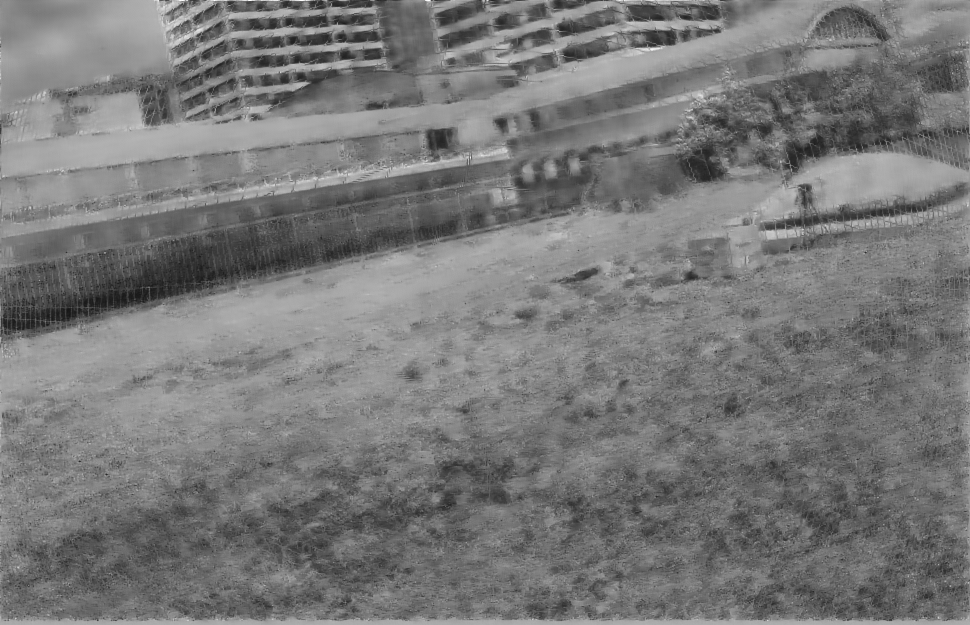}
        &\includegraphics[trim=10px 0px 10px 0px, clip, width=\linewidth]{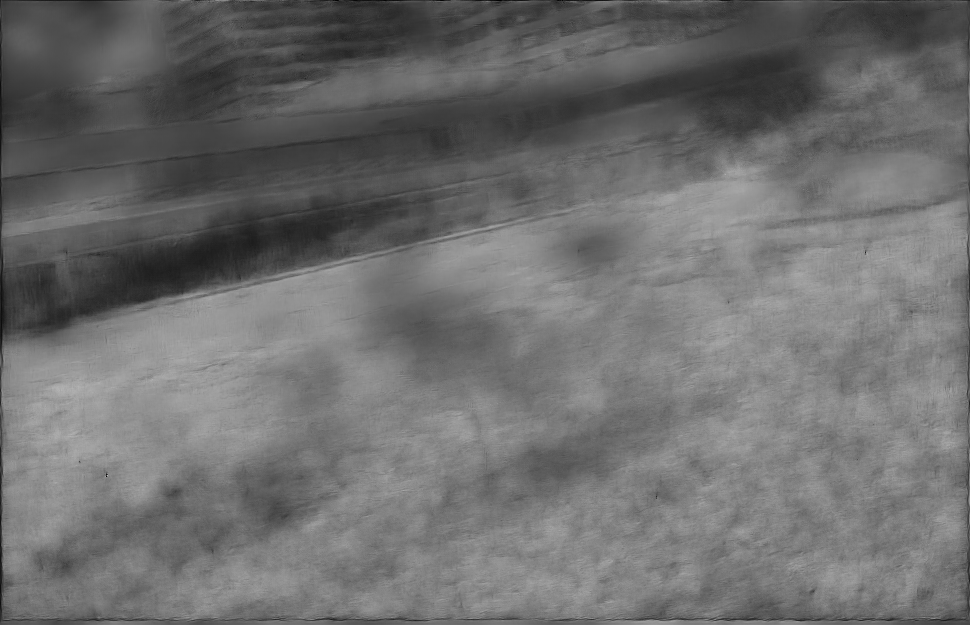}
        &\includegraphics[trim=10px 0px 10px 0px, clip, width=\linewidth]{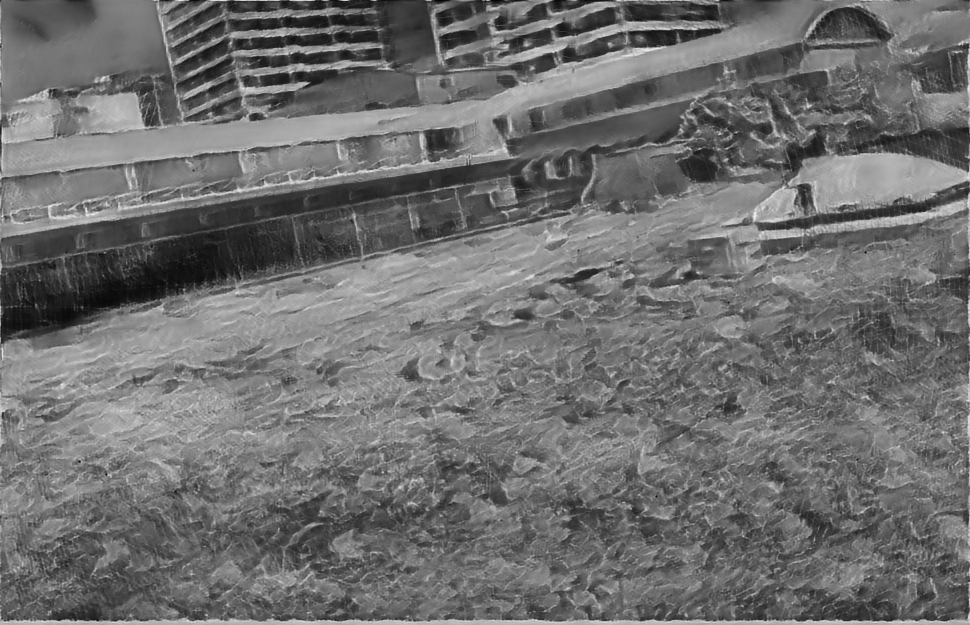}
        &\includegraphics[trim=10px 0px 10px 0px, clip, width=\linewidth]{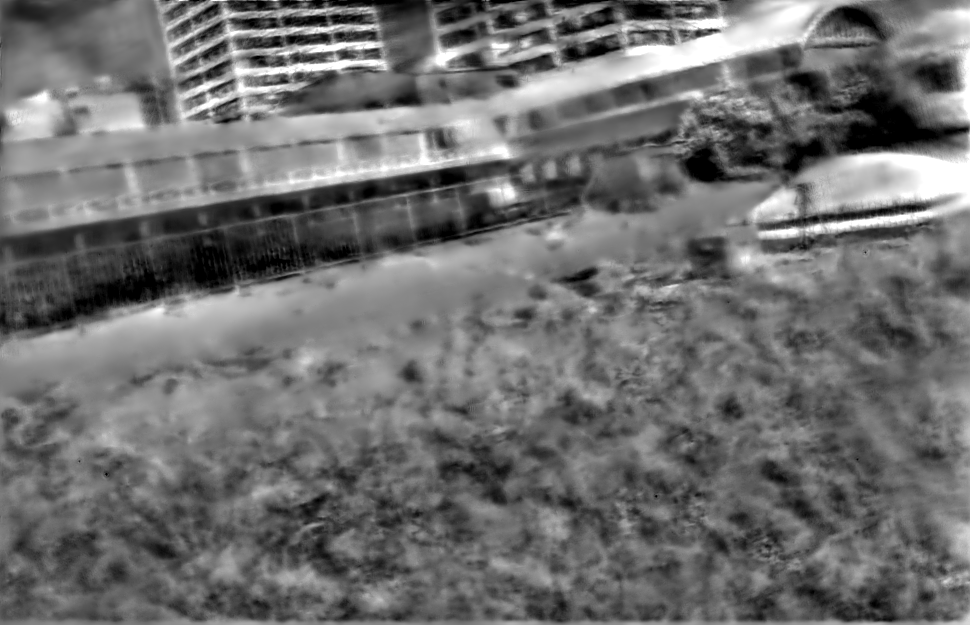}
	&\includegraphics[width=\linewidth]{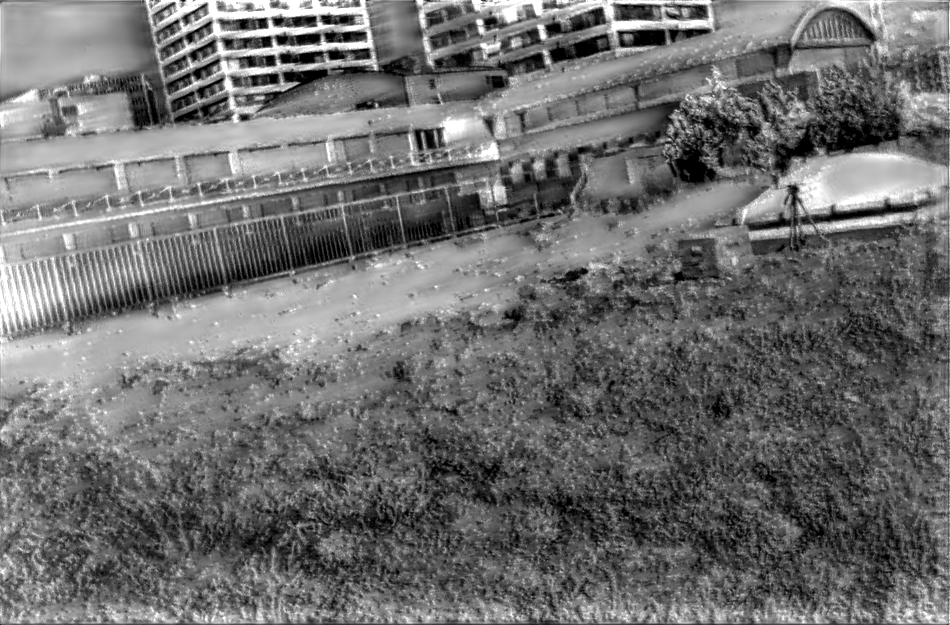}
        &\includegraphics[trim=10px 0px 10px 0px, clip, width=\linewidth]{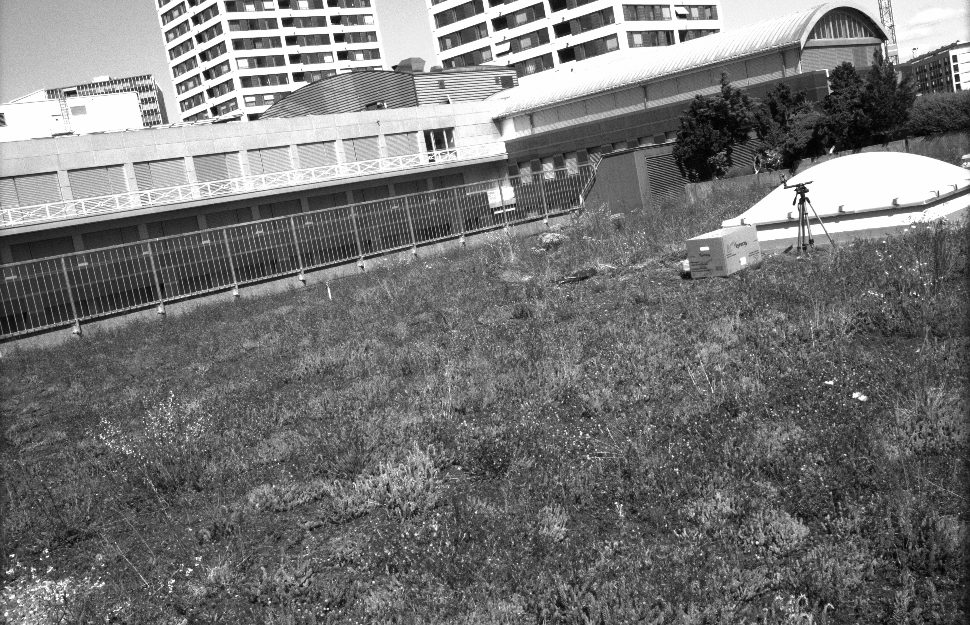}
	\\[-0.2ex]

        \rotatebox{90}{\makecell{rooftop\_03}}
	&\includegraphics[trim=10px 0px 10px 0px, clip, width=\linewidth]{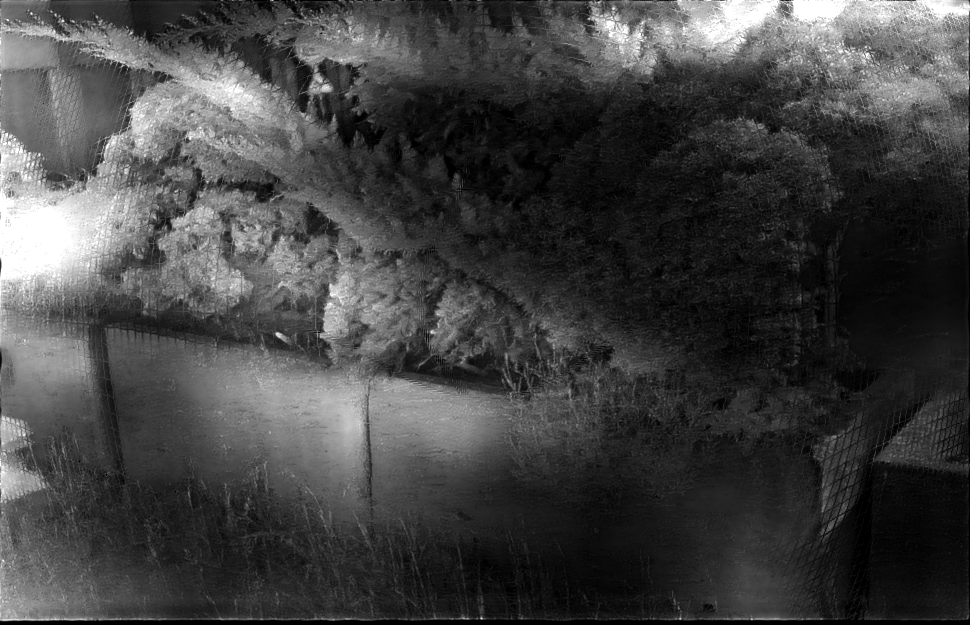}
	&\includegraphics[trim=10px 0px 10px 0px, clip, width=\linewidth]{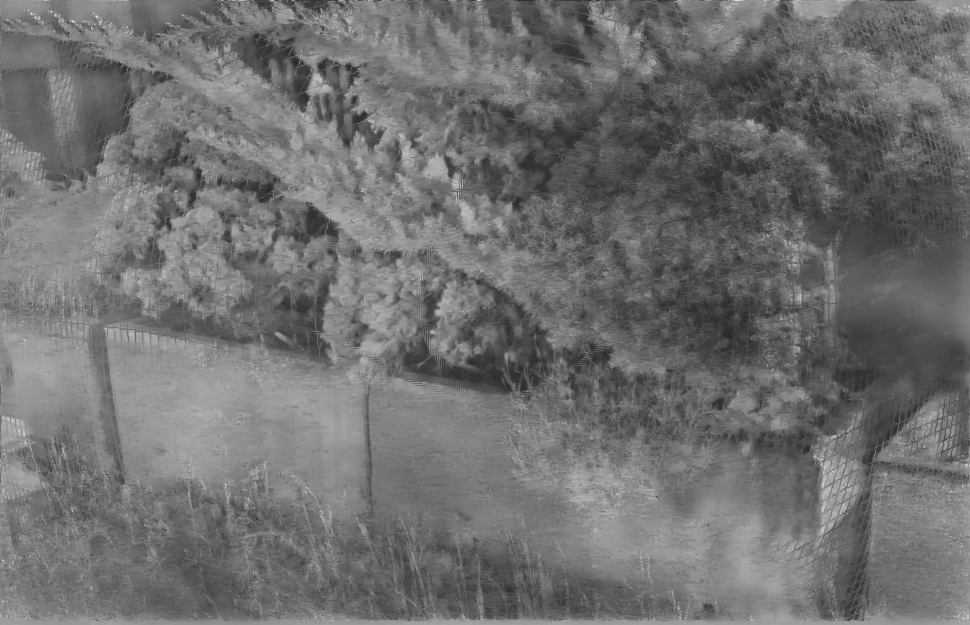}
        &\includegraphics[trim=10px 0px 10px 0px, clip, width=\linewidth]{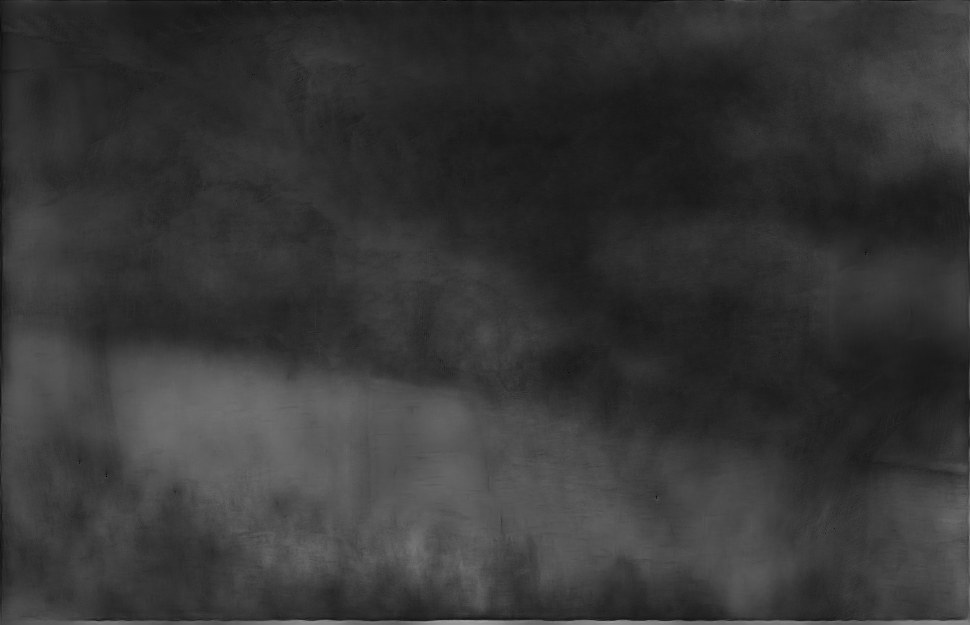}
        &\includegraphics[trim=10px 0px 10px 0px, clip, width=\linewidth]{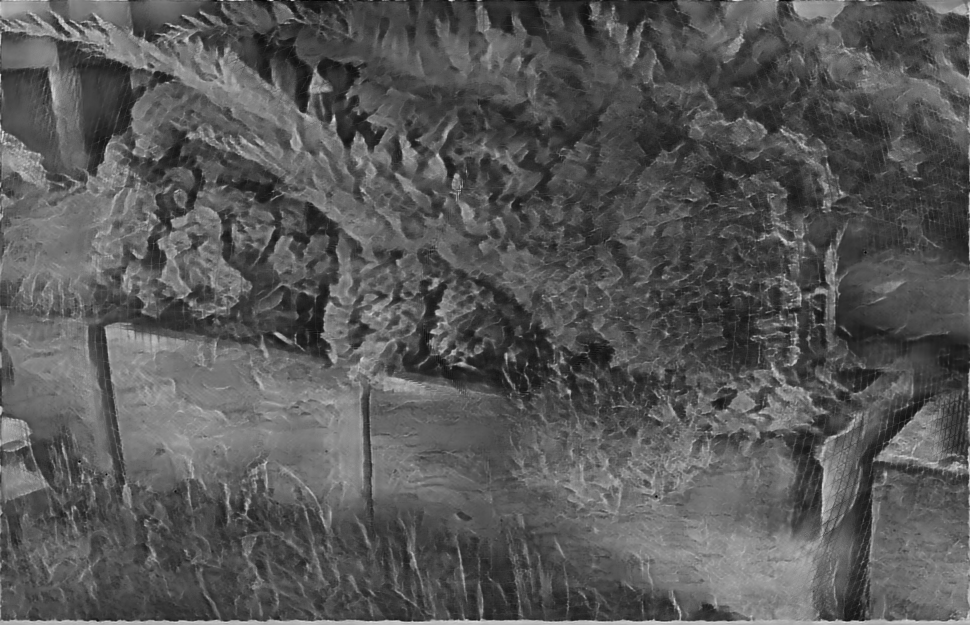}
        &\includegraphics[trim=10px 0px 10px 0px, clip, width=\linewidth]{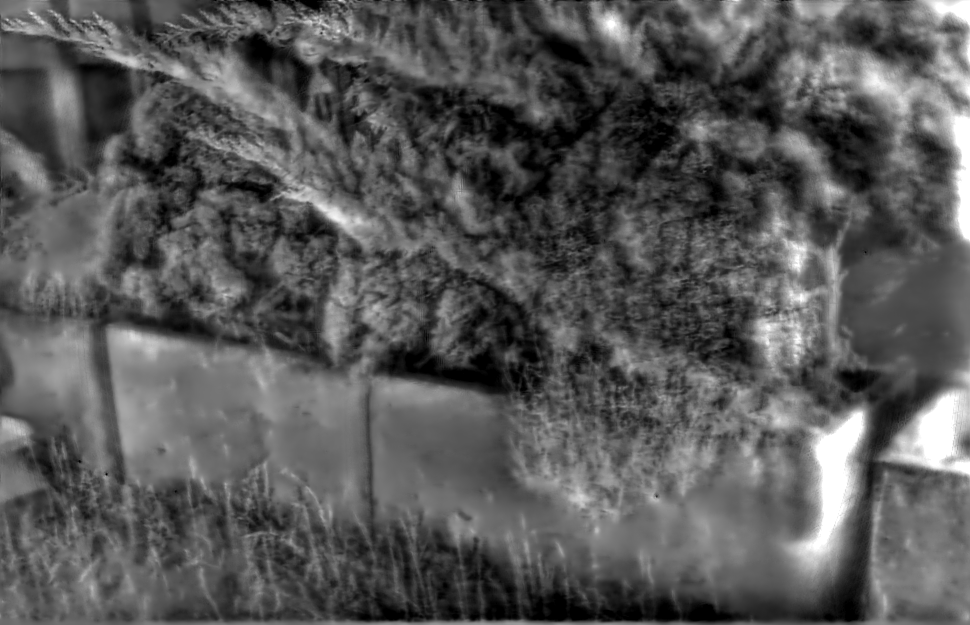}
	&\includegraphics[width=\linewidth]{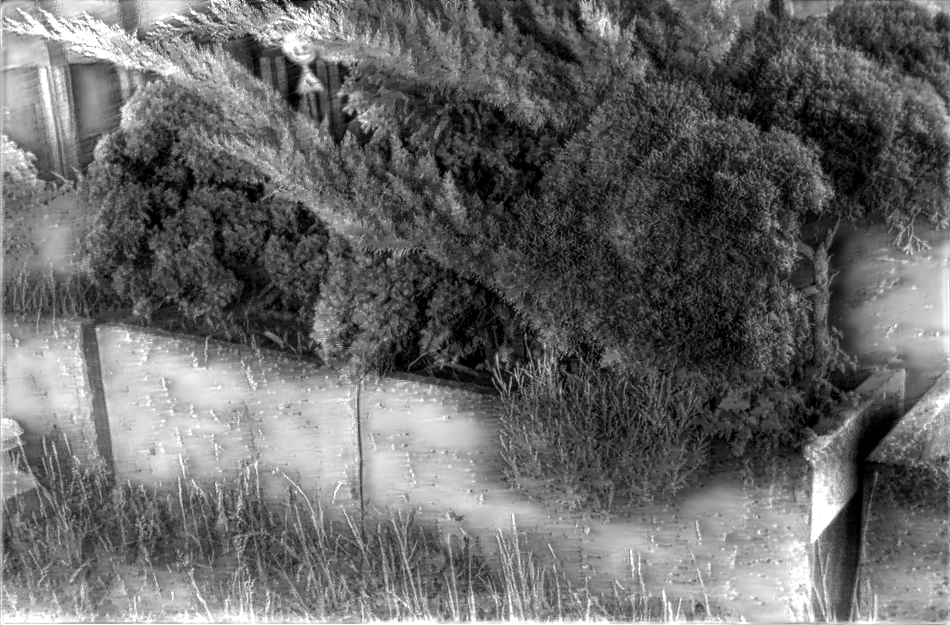}
        &\includegraphics[trim=10px 0px 10px 0px, clip, width=\linewidth]{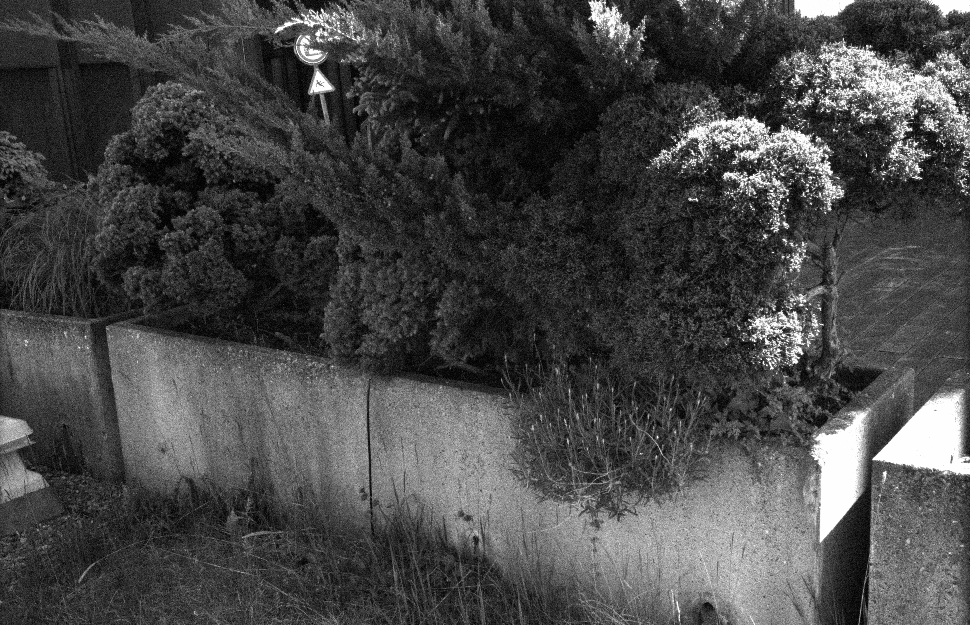}
	\\[-0.2ex]

        \rotatebox{90}{\makecell{rooftop\_02}}
	&\includegraphics[trim=10px 0px 10px 0px, clip, width=\linewidth]{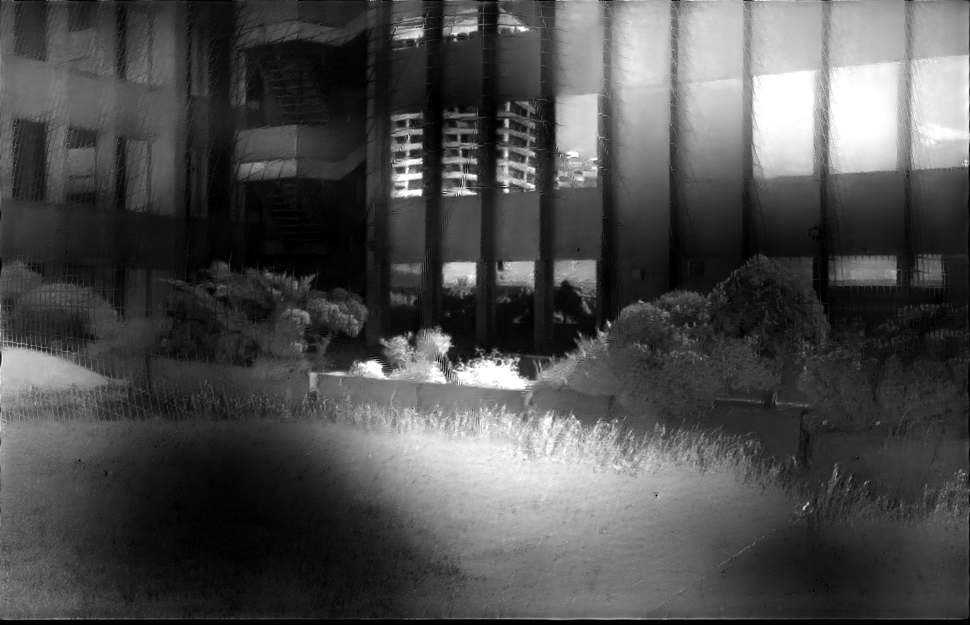}
	&\includegraphics[trim=10px 0px 10px 0px, clip, width=\linewidth]{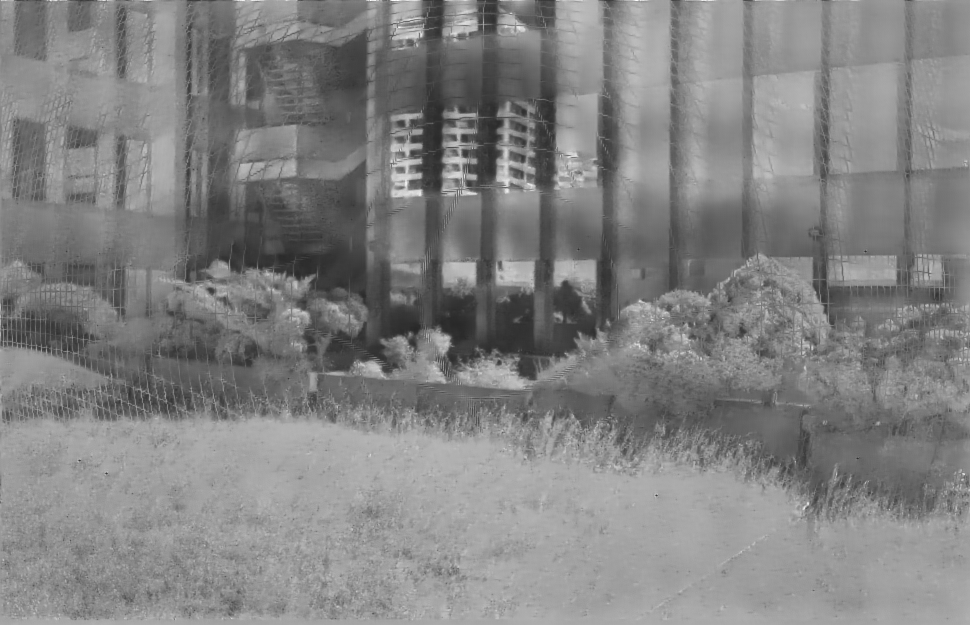}
        &\includegraphics[trim=10px 0px 10px 0px, clip, width=\linewidth]{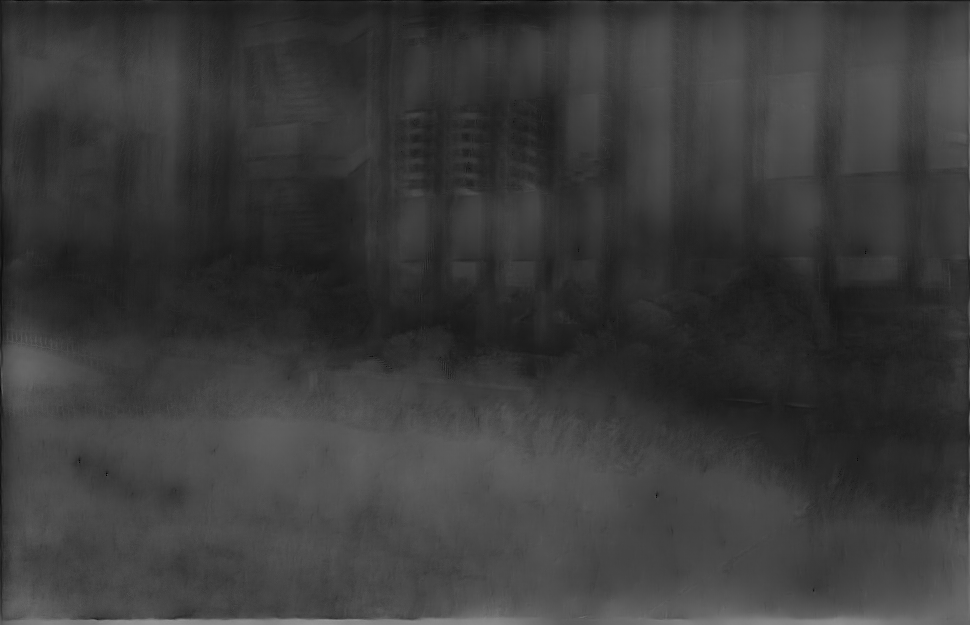}
        &\includegraphics[trim=10px 0px 10px 0px, clip, width=\linewidth]{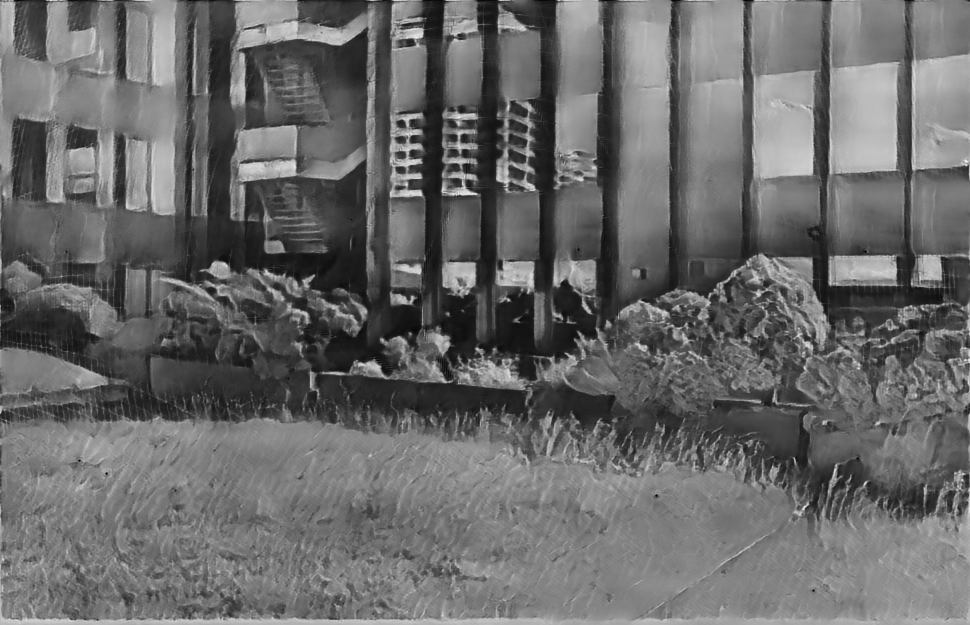}
        &\includegraphics[trim=10px 0px 10px 0px, clip, width=\linewidth]{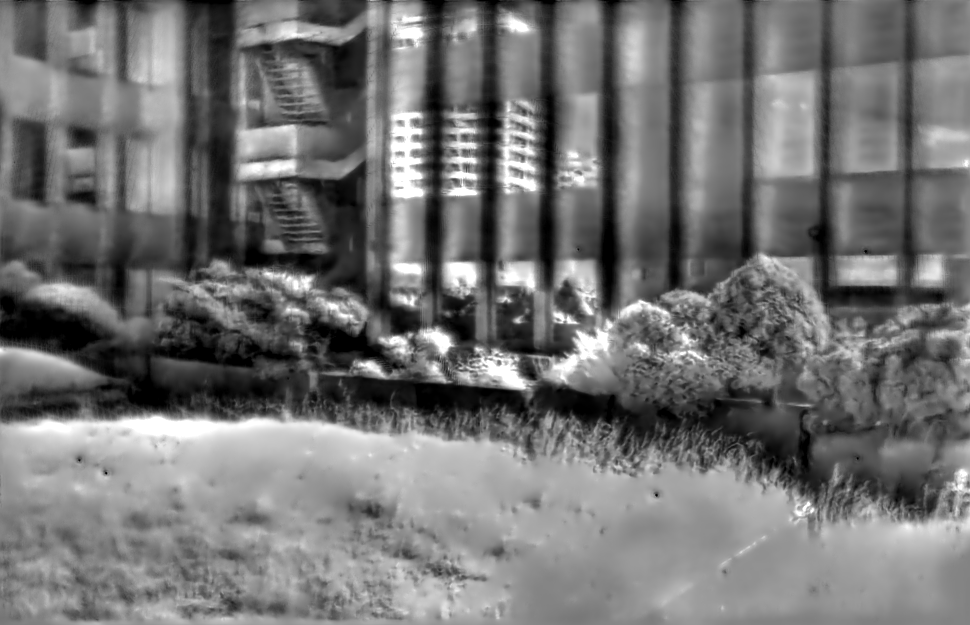}
	&\includegraphics[width=\linewidth]{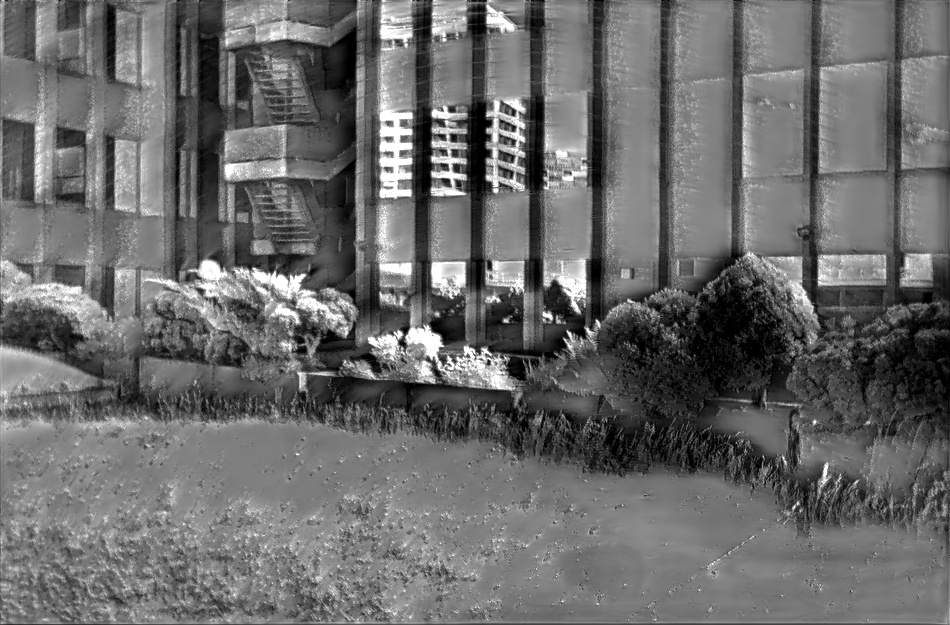}
        &\includegraphics[trim=10px 0px 10px 0px, clip, width=\linewidth]{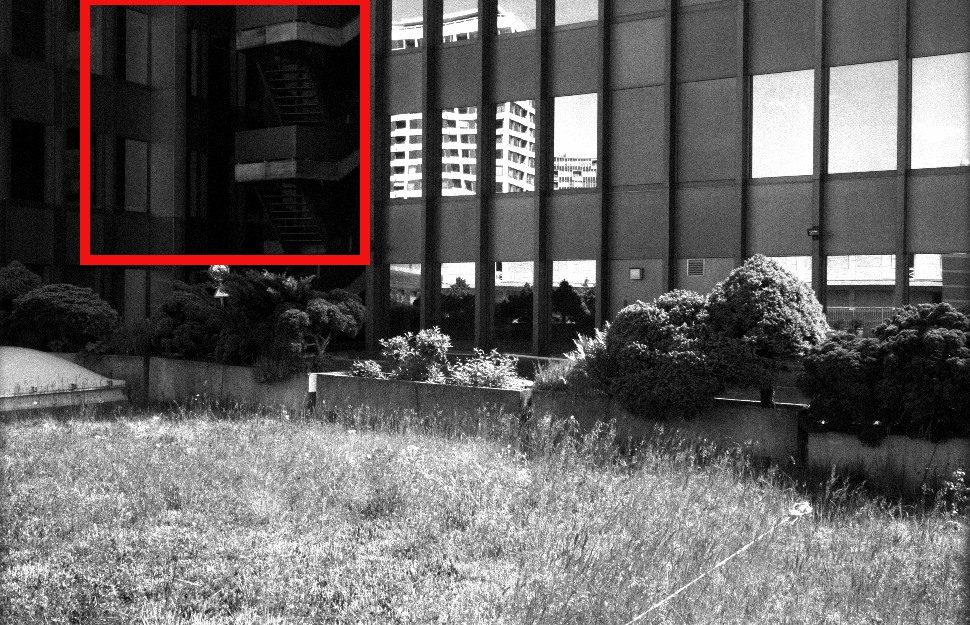}
	\\[-0.2ex]

        \rotatebox{90}{\makecell{rooftop\_05}}
	&\includegraphics[trim=10px 0px 10px 0px, clip, width=\linewidth]{images/suppl_comp/may29_rooftop_handheld_05/132/E2VID.png}
	&\includegraphics[trim=10px 0px 10px 0px, clip, width=\linewidth]{images/suppl_comp/may29_rooftop_handheld_05/132/FireNet.png}
        &\includegraphics[trim=10px 0px 10px 0px, clip, width=\linewidth]{images/suppl_comp/may29_rooftop_handheld_05/132/SPADE-E2VID.png}
        &\includegraphics[trim=10px 0px 10px 0px, clip, width=\linewidth]{images/suppl_comp/may29_rooftop_handheld_05/132/ET-Net.png}
        &\includegraphics[trim=10px 0px 10px 0px, clip, width=\linewidth]{images/suppl_comp/may29_rooftop_handheld_05/132/SSL-E2VID.png}
	&\includegraphics[width=\linewidth]{images/suppl_comp/may29_rooftop_handheld_05/132/intensity.png}
        &\includegraphics[trim=10px 0px 10px 0px, clip, width=\linewidth]{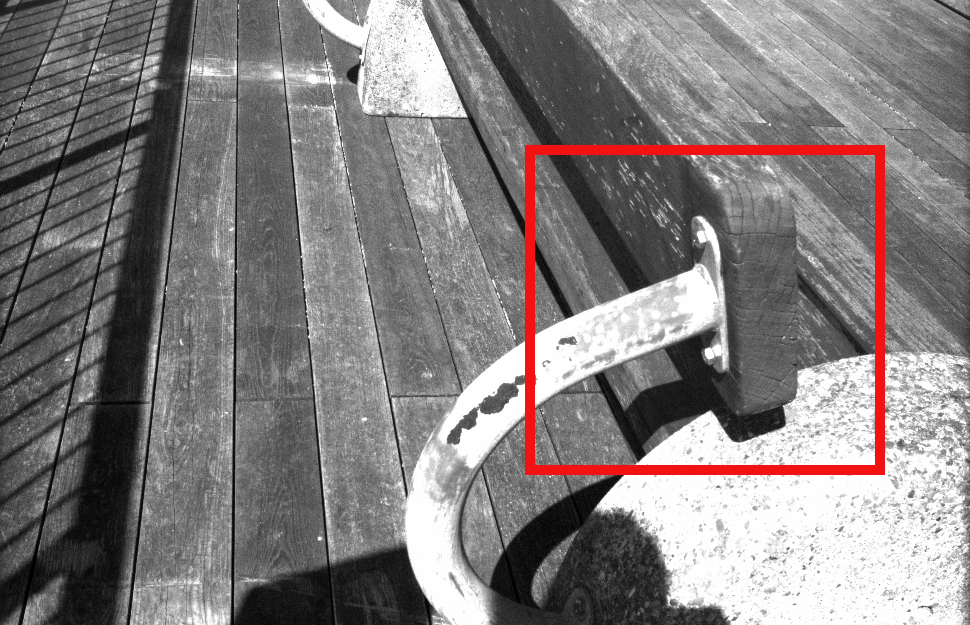}
	\\[-0.2ex]

        & E2VID
        & FireNet
        & SPADE-E2VID
        & ET-Net
        & BTEB
        & Ours
        & References
        \\%[-0.8ex]
	\end{tabular}
	}
    \caption{\emph{Additional qualitative comparison of image intensity reconstruction on the BS-ERGB dataset}. 
    For the last two rows, the regions marked by red boxes are enlarged in \cref{fig:suppl_zoom_in}.
    \label{fig:suppl_comp}}
\end{figure*}

\def\figWidth{0.135\linewidth}
\begin{figure*}[ht]
	\centering
    {\footnotesize
    \setlength{\tabcolsep}{1pt}
	\begin{tabular}{
        >{\centering\arraybackslash}m{0.3cm} 
	>{\centering\arraybackslash}m{\figWidth} 
	>{\centering\arraybackslash}m{\figWidth}
	>{\centering\arraybackslash}m{\figWidth}
	>{\centering\arraybackslash}m{\figWidth}
        >{\centering\arraybackslash}m{\figWidth}
        >{\centering\arraybackslash}m{\figWidth}
        >{\centering\arraybackslash}m{\figWidth}
        }

        \rotatebox{90}{\makecell{rooftop\_02}}
	&\includegraphics[width=\linewidth]{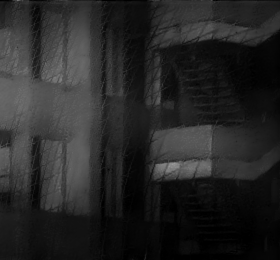}
	&\includegraphics[width=\linewidth]{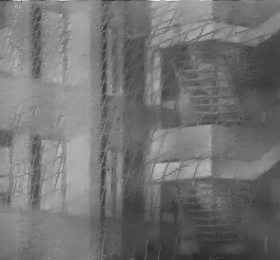}
        &\includegraphics[width=\linewidth]{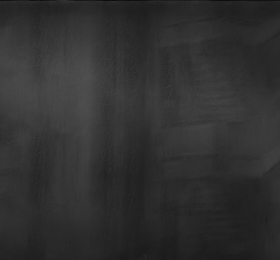}
        &\includegraphics[width=\linewidth]{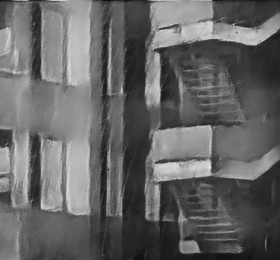}
        &\includegraphics[width=\linewidth]{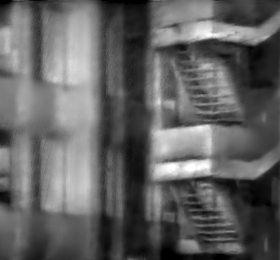}
	&\includegraphics[width=\linewidth]{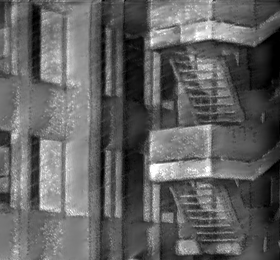}
        &\includegraphics[width=\linewidth]{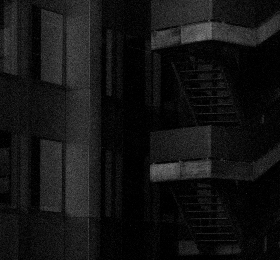}
	\\[-0.2ex]

        \rotatebox{90}{\makecell{rooftop\_05}}
	&\includegraphics[width=\linewidth]{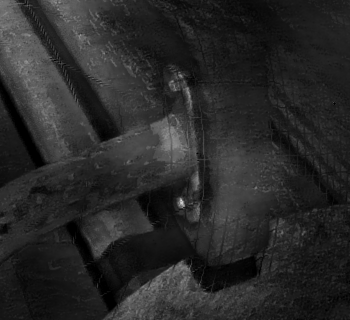}
	&\includegraphics[width=\linewidth]{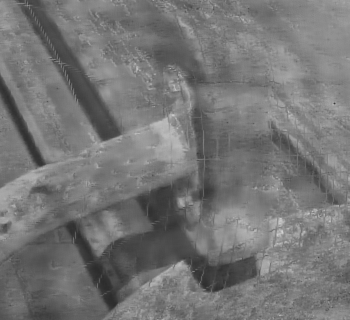}
        &\includegraphics[width=\linewidth]{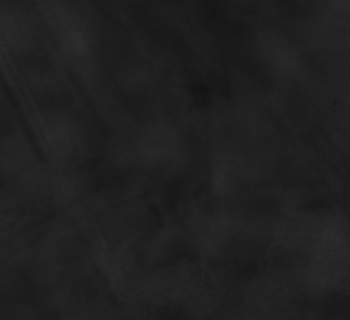}
        &\includegraphics[width=\linewidth]{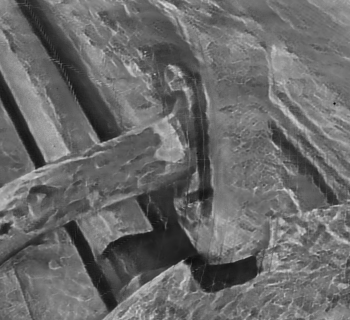}
        &\includegraphics[width=\linewidth]{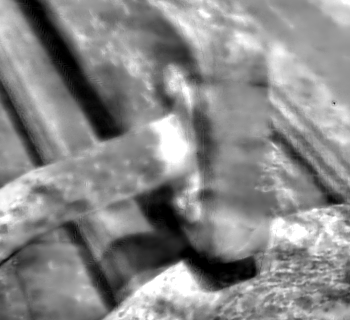}
	&\includegraphics[width=\linewidth]{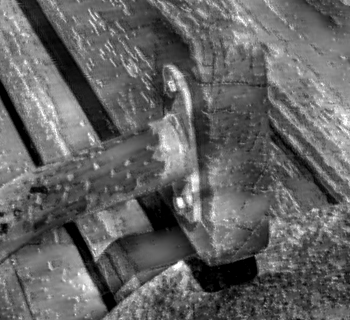}
        &\includegraphics[width=\linewidth]{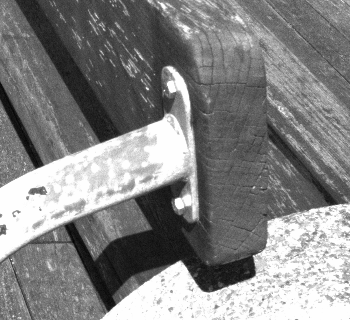}
	\\[-0.2ex]

        & E2VID
        & FireNet
        & SPADE-E2VID
        & ET-Net
        & BTEB
        & Ours
        & References
        \\%[-0.8ex]
	\end{tabular}
	}
    \caption{\emph{Additional qualitative comparison of image intensity reconstruction}. 
    Enlarged regions indicated by red boxes in \cref{fig:suppl_comp}.
    \label{fig:suppl_zoom_in}}
\end{figure*}

\Cref{fig:suppl_results} demonstrates that our model recovers precise optical flow and image intensity on unseen data (i.e., not used for training).
In column c (optical flow), independent moving objects (IMOs) are clearly identified with respect to the background (e.g., the cars in the first row and the motorbike in the second row).
Besides, flow discontinuities agree with the contours of different objects at different depths (e.g., the fence in the third and fourth rows and the bench in the last row).
For image intensity, our model reconstructs fine details at the valid pixels (e.g., the contours of objects), while it leverages the total variation regularization to partially fill in the regions that lack texture and rarely trigger events.

\Cref{fig:suppl_comp} confirms that our model produces competitive results compared to baseline methods.
Our reconstructed images are overall sharper, and are more precise in HDR conditions.
To highlight this, we select two HDR regions (i.e., the stairs of the building and the bench on the rooftop) and present their zoomed-in versions in \cref{fig:suppl_zoom_in}.
It can be clearly seen that:
E2VID and SPADE-E2VID report poor HDR performance;
FireNet produces intensity images with low contrast, where strong artifacts (like spider webs) caused by the rectification of event data using the camera's intrinsic parameters are clearly observed;
BTEB's intensity images are blurred;
ET-Net oversmooths the fine textures on the building wall, and shows strange wrong textures on the bench (especially the stone part in the bottom right).
In contrast, our reconstructed intensity reveals sharp edges and fine details in HDR illumination, where frame-based cameras suffer from under/over exposure problems.

{
    \small
    \bibliographystyle{ieeenat_fullname}
    %\bibliography{all,main}

\begin{thebibliography}{52}
\providecommand{\natexlab}[1]{#1}
\providecommand{\url}[1]{\texttt{#1}}
\expandafter\ifx\csname urlstyle\endcsname\relax
  \providecommand{\doi}[1]{doi: #1}\else
  \providecommand{\doi}{doi: \begingroup \urlstyle{rm}\Url}\fi

\bibitem[Bardow et~al.(2016)Bardow, Davison, and Leutenegger]{Bardow16cvpr}
Patrick Bardow, Andrew~J. Davison, and Stefan Leutenegger.
\newblock Simultaneous optical flow and intensity estimation from an event
  camera.
\newblock In \emph{{IEEE} Conf. Comput. Vis. Pattern Recog. (CVPR)}, pages
  884--892, 2016.

\bibitem[Boyd et~al.(2011)Boyd, Parikh, Chu, Peleato, and
  Eckstein]{Boyd2011admm}
Stephen Boyd, Neal Parikh, Eric Chu, Borja Peleato, and Jonathan Eckstein.
\newblock Distributed optimization and statistical learning via the alternating
  direction method of multipliers.
\newblock \emph{Foundations and Trends in Machine Learning}, 3:\penalty0
  1--122, 2011.

\bibitem[Brebion et~al.(2021)Brebion, Moreau, and Davoine]{Brebion21tits}
Vincent Brebion, Julien Moreau, and Franck Davoine.
\newblock Real-time optical flow for vehicular perception with low- and
  high-resolution event cameras.
\newblock \emph{{IEEE} Trans. Intell. Transport. Syst.}, pages 1--13, 2021.

\bibitem[Chen et~al.(2020)Chen, Cao, Conradt, Tang, Rohrbein, and
  Knoll]{Chen20msp}
Guang Chen, Hu Cao, Jorg Conradt, Huajin Tang, Florian Rohrbein, and Alois
  Knoll.
\newblock Event-based neuromorphic vision for autonomous driving: A paradigm
  shift for bio-inspired visual sensing and perception.
\newblock \emph{{IEEE} Signal Process. Mag.}, 37\penalty0 (4):\penalty0 34--49,
  2020.

\bibitem[Cook et~al.(2011)Cook, Gugelmann, Jug, Krautz, and
  Steger]{Cook11ijcnn}
Matthew Cook, Luca Gugelmann, Florian Jug, Christoph Krautz, and Angelika
  Steger.
\newblock Interacting maps for fast visual interpretation.
\newblock In \emph{Int. Joint Conf. Neural Netw. (IJCNN)}, pages 770--776,
  2011.

\bibitem[Ercan et~al.(2023)Ercan, Eker, Erdem, and Erdem]{Ercan23cvprw}
Burak Ercan, Onur Eker, Aykut Erdem, and Erkut Erdem.
\newblock {EVREAL}: Towards a comprehensive benchmark and analysis suite for
  event-based video reconstruction.
\newblock In \emph{{IEEE} Conf. Comput. Vis. Pattern Recog. Workshops (CVPRW)},
  2023.

\bibitem[Gallego et~al.(2018)Gallego, Rebecq, and Scaramuzza]{Gallego18cvpr}
Guillermo Gallego, Henri Rebecq, and Davide Scaramuzza.
\newblock A unifying contrast maximization framework for event cameras, with
  applications to motion, depth, and optical flow estimation.
\newblock In \emph{{IEEE} Conf. Comput. Vis. Pattern Recog. (CVPR)}, pages
  3867--3876, 2018.

\bibitem[Gallego et~al.(2019)Gallego, Gehrig, and Scaramuzza]{Gallego19cvpr}
Guillermo Gallego, Mathias Gehrig, and Davide Scaramuzza.
\newblock Focus is all you need: Loss functions for event-based vision.
\newblock In \emph{{IEEE} Conf. Comput. Vis. Pattern Recog. (CVPR)}, pages
  12272--12281, 2019.

\bibitem[Gallego et~al.(2022)Gallego, Delbruck, Orchard, Bartolozzi, Taba,
  Censi, Leutenegger, Davison, Conradt, Daniilidis, and
  Scaramuzza]{Gallego20pami}
Guillermo Gallego, Tobi Delbruck, Garrick Orchard, Chiara Bartolozzi, Brian
  Taba, Andrea Censi, Stefan Leutenegger, Andrew Davison, J{\"o}rg Conradt,
  Kostas Daniilidis, and Davide Scaramuzza.
\newblock Event-based vision: A survey.
\newblock \emph{{IEEE} Trans. Pattern Anal. Mach. Intell.}, 44\penalty0
  (1):\penalty0 154--180, 2022.

\bibitem[Gantier~Cadena et~al.(2021)Gantier~Cadena, Qian, Wang, and
  Yang]{Gantier21tip}
Pablo~Rodrigo Gantier~Cadena, Yeqiang Qian, Chunxiang Wang, and Ming Yang.
\newblock {SPADE}-{E2VID}: Spatially-adaptive denormalization for event-based
  video reconstruction.
\newblock \emph{{IEEE} Trans. Image Process.}, 30:\penalty0 2488--2500, 2021.

\bibitem[Gehrig et~al.(2019)Gehrig, Loquercio, Derpanis, and
  Scaramuzza]{Gehrig19iccv}
Daniel Gehrig, Antonio Loquercio, Konstantinos~G. Derpanis, and Davide
  Scaramuzza.
\newblock End-to-end learning of representations for asynchronous event-based
  data.
\newblock In \emph{Int. Conf. Comput. Vis. (ICCV)}, pages 5632--5642, 2019.

\bibitem[Gehrig et~al.(2021{\natexlab{a}})Gehrig, Aarents, Gehrig, and
  Scaramuzza]{Gehrig21ral}
Mathias Gehrig, Willem Aarents, Daniel Gehrig, and Davide Scaramuzza.
\newblock {DSEC}: A stereo event camera dataset for driving scenarios.
\newblock \emph{{IEEE} Robot. Autom. Lett.}, 6\penalty0 (3):\penalty0
  4947--4954, 2021{\natexlab{a}}.

\bibitem[Gehrig et~al.(2021{\natexlab{b}})Gehrig, Millhäusler, Gehrig, and
  Scaramuzza]{Gehrig21threedv}
Mathias Gehrig, Mario Millhäusler, Daniel Gehrig, and Davide Scaramuzza.
\newblock {E-RAFT}: Dense optical flow from event cameras.
\newblock In \emph{Int. Conf. 3D Vision (3DV)}, pages 197--206,
  2021{\natexlab{b}}.

\bibitem[Guo and Gallego(2024)]{Guo24eccv}
Shuang Guo and Guillermo Gallego.
\newblock Event-based mosaicing bundle adjustment.
\newblock In \emph{Eur. Conf. Comput. Vis. (ECCV)}, pages 479--496, 2024.

\bibitem[Guo and Gallego(2025)]{Guo24epba}
Shuang Guo and Guillermo Gallego.
\newblock Event-based photometric bundle adjustment.
\newblock \emph{{IEEE} Trans. Pattern Anal. Mach. Intell.}, 2025.

\bibitem[Hamann et~al.(2024)Hamann, Wang, Asmanis, Chaney, Gallego, and
  Daniilidis]{Hamann24eccv}
Friedhelm Hamann, Ziyun Wang, Ioannis Asmanis, Kenneth Chaney, Guillermo
  Gallego, and Kostas Daniilidis.
\newblock Motion-prior contrast maximization for dense continuous-time motion
  estimation.
\newblock In \emph{Eur. Conf. Comput. Vis. (ECCV)}, pages 18--37, 2024.

\bibitem[Li et~al.(2023)Li, Huang, Chen, Shi, Li, Bao, Cui, and
  Zhang]{Li23iros}
Yijin Li, Zhaoyang Huang, Shuo Chen, Xiaoyu Shi, Hongsheng Li, Hujun Bao,
  Zhaopeng Cui, and Guofeng Zhang.
\newblock {B}link{F}low: A dataset to push the limits of event-based optical
  flow estimation.
\newblock In \emph{IEEE/RSJ Int. Conf. Intell. Robot. Syst. (IROS)}, pages
  3881--3888, 2023.

\bibitem[Lichtsteiner et~al.(2008)Lichtsteiner, Posch, and
  Delbruck]{Lichtsteiner08ssc}
Patrick Lichtsteiner, Christoph Posch, and Tobi Delbruck.
\newblock {A 128$\times$128 120 dB 15 $\mu$s latency asynchronous temporal
  contrast vision sensor}.
\newblock \emph{{IEEE} J. Solid-State Circuits}, 43\penalty0 (2):\penalty0
  566--576, 2008.

\bibitem[Liu et~al.(2023)Liu, Chen, Qu, Zhang, Li, Knoll, and Jiang]{Liu23iccv}
Haotian Liu, Guang Chen, Sanqing Qu, Yanping Zhang, Zhijun Li, Alois Knoll, and
  Changjun Jiang.
\newblock {TMA}: Temporal motion aggregation for event-based optical flow.
\newblock In \emph{Int. Conf. Comput. Vis. (ICCV)}, pages 9651--9660, 2023.

\bibitem[Loshchilov and Hutter(2019)]{Loshchilov2019iclr}
Ilya Loshchilov and Frank Hutter.
\newblock Decoupled weight decay regularization.
\newblock In \emph{Int. Conf. Learn. Representations ({ICLR})}, 2019.

\bibitem[Luo et~al.(2023)Luo, Luo, Luo, Wang, Tan, and Liu]{Luo23iccv}
Xinglong Luo, Kunming Luo, Ao Luo, Zhengning Wang, Ping Tan, and Shuaicheng
  Liu.
\newblock Learning optical flow from event camera with rendered dataset.
\newblock In \emph{Int. Conf. Comput. Vis. (ICCV)}, pages 9847--9857, 2023.

\bibitem[Mitrokhin et~al.(2018)Mitrokhin, Fermuller, Parameshwara, and
  Aloimonos]{Mitrokhin18iros}
Anton Mitrokhin, Cornelia Fermuller, Chethan Parameshwara, and Yiannis
  Aloimonos.
\newblock Event-based moving object detection and tracking.
\newblock In \emph{IEEE/RSJ Int. Conf. Intell. Robot. Syst. (IROS)}, pages
  1--9, 2018.

\bibitem[Mittal et~al.(2012)Mittal, Moorthy, and Bovik]{Mittal12tip}
Anish Mittal, Anush~Krishna Moorthy, and Alan~Conrad Bovik.
\newblock No-reference image quality assessment in the spatial domain.
\newblock \emph{{IEEE} Trans. Image Process.}, 21\penalty0 (12):\penalty0
  4695--4708, 2012.

\bibitem[Mittal et~al.(2013)Mittal, Soundararajan, and Bovik]{Mittal13spl}
A. Mittal, R. Soundararajan, and A.~C. Bovik.
\newblock Making a “completely blind” image quality analyzer.
\newblock \emph{{IEEE} Signal Process. Lett.}, 20\penalty0 (3):\penalty0
  209--212, 2013.

\bibitem[Mostafavi~I. et~al.(2021)Mostafavi~I., Wang, and
  Yoon]{Mostafavi21ijcv}
S.M. Mostafavi~I., Lin Wang, and Kuk-Jin~Yoon Yoon.
\newblock Learning to reconstruct {HDR} images from events, with applications
  to depth and flow prediction.
\newblock \emph{Int. J. Comput. Vis.}, 129\penalty0 (4):\penalty0 900--920,
  2021.

\bibitem[Mueggler et~al.(2017)Mueggler, Rebecq, Gallego, Delbruck, and
  Scaramuzza]{Mueggler17ijrr}
Elias Mueggler, Henri Rebecq, Guillermo Gallego, Tobi Delbruck, and Davide
  Scaramuzza.
\newblock The event-camera dataset and simulator: Event-based data for pose
  estimation, visual odometry, and {SLAM}.
\newblock \emph{Int. J. Robot. Research}, 36\penalty0 (2):\penalty0 142--149,
  2017.

\bibitem[Munda et~al.(2018)Munda, Reinbacher, and Pock]{Munda18ijcv}
Gottfried Munda, Christian Reinbacher, and Thomas Pock.
\newblock Real-time intensity-image reconstruction for event cameras using
  manifold regularisation.
\newblock \emph{Int. J. Comput. Vis.}, 126\penalty0 (12):\penalty0 1381--1393,
  2018.

\bibitem[Paredes-Vall{\'e}s and de~Croon(2021)]{Paredes21cvpr}
Federico Paredes-Vall{\'e}s and Guido C. H.~E. de Croon.
\newblock Back to event basics: Self-supervised learning of image
  reconstruction for event cameras via photometric constancy.
\newblock In \emph{{IEEE} Conf. Comput. Vis. Pattern Recog. (CVPR)}, pages
  3445--3454, 2021.

\bibitem[Paredes-Vall{\'e}s et~al.(2023)Paredes-Vall{\'e}s, Scheper, De~Wagter,
  and de~Croon]{Paredes23iccv}
Federico Paredes-Vall{\'e}s, Kirk~YW Scheper, Christophe De~Wagter, and
  Guido~CHE de Croon.
\newblock Taming contrast maximization for learning sequential, low-latency,
  event-based optical flow.
\newblock In \emph{Int. Conf. Comput. Vis. (ICCV)}, pages 9661--9671, 2023.

\bibitem[Posch et~al.(2014)Posch, Serrano-Gotarredona, Linares-Barranco, and
  Delbruck]{Posch14ieee}
Christoph Posch, Teresa Serrano-Gotarredona, Bernabe Linares-Barranco, and Tobi
  Delbruck.
\newblock Retinomorphic event-based vision sensors: Bioinspired cameras with
  spiking output.
\newblock \emph{Proc. {IEEE}}, 102\penalty0 (10):\penalty0 1470--1484, 2014.

\bibitem[Rebecq et~al.(2021)Rebecq, Ranftl, Koltun, and
  Scaramuzza]{Rebecq19pami}
Henri Rebecq, Ren{\'{e}} Ranftl, Vladlen Koltun, and Davide Scaramuzza.
\newblock High speed and high dynamic range video with an event camera.
\newblock \emph{{IEEE} Trans. Pattern Anal. Mach. Intell.}, 43\penalty0
  (6):\penalty0 1964--1980, 2021.

\bibitem[Ronneberger et~al.(2015)Ronneberger, Fischer, and
  Brox]{Ronneberger15icmicci}
Olaf Ronneberger, Philipp Fischer, and Thomas Brox.
\newblock {U}-{N}et: Convolutional networks for biomedical image segmentation.
\newblock In \emph{Int. Conf. Medical Image Computing and Computer-Assisted
  Intervention (MICCAI)}, pages 234--241, 2015.

\bibitem[Rudin et~al.(1992)Rudin, Osher, and Fatemi]{Rudin92physica}
Leonid~I. Rudin, Stanley Osher, and Emad Fatemi.
\newblock Nonlinear total variation based noise removal algorithms.
\newblock \emph{Physica D: Nonlinear Phenomena}, 60\penalty0 (1--4):\penalty0
  259--268, 1992.

\bibitem[Scheerlinck et~al.(2018)Scheerlinck, Barnes, and
  Mahony]{Scheerlinck18accv}
Cedric Scheerlinck, Nick Barnes, and Robert Mahony.
\newblock Continuous-time intensity estimation using event cameras.
\newblock In \emph{Asian Conf. Comput. Vis. (ACCV)}, pages 308--324, 2018.

\bibitem[Scheerlinck et~al.(2020)Scheerlinck, Rebecq, Gehrig, Barnes, Mahony,
  and Scaramuzza]{Scheerlinck20wacv}
Cedric Scheerlinck, Henri Rebecq, Daniel Gehrig, Nick Barnes, Robert Mahony,
  and Davide Scaramuzza.
\newblock Fast image reconstruction with an event camera.
\newblock In \emph{{IEEE} Winter Conf. Appl. Comput. Vis. (WACV)}, pages
  156--163, 2020.

\bibitem[Shiba et~al.(2022{\natexlab{a}})Shiba, Aoki, and Gallego]{Shiba22aisy}
Shintaro Shiba, Yoshimitsu Aoki, and Guillermo Gallego.
\newblock A fast geometric regularizer to mitigate event collapse in the
  contrast maximization framework.
\newblock \emph{Adv. Intell. Syst.}, page 2200251, 2022{\natexlab{a}}.

\bibitem[Shiba et~al.(2022{\natexlab{b}})Shiba, Aoki, and Gallego]{Shiba22eccv}
Shintaro Shiba, Yoshimitsu Aoki, and Guillermo Gallego.
\newblock Secrets of event-based optical flow.
\newblock In \emph{Eur. Conf. Comput. Vis. (ECCV)}, pages 628--645,
  2022{\natexlab{b}}.

\bibitem[Shiba et~al.(2022{\natexlab{c}})Shiba, Aoki, and
  Gallego]{Shiba22sensors}
Shintaro Shiba, Yoshimitsu Aoki, and Guillermo Gallego.
\newblock Event collapse in contrast maximization frameworks.
\newblock \emph{Sensors}, 22\penalty0 (14):\penalty0 1--20, 2022{\natexlab{c}}.

\bibitem[Shiba et~al.(2024)Shiba, Klose, Aoki, and Gallego]{Shiba24pami}
Shintaro Shiba, Yannick Klose, Yoshimitsu Aoki, and Guillermo Gallego.
\newblock Secrets of event-based optical flow, depth, and ego-motion by
  contrast maximization.
\newblock \emph{{IEEE} Trans. Pattern Anal. Mach. Intell.}, 46\penalty0
  (12):\penalty0 7742--7759, 2024.

\bibitem[Shiba et~al.(2025)Shiba, Aoki, and Gallego]{Shiba25iccv}
Shintaro Shiba, Yoshimitsu Aoki, and Guillermo Gallego.
\newblock Simultaneous motion and noise estimation with event cameras.
\newblock In \emph{Int. Conf. Comput. Vis. (ICCV)}, 2025.

\bibitem[Stoffregen et~al.(2020)Stoffregen, Scheerlinck, Scaramuzza, Drummond,
  Barnes, Kleeman, and Mahony]{Stoffregen20eccv}
Timo Stoffregen, Cedric Scheerlinck, Davide Scaramuzza, Tom Drummond, Nick
  Barnes, Lindsay Kleeman, and Robert Mahony.
\newblock Reducing the sim-to-real gap for event cameras.
\newblock In \emph{Eur. Conf. Comput. Vis. (ECCV)}, pages 534--549, 2020.

\bibitem[Tulyakov et~al.(2022)Tulyakov, Bochicchio, Gehrig, Georgoulis, Li, and
  Scaramuzza]{Tulyakov22cvpr}
Stepan Tulyakov, Alfredo Bochicchio, Daniel Gehrig, Stamatios Georgoulis,
  Yuanyou Li, and Davide Scaramuzza.
\newblock {Time Lens}++: Event-based frame interpolation with parametric
  non-linear flow and multi-scale fusion.
\newblock In \emph{{IEEE} Conf. Comput. Vis. Pattern Recog. (CVPR)}, pages
  17755--17764, 2022.

\bibitem[Wang et~al.(2004)Wang, Bovik, Sheikh, and Simoncelli]{Wang04tip}
Zhou Wang, Alan~C. Bovik, Hamid~R. Sheikh, and Eero~P. Simoncelli.
\newblock Image quality assessment: From error visibility to structural
  similarity.
\newblock \emph{{IEEE} Trans. Image Process.}, 13\penalty0 (4):\penalty0
  600--612, 2004.

\bibitem[Weng et~al.(2021)Weng, Zhang, and Xiong]{Weng21cvpr}
Wenming Weng, Yueyi Zhang, and Zhiwei Xiong.
\newblock Event-based video reconstruction using transformer.
\newblock In \emph{Int. Conf. Comput. Vis. (ICCV)}, pages 2543--2552, 2021.

\bibitem[Wu et~al.(2024)Wu, Paredes-Vallés, and de~Croon]{Wu24icra}
Yilun Wu, Federico Paredes-Vallés, and Guido C. H.~E. de Croon.
\newblock Lightweight event-based optical flow estimation via iterative
  deblurring.
\newblock In \emph{{IEEE} Int. Conf. Robot. Autom. (ICRA)}, pages 14708--14715,
  2024.

\bibitem[Yang et~al.(2022)Yang, Wu, Shi, Lao, Gong, Cao, Wang, and
  Yang]{Yang22cvprw}
Sidi Yang, Tianhe Wu, Shuwei Shi, Shanshan Lao, Yuan Gong, Mingdeng Cao, Jiahao
  Wang, and Yujiu Yang.
\newblock {MANIQA}: Multi-dimension attention network for no-reference image
  quality assessment.
\newblock In \emph{{IEEE} Conf. Comput. Vis. Pattern Recog. Workshops (CVPRW)},
  pages 1191--1200, 2022.

\bibitem[You et~al.(2024)You, Cao, Yuan, Wang, Qiao, and Li]{You24arxiv}
Hongzhi You, Yijun Cao, Wei Yuan, Fanjun Wang, Ning Qiao, and Yongjie Li.
\newblock Vector-symbolic architecture for event-based optical flow.
\newblock \emph{ar{X}iv e-prints}, 2024.

\bibitem[Zhang et~al.(2018)Zhang, Isola, Efros, Shechtman, and
  Wang]{Zhang18cvprLPIPS}
Richard Zhang, Phillip Isola, Alexei~A. Efros, Eli Shechtman, and Oliver Wang.
\newblock The unreasonable effectiveness of deep features as a perceptual
  metric.
\newblock In \emph{{IEEE} Conf. Comput. Vis. Pattern Recog. (CVPR)}, 2018.

\bibitem[Zhang et~al.(2023)Zhang, Yezzi, and Gallego]{Zhang22pami}
Zelin Zhang, Anthony Yezzi, and Guillermo Gallego.
\newblock Formulating event-based image reconstruction as a linear inverse
  problem with deep regularization using optical flow.
\newblock \emph{{IEEE} Trans. Pattern Anal. Mach. Intell.}, 45\penalty0
  (7):\penalty0 8372--8389, 2023.

\bibitem[Zheng et~al.(2023)Zheng, Liu, Lu, Hua, Pan, Zhang, Tao, and
  Wang]{Zheng23arxiv}
Xu Zheng, Yexin Liu, Yunfan Lu, Tongyan Hua, Tianbo Pan, Weiming Zhang, Dacheng
  Tao, and Lin Wang.
\newblock Deep learning for event-based vision: A comprehensive survey and
  benchmarks.
\newblock \emph{ar{X}iv e-prints}, 2023.

\bibitem[Zhu et~al.(2018)Zhu, Yuan, Chaney, and Daniilidis]{Zhu18rss}
Alex~Zihao Zhu, Liangzhe Yuan, Kenneth Chaney, and Kostas Daniilidis.
\newblock {EV-FlowNet}: Self-supervised optical flow estimation for event-based
  cameras.
\newblock In \emph{Robotics: Science and Systems (RSS)}, pages 1--9, 2018.

\bibitem[Zhu et~al.(2019)Zhu, Yuan, Chaney, and Daniilidis]{Zhu19cvpr}
Alex~Zihao Zhu, Liangzhe Yuan, Kenneth Chaney, and Kostas Daniilidis.
\newblock Unsupervised event-based learning of optical flow, depth, and
  egomotion.
\newblock In \emph{{IEEE} Conf. Comput. Vis. Pattern Recog. (CVPR)}, pages
  989--997, 2019.

\end{thebibliography}

}

\end{document}